\documentclass[runningheads]{llncs}
\usepackage{graphicx}
\usepackage{amsmath,amssymb} % define this before the line numbering.
\usepackage{color}

\include{Definitions}

\usepackage{multirow}
\usepackage{wrapfig}
\usepackage{hhline}
\usepackage{tabularx}
\usepackage{makecell}
\usepackage{comment}
\usepackage{appendix}
\usepackage{algorithm2e}

\begin{document}
% \linenumbers
\pagestyle{headings}
\mainmatter
\def\ECCVSubNumber{6568}  % Insert your submission number here

\title{PACTran: PAC-Bayesian Metrics for Estimating the Transferability of Pretrained Models to Classification Tasks}

\begin{comment}
\titlerunning{ECCV-22 submission ID \ECCVSubNumber}
\authorrunning{ECCV-22 submission ID \ECCVSubNumber}
\author{Anonymous ECCV submission}
\institute{Paper ID \ECCVSubNumber}
\end{comment}

%\begin{comment}
\titlerunning{PACTran}
\author{Nan Ding,
Xi Chen,
Tomer Levinboim,
Soravit Changpinyo, 
Radu Soricut}
\authorrunning{N. Ding et al.}
\institute{Google Research\\
\email{\{dingnan,chillxichen,tomerl,schangpi,rsoricut\}@google.com}}
%\end{comment}

\maketitle

\begin{abstract}

With the increasing abundance of pretrained models in recent years, the problem of selecting the best pretrained checkpoint for a particular downstream classification task has been gaining increased attention. Although several methods have recently been proposed to tackle the selection problem (e.g. LEEP, H-score), these methods resort to applying heuristics that are not well motivated by learning theory. In this paper we present PACTran, a theoretically grounded family of metrics for pretrained model selection and transferability measurement. We first show how to derive PACTran metrics from the optimal PAC-Bayesian bound under the transfer learning setting. We then empirically evaluate three metric instantiations of PACTran on a number of vision tasks (VTAB) as well as a language-and-vision (OKVQA) task. An analysis of the results shows PACTran is a more consistent and effective transferability measure compared to existing selection methods.
%\keywords{Transfer Learning, PAC-Bayes}
\end{abstract}

\section{Introduction}

Recent advances in machine learning and neural networks have resulted in effective but extremely over-parameterized models \cite{t5,gpt3}, sometimes referred to as foundation models~\cite{foundation_models_review}. 
Despite the fact that their training recipe and data are often available, training such models requires access to computational resources that are well beyond the reach of an average machine learning user or group.
At the same time, many such model checkpoints (parameter snapshots at a particular training step) have been made publicly available in platforms such as Tensorflow-Hubs\footnote{\url{https://www.tensorflow.org/hub}}~\cite{tensorflow2015-whitepaper} and Huggingface\footnote{\url{https://huggingface.co/}}~\cite{wolf2019huggingface}, so that ML users who are interested in a certain model configuration need only write a few lines of code to initialize their own model from a public checkpoint, and continue fine-tuning on their downstream task of interest without incurring the cost of pretraining a model themselves.
However, as the number of such models and checkpoints increases, a natural question of selection arises -- is it possible to tell which initialization checkpoint is most suitable for a given downstream task without brute-force fine-tuning from all the available checkpoints?

To answer this question in the context of classification tasks, a number of existing approaches have been recently proposed.
For example, the LEEP metric~\cite{nguyen2020leep} assumes the pretrained model was trained on a source classification task, and then estimates the likelihood of an empirical predictor which maps source labels to target (downstream) labels for classifying the target data.
% (dingnan): previous description was wrong. now revised.
On the other hand, the H-score metric~\cite{bao2019information} casts the classification problem as a linear regression task involving the representations of the penultimate layer and the target labels.

Unfortunately, both these methods resort to heuristics or approximations to arrive at their estimate. 
Specifically, the empirical predictor used by LEEP is not the optimal solution to its associated objective function. Furthermore, the predictor and the metric are estimated on the same dataset, which is prone to overfitting.
On the other hand, the least squares 
solution of H-score is generally not a valid approximation to the commonly-used cross-entropy loss for classification, unless the dependence between the label and the input feature is weak~\cite{huang2019universal}, which rarely holds in practice.

In this paper we present PACTran, a theoretically grounded framework for deriving metrics that measure the transferability of pretrained models to downstream classification tasks.
Our framework seeks an optimal yet efficient PAC-Bayesian bound~\cite{mcallester1999some,germain2016pac} to the generalization error in a transfer learning setting, and the error is based on the cross-entropy loss between the prediction and the labels, as is commonly used in classification. 
That is, the PACTran framework enjoys at least one of two advantages compared to previous methods: 
(1) It is based on learning theory (as opposed to LEEP) through PAC-Bayesian bounds that measure the generalization gap, and 
(2) it is compatible with classification, since it relies on the cross-entropy loss (as opposed to H-score).

We instantiate the PACTran framework with three different priors, yielding three new transferability metrics: PACTran-Dirichlet, PACTran-Gamma and PACTran-Gaussian. 
Our experiments empirically evaluate and compare these new metrics against a number of baseline metrics over various image classification tasks in the Visual Task Adaptation Benchmark (VTAB)~\cite{zhai2019large}. Furthermore, we also evaluate our metrics over the multimodal Open-Knowledge VQA~\cite{balanced_vqa_v2,okvqa} task, which contains both image and text. 

\section{Transferability Metrics: A Quick Review}
In this section, we describe the transferability problem and review several transferability metrics which will be used as baselines in our experiments.
We begin by describing the setup of Transfer Learning, where the goal is to leverage knowledge acquired on a source task in order to solve a new target task.
Specifically, let $M$ denote a model checkpoint already pretrained to solve a source task,
%by mapping input representation $x \in \Xcal$ to output labels $z \in \Zcal$, 
and let $S$ denote a dataset for the target (downstream) task, 
such that $S = \cbr{(\xb_1, y_1), \ldots, (\xb_n, y_N)}$ with inputs $\xb \in \Xcal$ and target labels $y \in \Ycal$. Transfer learning seeks to transfer the knowledge already encoded in $M$ by finetuning from $M$ using $S$.

At the same time, given multiple pretrained checkpoints $M$, another problem arises -- is it possible to know which of the pretrained checkpoints is most suitable for the downstream task $S$, without incurring the cost of fine-tuning from each of them?
To this end, several effective and computationally efficient transferability metrics have been proposed and are summarized below.

\subsection{LEEP}
Given a checkpoint $M$ of a model pretrained on a source classification task, LEEP~\cite{nguyen2020leep} estimates the transferability of $M$ to the target dataset $S$ by first computing two types of probabilities: (1) the predicted distribution $M(\xb_i)$ over the source label set $\Zcal$ of the pretraining task, 
where we let $M(\xb_i)_z$ denote the output probability of the $z$-th source label, and (2) the empirical conditional distribution $\hat{p}(y|z)$ of the target label $y$ given the source label $z$,
\begin{align}
    \hat{p}(y|z) = \frac{\hat{p}(y, z)}{\sum_y \hat{p}(y, z)}, \text{where } \hat{p}(y, z) = \frac{1}{N} \sum_{(x_i, y_i)\in S} M(\xb_i)_z \cdot \delta(y_i = y). \label{eq:leep-pycz}
\end{align}
The LEEP measure is then defined as the logarithm of the marginal likelihood $\hat{p}(\yb | \xb)$ (called EEP) given the empirical predictor $\hat{p}(y|z)$ and $M(\xb)$, 
\begin{align}
    R_{LEEP} = \frac{1}{N} \sum_{(x_i, y_i)\in S} \log \hat{p}(y_i | \xb_i) = \frac{1}{N} \sum_{(x_i, y_i)\in S} \log \rbr{\sum_{z \in \Zcal} \hat{p}(y_i | z) M(\xb_i)_z }. \label{eq:leep}
\end{align}
LEEP was proposed as an improvement over the Conditional Entropy (CE) measure~\cite{tran2019transferability} 
which itself is an information theoretic approach that measures the transferability between two classification tasks by analyzing the correlation between their label sequences $Y = \cbr{y_1, \ldots, y_N}$ and $Z = \cbr{z_1, \ldots, z_N}$. 
%\begin{align}
%    H(Y|Z) = \sum_{y \in \Ycal} \sum_{z \in \Zcal} \hat{p}(y, z) \log \frac{\hat{p}(y, z)}{\hat{p}(z)}, \label{eq:ce}
%\end{align}
%where $\hat{p}(y, z) = \frac{1}{n} \sum_i \delta(y_i=y) \delta(z_i=z)$ and $z_i = \argmax_{z \in \Zcal} M(\xb_i)_z$. 
In \cite{nguyen2020leep} the authors show that LEEP is an upper bound of negative CE and outperforms it empirically as a transferability metric.
However, from a theoretic stand point the LEEP formulation suffers from a few deficiencies. 
For example, plugging in the empirical conditional distribution $\hat{p}(y|z)$ into Eq.\eqref{eq:leep} is not guaranteed to maximize the target log-likelihood $\log \hat{p}(\yb | \xb)$. 
Furthermore, both $\hat{p}(y|z)$ and $\log \hat{p}(\yb | \xb)$ are computed over $S$, which make the latter prone to overfitting and behave more similarly to training error as opposed to generalization error.

\subsection{$\Ncal$-LEEP}
Another limitation of the LEEP measure (as well as CE) is that it can only be applied to measure the transferability of pretrained classification models. 
In addition, LEEP's performance degrades when the number of source classes is considerably smaller than the number of target classes.
To overcome these issues, several methods that propose using the outputs $f(\xb)$ of the penultimate layer.

In $\Ncal$-LEEP~\cite{li2021ranking}, the authors suggest to first apply Principal Component Analysis (PCA) on the penultimate layer outputs $f(\xb)$ to reduce their dimension and then fit a Gaussian Mixture Model (GMM) to the PCA-reduced representation $\sbb(\xb)$, so that $p(\sbb) = \sum_{v \in \Vcal} \alpha_v \Ncal(\sbb|\ub_v, \Sigmab_v)$ and the posterior of the cluster assignment 
\begin{align}
p(v | \xb) = p(v | \sbb) \propto \alpha_v \Ncal(\sbb|\ub_v, \Sigmab_v) \label{eq:nleep}
\end{align}
are used to replace $M(\xb)_z$ in Eq.\eqref{eq:leep-pycz}. 
The rest of the procedure follows the same as in Eq. \eqref{eq:leep}. 

The $\Ncal$-LEEP method~\cite{li2021ranking} conjectures that the cluster assignment $p(v | \sbb)$ is more reliable than the class assignment $M(\xb)_z$, because the GMM fitting is learned from the downstream target data, while the softmax classifier of LEEP is learned over the pretrained source data.
Since $\Ncal$-LEEP is a extension of LEEP, it also inherits its aforementioned problems such as the non-optimality of the log-likelihood, as well as the lack of generalization consideration.

\subsection{H-Score}
The H-score~\cite{bao2019information} transferability metric is also not restricted to pretrained classifiers.
The idea for H-score comes from the matrix factorization of the divergence transition matrix (DTM) $\tilde{B} = \frac{p(x, y)}{\sqrt{p(x)p(y)}} - \sqrt{p(x)p(y)}$, for discrete random variables $\xb$ and $\yb$. 
It is shown in \cite{huang2019universal} that, under the assumption of sufficiently small $\tilde{B}$, the solution of the cross-entropy loss coincides with the following solution of the matrix decomposition: 
\begin{align}
    \Psi^* = \argmin_{\Psi} \|\tilde{B} - \Phi(\xb)^\top \Psi  \|_F^2,\; \text{where }\; \Phi(\xb) = \sqrt{p(\xb)} f(\xb). \label{eq:hscore-mf}
\end{align}
After plugging in the least squares solution $\Psi^* = \tilde{B}\Phi(\Phi^\top \Phi)^{-1}$, Eq.\eqref{eq:hscore-mf} becomes $\|\tilde{B}\|_F^2 - \|\tilde{B} \Phi(\Phi^\top \Phi)^{-\frac{1}{2}} \|_F^2$, in which the second term is defined as the H-score:
\begin{align}
H =& \|\tilde{B} \Phi(\Phi^\top \Phi)^{-\frac{1}{2}} \|_F^2
= \tr(\text{cov}(f(\xb)))^{-1} \text{cov}(\EE_{p(x|y)}[f(\xb) | y]).
\end{align}
Compared to LEEP, H-score is more theoretically solid, in that it is optimal with respect to its loss. However, the key drawback of the H-score is that the optimality is based on the least squares objective, which is rarely used for classification. 
As proven in \cite{huang2019universal}, the least squares solution is a valid approximation to the cross-entropy classification loss only when label $\yb$ and input $\xb$ are weakly dependent, which is clearly not the case in general.

\subsection{LogME}
Similarly to H-Score, the Log Marginal Evidence (LogME)~\cite{you2021logme} transferability metric also uses a least squares objective function.
However, to avoid over-fitting, instead of directly minimizing the Gaussian based log-likelihood (a.k.a the squared-loss) $\wb^* = \argmin_{\wb} \|\yb - \fb^\top \wb\|_F^2$, LogME uses Bayesian averaging to improve its generalization ability. That is, the LogME metric uses the marginal evidence of the target task $p(y | \fb) = \int p(\wb) p(y | \fb, \wb) d \wb$.
When $p(\wb)$ is defined as a Gaussian prior and $p(y | \fb, \wb)$ is a Gaussian likelihood, then $p(y|\fb)$ can be analytically estimated.

LogME shares the same theoretical problems as H-score due to its dependence on the least squares objective, however, by relying on the marginal evidence it is less prone to overfitting~\cite{you2021logme} which potentially improves its generalization ability.

\section{PACTran}
In this section, we first briefly review the PAC-Bayesian bound~\cite{mcallester1999some,germain2009pac} in the supervised learning setting. 
We then show how to leverage this bound for measuring transferability as the PACTran metric (Section \ref{sec:parctran}).
Specifically, we derive three instances of the PACTran metric based on the cross entropy loss using three different prior distributions: two based on conjugate priors with the Dirichlet and Gamma distributions  (Section \ref{sec:pt-dir} and \ref{sec:pt-gam}), and a third with a non-conjugate Gaussian prior (Section \ref{sec:pt-gau}). 

\subsection{PAC-Bayesian Bounds for Supervised and Transfer Learning}
\label{sec:parctran}
Consider a learning task with data distribution $D$ where examples are denoted as $u = (x, y)$. 
A hypothesis $h$ from the hypothesis space $H$ allows us to make predictions for each input $x$. 
The quality of the predictions is measured by a loss function $l(h, u)$, and the goal is to minimize the expected loss 
$L(h, D) = \EE_{u \sim D} l(h, u)$. 
Typically, the data distribution $D$ is unknown, and instead we are given a set of $N$ (training) examples 
$S \sim D^N = \{u_i \sim D\}_{i=1}^N$, in which case the empirical error on $S$ is simply
$\hat{L}(h, S) = \frac{1}{N} \sum_{i=1}^N l(h, u_i)$. 
The gap between $L(h, D)$ and $\hat{L}(h, S)$ is known as the generalization gap of $h$. Based on this, various forms of PAC (Probably Approximately Correct) bounds have been studied in the ML community over the last few decades \cite{bousquet2003introduction,tripuraneni2020theory}.

A key drawback of the PAC bounds is that the worst-case analysis (via the union bound over all $h \in H$) makes the bound vacuous for modern machine learning approaches~\cite{48754,zhang2021understanding}. 
To address this drawback, PAC-Bayesian learning~\cite{mcallester1999some,germain2009pac} goes one step further by bounding the generalization gap of distributions over $H$, which can be optimized to obtain a non-vacuous bound~\cite{dziugaite2017computing,48754}.
In particular, let us assume that the learner has some prior knowledge of the hypothesis space $H$ in the form of a prior distribution $P(h)$. 
Once the learner observes a training dataset $S$, it updates its prior $P$ into a posterior distribution $Q$. 
The expected error of the posterior $Q$ is called the Gibbs error $L(Q, D) = \EE_{h \sim Q} L(h, D)$, and its empirical counterpart is $\hat{L}(Q, S) = \EE_{h \sim Q} \hat{L}(h, S)$.
The PAC-Bayesian framework provides the following upper bound~\cite{mcallester1999some,germain2016pac} over $L(Q, D)$ based on its empirical estimate $\hat{L}(Q, S)$:

\begin{theorem}\cite{germain2016pac}
Given a data distribution $D$, a hypothesis space $H$, a prior $P$, a confidence level $\delta \in (0, 1]$, and $\lambda > 0$, with probability at least $1 - \delta$ over samples $S \sim D^N$, for all posterior $Q$,
\begin{align}
    L(Q,D)
    &\le \hat{L}(Q, S) + \frac{1}{\lambda} D_{KL}(Q \| P) + C(\delta, \lambda, N) \label{eq:super}
\end{align}
where $C(\delta, \lambda, N)$ is a constant independent of the posterior $Q$.
\end{theorem}
The hyperparameter $\lambda$ can be adjusted to balance between the divergence and the constant $C$ terms, where a common choice is $\lambda \propto N$ (see \cite{germain2016pac,rothfuss2021pacoh,ding2021bridging}).

In the transfer learning setting, starting from a pretrained checkpoint $M$ that is encoded within the prior $P(h)$, $L(Q, D)$ measures the generalization error of a posterior $Q$ after it was finetuned over the downstream data $S$. 
Furthermore, by minimizing the RHS of the bound (Eq.~\eqref{eq:super}) with respect to $Q \in \Qcal_M$, one can obtain a posterior $Q$ that has low transfer error $L(Q, D)$.
Therefore, to measure the transferability of a pretrained checkpoint $M$, we define a family of metrics \emph{PACTran} by optimizing the PAC-Bayesian bound (ignoring the constant $C$ since it is the same for all checkpoints):
\begin{align}
\min_{Q \in \Qcal_M} \hat{L}(Q, S) + \frac{1}{\lambda} D_{KL}(Q \| P). \label{eq:pactran}
\end{align}

For computational efficiency, we restrict the domain of $\Qcal_M$ in which the feature network of the pretrained checkpoint $M$ remains fixed. Since all $h \in dom(\Qcal_M)$ shares the same feature network, we can simplify $P$ as the prior distribution of the top classification layer of the network only; and $Q$ as the posterior distribution of the top layer after finetuning. 
Despite this restriction, PACTran appears promising in comparing the transferability of pretrained checkpoints even after full-model finetuning.

According to~\cite{germain2016pac}, the so-called Gibbs posterior $Q^*$ that minimizes the objective of Eq.\eqref{eq:pactran} takes the form of  
$Q^*(h) = P(h) \exp(-\lambda \hat{L}(h, S)) / Z(S)$,
where $Z(S)$ is equal to the marginal evidence $\int P(h) \exp(-\lambda \hat{L}(h, S)) dh$. 
Plugging in $Q^*(h)$ back into Eq.\eqref{eq:pactran}, the resulting optimal PAC-Bayesian bound equals to $-\frac{1}{\lambda} \log Z(S)$.
Note however, that computing $\log Z(S)$ is only analytically feasible, 
when the prior $P(h)$ and the likelihood function $\exp(-\lambda \hat{L}(h, S))$ are conjugate, for example, when both are Gaussians as in LogME~\cite{you2021logme}. 

In this paper, we focus on metrics for which $\hat{L}(h, S)$ is based on the cross-entropy loss, as it is more compatible with classification tasks (in which case $\exp(-L(Q, D))$ is an estimate of the expected test accuracy).
From a theoretical perspective, this makes the PACTran metric preferable to LEEP (which is not optimal over the cross entropy loss) as well as LogME and H-score metrics (whose proposed solution is based on the squared loss instead of the classification loss of the downstream task).

In what follows, we derive three instantiations of the bound using the conjugate Dirichlet and Gamma priors whose solution can be found with a fast variational approach, and the non-conjugate Gaussian prior, which requires gradient optimization.

\subsection{PACTran with a Dirichlet Prior}
\label{sec:pt-dir}
Given the target data $S = \cbr{(\xb_1, y_1), \ldots, (\xb_N, y_N)} \in (\Xcal, \Ycal)$, let us assume that the pretrained model $M$ provides a probability vector $M(\xb)$ where $\sum_z M(\xb)_z = 1$.
Here, $z \in \Zcal$ can either be defined over the set of source label as in LEEP or over the Gaussian clusters as in $\Ncal$-LEEP. 
We restrict the top layer to a set of 
$l_1$-normalized vectors $\Wb = \cbr{\wb_1, \ldots, \wb_{|\Zcal|}}$ in the probability simplex, where for each vector $\wb_z$ we have $\sum_{y \in \Ycal} w_{yz} = 1, w_{yz} \ge 0$, 
and then define the marginal likelihood as:
\begin{align}
p(y_i| \xb_i, \Wb) = \sum_z p(y_i, z | \xb_i, \Wb) = \sum_z M(\xb_i)_z w_{y_i z}.  \label{eq:pac-lda-pyx}
\end{align}
We assign a Dirichlet prior $P(\wb_z)$ on these vectors and let $\lambda = N$ for simplicity. Using the above definitions, we can rewrite $\log Z(S)$ as
\begin{align}
    &\log \int \prod_z P(\wb_z) \prod_i\rbr{\sum_z p(y_i, z_i=z | \xb_i, \Wb)} d \Wb \nonumber\\
%    =&\log \int \prod_z \rbr{\frac{\Gamma(\sum_y {\alpha_y}) }{\prod_y \Gamma(\alpha_y)}\prod_y w_{yz}^{\alpha_y - 1}} \prod_i\rbr{\sum_z M(\xb_i)_z w_{y_i z}} d \Wb \nonumber\\
%    =&\log \sum_{z_1} \ldots \sum_{z_n}\int \prod_z \rbr{\frac{\prod_z \Gamma(\sum_y {\alpha_y})}{\prod_z \prod_y \Gamma(\alpha_y)}\prod_y w_{yz}^{\alpha_y - 1}} \prod_i \rbr{ M(\xb_i)_{z_i} w_{y_i z_i}} d \Wb \nonumber\\
    =&\log \sum_{z_1} \ldots \sum_{z_N}\int \rbr{\frac{\prod_z \Gamma(\sum_y {\alpha_y})}{\prod_z \prod_y \Gamma(\alpha_y)}} \prod_z \prod_y w_{yz}^{n_{yz} + \alpha_y - 1} \prod_i M(\xb_i)_{z_i} d \Wb, \label{eq:pac-lda}
\end{align}
where $\Gamma(\cdot)$ is the well-known Gamma function. 
The form of the resulting Bayesian model is similar to Latent Dirichlet Allocation (LDA)~\cite{blei2003latent}. 
Evaluating Eq.\eqref{eq:pac-lda} exactly is considered intractable, as it involves a summation over $\zb$ which has $|\Zcal|^N$ different configurations. 
%One could attempt a Monte-Carlo sampling based approximation, however in practice we find that the resulting estimation is unreliable, having a large variance. 
Therefore, we turn to the variational inference approach as in~\cite{blei2003latent,blei2017variational} to optimize the evidence lower bound (ELBO). 
The PACTran-Dirichlet is the negation of the optimal ELBO, and equals to (see details in A.1): %\ref{sec:derive-pacdir}
{\small
\begin{align}
    \sum_z \rbr{\log C(\tilde{\alphab_z}) - \log C(\alphab_z) + \sum_i q^*(z_i=z)\rbr{\log q^*(z_i=z) - \log M(\xb_i)_z}}, \label{eq:pac-lda-v}
\end{align}}
where, 
\begin{align*}
    &q^*(z_i = z) = \text{softmax}\rbr{\log M(\xb_i)_z + \Psi(\tilde{\alpha}_{y_i z}) - \Psi(\sum_y \tilde{\alpha}_{y z})}, \\
    &\tilde{\alpha}_{yz} = \alpha_{yz} + \sum_i q^*(z_i=z)\delta(y_i=y),\; \text{and }\; C(\alphab_z) = \frac{\Gamma(\sum_y \alpha_{yz})}{\prod_y \Gamma(\alpha_{yz})},
\end{align*}
where $\Psi(\cdot)$ denotes the digamma-function. 

It is worth noting that the PACTran-Dirichlet metric Eq.\eqref{eq:pac-lda-v} is a valid PAC-Bayesian upper bound to the generalization error (up to a constant). That is because Eq.\eqref{eq:pac-lda-v} is the negation of ELBO which upper bounds the negative log evidence $-\log Z(S)$ which itself is an upper bound of $L(Q^*, D)$. Furthermore, both upper bounds are optimally tight with respect to their hypothesis spaces in consideration: the variational distribution $q^*$ optimizes the ELBO over all the independent approximate distributions $q$, and the Gibbs posterior $Q^*(h)$ optimizes the PAC-Bayes bound \eqref{eq:super} over all the base-learner $Q$. 

\subsection{PACTran with a Gamma Prior}
\label{sec:pt-gam}
Instead of using a set of $l_1$-normalized vectors $\Wb$, we can also relax the constraint by working on a matrix of non-negative variables $\Vb = \cbr{v_{yz}}$ whose prior is chosen to be the gamma distribution $P(v_{yz})=Gamma(a_y, b)$. Unlike the normalized vectors $\Wb$, where $\sum_{y \in \Ycal} \sum_{z \in \Zcal} M(\xb)_z w_{yz} = 1$ is automatically satisfied, when using unnormalized $\Vb$, we need to normalize the output explicitly, 
\begin{align}
p(y_i, z | \xb_i, \Vb) = \frac{M(\xb_i)_z v_{y_i z}}{\sum_{y \in \Ycal} \sum_{z \in \Zcal} M(\xb_i)_z v_{yz}}.  \label{eq:pac-gamma-pyx}
\end{align}
Note that $M(\xb_i)$ is also not required to be normalized, which potentially makes the use case of Eq.\eqref{eq:pac-gamma-pyx} broader. Even with a normalized $M(\xb_i)$, Eq.\eqref{eq:pac-gamma-pyx} strictly subsumes Eq.\eqref{eq:pac-lda-pyx}, because the former is only normalized once, while the latter is normalized $|\Zcal|$ times for each $\wb_z$. In addition, since $v_{yz}$ appears in both denominator and numerator, their scaling cancels out. Therefore, we fix a simple scaling coefficient $b = 1$ for all Gamma priors.

The rest of the Bayesian inference is similar to that of PACTran-Dirichlet. 
The PACTran-Gamma metric is the negative ELBO after applying the variational principles, and equals to (see details in A.2): %\ref{sec:derive-pacgam}
\begin{align}
& \sum_y \sum_z \rbr{\log \Gamma(a_y) - \log \Gamma(\tilde{a}_{yz})} + \sum_i \log \tilde{\lambda}_i \nonumber\\
+& \sum_i \sum_z q^*(z_i=z)\rbr{\log q^*(z_i=z)- \log M(x_i)_z}, \label{eq:pac-gamma-v}
\end{align}
where,
\begin{align*}
    &q^*(z_i=z) = \text{softmax}\rbr{\log M(\xb_i)_z + \Psi(\tilde{a}_{y_i z})},\\
    &\tilde{a}_{yz} = a_y + \sum_i q^*(z_i=z)\delta(y_i=y),\;\;
    \tilde{\lambda}_i = \sum_y \sum_z M(\xb_i)_z \tilde{a}_{yz} .
\end{align*}
PACTran-Gamma metric is also a valid PAC-Bayesian upper bound to the generalization error, for the same reasons as the PACTran-Dirichlet metric.

\subsection{PACTran with a Gaussian Prior}
\label{sec:pt-gau}
In the previous sections, we focus on the cases when the source model outputs normalized (in the Dirichlet prior case) or non-negative vectors (in the Gamma prior case). When the pretraining model is not based on classification tasks, one needs to add additional components (such as the Gaussian mixture models in $\Ncal$-LEEP) to obtain those outputs. Here, we present another metric PACTran-Gaussian which relies only on penultimate layer representations $f(\xb)$. In PACTran-Gaussian, the prior $P$ and posterior $Q$ are both Gaussian distributions, where $P(\thetab) \sim \Ncal(0, \sigma_0^2\Ib)$ and $Q(\thetab) \sim \Ncal(\thetab_q, \Sigmab_q)$. 
%Although it is possible to use a full multivariate covariance matrix for $\Sigmab_q$, 
For computational efficiency, we consider $\Sigmab_q = \sigma_q^2 \Ib$ only. 
Note that although both LogME and PACTran-Gaussian use Gaussian priors and posteriors on $\thetab$, a main difference is that the former applies the squared loss, while the latter applies the cross-entropy loss (see more discussions in A.3). %\ref{sec:logme_pt_gauss}
However, since the Gaussian prior is not conjugate to the exponentiated cross-entropy loss, we derive the bound using 2nd order approximations and a reparameterization trick as in~\cite{tsuzuku20a},
\begin{align}
    &\hat{L}(Q, S) + \frac{1}{\lambda} D_{KL}(Q \| P) \nonumber\\
    \simeq & \hat{L}(\thetab_q, S) + \EE_{\epsilonb \sim \Ncal(0, \Ib)} [\sigma_q \epsilonb^{\top} \nabla \hat{L}(\thetab_q, S) + \frac{\sigma_q^2}{2} \epsilonb^{\top} \nabla^2 \hat{L}(\thetab_q, S) \epsilonb \nonumber\\
    &\qquad + \frac{1}{\lambda} \log \Ncal(\thetab_q + \sigma_q \epsilonb | \thetab_q, \sigma_q^2 \Ib) - \frac{1}{\lambda} \log \Ncal(\thetab_q + \sigma_q \epsilonb | 0, \sigma_0^2 \Ib)] \nonumber\\
    %=& \hat{L}(\thetab_*, S) + \EE_{\epsilon \sim \Ncal(0, 1)} [\sigma_* \epsilon^{\top} \nabla \hat{L}(\thetab_*, S) + \frac{\sigma_*^2}{2} \epsilon^{\top} \nabla^2 \hat{L}(\thetab_*, S) \epsilon \nonumber\\
    %&\qquad - \frac{1}{2\lambda} \rbr{KD \log \sigma_*^2 + \epsilon^2} + \frac{1}{2\lambda} (KD \log \sigma_0^2 + \frac{\|\thetab_*\|_F^2 + 2 \sigma_* \epsilon^{\top} \thetab_* + \sigma_*^2 \epsilon^2}{\sigma_0^2})] \nonumber\\
    =& \hat{L}(\thetab_q, S) + \frac{\sigma_q^2}{2} \text{Tr}(\nabla^2 \hat{L}(\thetab_q, S)) + \frac{KD}{2\lambda} (\log \frac{\sigma_0^2}{\sigma_q^2} - 1 + \frac{\sigma_q^2}{\sigma_0^2} + \frac{\|\thetab_q\|_F^2}{KD \sigma_0^2}).  \label{eq:pac-gauss-1}
\end{align}
The results of minimizing Eq.~\eqref{eq:pac-gauss-1} with respect to $\sigma_q$ and $\thetab_q$ yield the following optimal $\sigma_*$ and $\thetab_*$ (see details in A.3), %\ref{sec:derive-pacgau}
\begin{align*}
    \frac{\sigma_0^2}{\sigma_*^2} &= 1 + \frac{\beta}{KD} \text{Tr}(\nabla^2 \hat{L}(\thetab_*, S)),\ \ \ \ 
    \thetab_* = \argmin_{\thetab_q} \cbr{\hat{L}(\thetab_q, S) + \frac{\|\thetab_q\|_F^2}{2 \beta}},
\end{align*}
where $\beta = \lambda \sigma_0^2$,  So that we reach the following PACTran-Gaussian metric,
\begin{align}
    \underbrace{\hat{L}(\thetab_*, S) + 
    \frac{\|\thetab_*\|_F^2}{2 \beta}}_{RER} + 
    \underbrace{\frac{KD \sigma_0^2}{2 \beta} \log \frac{\sigma_0^2}{\sigma_*^2}}_{FR}.  \label{eq:pac-gauss-2}
\end{align}
In Eq.\eqref{eq:pac-gauss-2}, the first two terms are simply the $l_2$-regularized empirical risk (RER). The third term is a "flatness regularizer" (FR) that involves the trace of the Hessian of the empirical risk $\text{Tr}(\nabla^2 \hat{L}(\thetab_*, S))$ and has a simple closed-form solution for the cross-entropy loss (provided in A.3). %\ref{sec:derive-pacgau} 
It is accepted wisdom that a model generalizes better when its optimum is relatively flat~\cite{tsuzuku20a,dziugaite2017computing,neyshabur2017exploring} (low trace of Hessian). Empirically, we observe that the FR term is extremely effective in preventing the metrics from overfitting even though metric evaluation is done only on the training set. 

It is also worth noting that there are two subtle, yet critical, differences between the derivations of our bound Eq.\eqref{eq:pac-gauss-2} and the ones in~\cite{tsuzuku20a}. 
First, our mean parameter $\thetab_*$ is a minimum of Eq.\eqref{eq:pac-gauss-1}, while in \cite{tsuzuku20a} it is an arbitrary model parameter. Second, in~\cite{tsuzuku20a}, $\sigma_0$ and $\sigma_*$ were tied together during the optimization of $\sigma_*$, which violates the assumption of the PAC-Bayes theorem where the prior must be data independent. Instead, our $\sigma_*$ is not only optimal, but also leaves $\sigma_0$ data independent.

\section{Empirical Studies}
\label{sec:experiments}
In this section, we evaluate the PACTran metrics: PACTran-Dirichlet, PACTran-Gamma and PACTran-Gaussian, over several transfer learning benchmarks, and compare them against other existing transferability metrics including LEEP, NCE, $\Ncal$-LEEP, H-Score, LogME.
\vspace{-0.1in}
\subsection{The Neural Checkpoint Ranking Benchmark (NeuCRaB)}
\label{sec:neucrab}
\subsubsection{Pretrained Checkpoints}
Following NeuCRaB~\cite{li2021ranking} (Group I), we collected a set of 16 ResNet-50 based checkpoints trained with various types of supervision. 
These checkpoints were pretrained on ImageNet with different training strategies, which
include 5 models via self-supervised learning (Jigsaw~\cite{jigsaw_noroozi2017}, Relative Patch Location~\cite{relative_doersch2016}, Exemplar~\cite{exemplar}, Rotation~\cite{rotation}, and Sup-Rotation~\cite{semi_rotation_exemplar}), 
6 models via discriminators of generative models (WAE-UKL~\cite{wae_ukl}, WAE-GAN, WAE-MMD~\cite{wae_gan_mmd}, Cond-BigGAN, Uncond-BigGAN~\cite{biggan}, and VAE~\cite{vae_2013}), 
2 via semi-supervised learning (Semi-Rotation-10\% and Semi-Exemplar-10\%~\cite{semi_rotation_exemplar}), one with a hybrid supervised loss (Sup-Exemplar-100\%~\cite{semi_rotation_exemplar}), 
one by supervised learning of a standard Resnet50 (Sup-100\%~\cite{sup_100_img}), and lastly, one by supervised learning of a Resnet50 with identity mappings (Feature Vector~\cite{feature_vec}). 
\vspace{-0.1in}
\subsubsection{Downstream Tasks}
Following NeuCRaB~\cite{li2021ranking}, we adopt the Visual Task Adaptation Benchmark (VTAB)~\cite{zhai2019large} and study diverse downstream tasks. 
The original NeuCRaB only contains four tasks: Caltech101~\cite{caltech101_FeiFei2004}, Flowers102~\cite{flowers102_Nilsback08}, Patch Camelyon~\cite{patch_camelyon} and Sun397~\cite{sun397_Xiao:2010}. 
In order to compare the transferability metrics on a wider variety of downstream tasks, we added 5 more tasks: 
DMLAB~\cite{zhai2019large}, 
CBIS-DDSM~\cite{CBIS_DDSM_Citation}, 
Cifar10~\cite{cifar10_Krizhevsky09}, 
Oxford IIIT Pet~\cite{oxford_iiit_pet_parkhi12a} and Smallnorb(azimuth)~\cite{smallnorb_LeCun2004}. 
These new tasks not only enrich the task categories, but also span the full range of the number of classes per tasks (single-digit, double-digit, and 100+ classes), which allows us to analyze the performance of transferability metrics according to the number of classes. 
In particular, we group these tasks according to the number of output classes: tasks with 100+ classes include Caltech101 (102 classes), Flowers102 (102 classes), Sun397 (397 classes); tasks with 10-99 classes include Cifar10 (10 classes), Oxford IIIT Pet (37 classes) and Smallnorb(azimuth) (18 classes); tasks with 2-9 classes include Patch Camelyon (2 classes), DMLAB (6 classes), and CBIS-DDSM (5 classes).
\vspace{-0.1in}
\subsubsection{Evaluating the Transferability Metrics}
We use the Kendall-Tau rank correlation coefficient to correlate between the transferability metric scores and the testing error of the finetuned checkpoints. 
The "ground-truth" testing error that corresponds to each pretrained checkpoint $M$ is obtained by finetuning $M$ on the downstream training set multiple times and 
setting the ground-truth testing error $e_M$ to the lowest test error among the runs (See details in B.3). %\ref{sec:finetune}
\vspace{-0.1in}
\subsubsection{Experimental Settings}
Since it is crucial for a transferability metric to be highly efficient compared to the finetuning, we focus our experiments on limited-data settings. 
Let $K$ denote the number of classes, $D$ the feature dimension and $N$ the number of examples for computing the metric. 
We consider three data settings with increasing average number of samples per class $N/K \in \cbr{2, 5, 10}$ (to avoid having too few examples, we also set a lower bound for $N \ge 20$). 
For each $N/K$ setting, we subsample $N$ samples from the training set of each downstream task 5 times.
The transferability metrics are then evaluated over the 5 splits and their 
average Kendall-Tau correlation is reported.
Compared to evaluating the metrics on the full training set, the limited-data setting significantly reduces the cost of penultimate-layer feature extraction, which is usually orders of magnitude more expensive than computing the metrics themselves (see Table.~\ref{tab:vtab10-gflops}). 

Besides the aforementioned baseline transferability metrics (LEEP, $\Ncal$-LEEP, H-score, LogME),
we also include two additional metrics based on linear classification. 
The LINEAR metric is based on the training loss of a regularized linear classifier (the sum of the first two terms of Eq.\eqref{eq:pac-gauss-2}). 
The second metric LINEAR-VALID splits the subsampled dataset into two equally sized folds, trains a linear classifer on one fold and computes the validation error on the other. 
The regularizing coefficients for both metrics are $\beta \in \cbr{0.1, 1.0, 10} \cdot N$. 
For LINEAR-VALID, the model with the lowest validation error is chosen. 
For LINEAR, since there is no validation set, we select the $\beta$ that maximizes the Kendall correlation between the loss and LINEAR-VALID's validation error across checkpoints.

For $\Ncal$-LEEP, we follow the recipe from \cite{li2021ranking} and set the PCA energy percentage to 80\% and the number of Gaussian components to the number of classes in the downstream task. 
For PACTran-Dirichlet and PACTran-Gamma, we set $\alpha_y = \hat{p}(y)$. 
For PACTran-Gaussian, we report two sets of results: 
In PT-Gauss$_{fix}$, we fix the two hyperparameters to $\beta = 10 N$ and $\sigma_0^2 = \frac{100}{D}$.
In PT-Gauss$_{grid}$, we perform a hyperparameters grid-search over $\beta \in \cbr{0.1, 1.0, 10} \cdot N$ and $\sigma_0^2 \in \cbr{1.0, 10, 100, 1000} \cdot \frac{1}{D}$ and select the hyperparameters $(\beta, \sigma_0^2)$ that maximize the 
Kendall correlation between the PT-Gauss$_{grid}$ metric scores and LINEAR-VALID's validation errors across checkpoints.
More detailed discussions about the hyperparameters are available in B.5. %\ref{sec:hparams-pac-gauss}

\begin{table}[!ht]
\centering
  \begin{tabular}{c | c | c | c | c  }
$N/K=2$ & 100+ classes & 10-99 classes & 2-9 classes & Average \\\hline
%LEEP & 0.080 & -0.014 & 0.013 & 0.027  \\
LEEP & 0.202 & 0.005 & 0.041 & 0.083  \\
$\Ncal$-LEEP & 0.723 & 0.401 & 0.077 & 0.401 \\
H-score & 0.413 & 0.106 &0.185 & 0.235 \\
LogME & 0.308 & 0.067 &0.071 & 0.149 \\
LINEAR & 0.231 & 0.072 & 0.114 & 0.139\\
LINEAR-VALID & 0.750 & 0.309 & 0.063 & 0.374\\ \hline
$\Ncal$-PT-Dir & 0.760 & 0.327 &0.099 & 0.395 \\
$\Ncal$-PT-Gam & 0.763 & 0.333 &0.108 & 0.401 \\
PT-Gauss$_{grid}$ & {\bf 0.868} & 0.664 & {\bf 0.509} & {\bf 0.680} \\
PT-Gauss$_{fix}$ & 0.770 & {\bf 0.683} & {\bf 0.509} & 0.654 \\\hline\hline
  \end{tabular}
  \begin{tabular}{c | c | c | c | c  }
$N/K=5$ & 100+ classes & 10-99 classes & 2-9 classes & Average  \\\hline
%LEEP & 0.012 & 0.025 & 0.067& 0.035 \\
LEEP & 0.112 & 0.082 & 0.023 & 0.109  \\
$\Ncal$-LEEP & 0.795 & 0.536 & 0.096 & 0.476  \\
H-score & 0.412 & 0.141 & 0.118 & 0.224 \\
LogME & 0.421 & 0.093 & 0.075 & 0.196\\
LINEAR & 0.253 & 0.084 & 0.122 & 0.153  \\
LINEAR-VALID  & 0.807 & 0.411 & 0.044  & 0.420 \\ \hline
$\Ncal$-PT-Dir & {\bf 0.826} &0.458 &0.140 & 0.475 \\
$\Ncal$-PT-Gam & {\bf 0.825} &0.462 &0.151 & 0.479 \\
PT-Gauss$_{grid}$ & 0.793 & {\bf 0.716} & 0.412 & 0.641  \\
PT-Gauss$_{fix}$ & {\bf 0.832} & 0.675 & {\bf 0.512} & {\bf 0.673}  \\\hline\hline
  \end{tabular}
  \begin{tabular}{c | c | c | c | c  }
$N/K=10$ & 100+ classes & 10-99 classes & 2-9 classes & Average \\\hline
%LEEP & 0.036 & 0.041 & 0.032& 0.036 \\
LEEP & 0.276 & 0.079 & 0.049 & 0.134  \\
$\Ncal$-LEEP & 0.822 &0.520 &0.148 & 0.497  \\
Hscore & 0.461 &0.318 &0.158 & 0.313 \\
LogME & 0.488 & 0.138 & 0.073 & 0.233 \\
LINEAR & 0.325 & 0.089 & 0.109 & 0.174 \\
LINEAR-VALID & {\bf 0.835} & 0.482  & 0.123 & 0.480 \\ \hline
$\Ncal$-PT-Dir & {\bf 0.839} & 0.446 & 0.134 & 0.473 \\
$\Ncal$-PT-Gam & {\bf 0.839} & 0.452 & 0.140 & 0.477 \\
PT-Gauss$_{grid}$ & 0.769 & {\bf 0.678} & 0.429 & 0.625 \\
PT-Gauss$_{fix}$ & 0.778 & 0.609 & {\bf 0.534} & {\bf 0.641} \\\hline
  \end{tabular}
  \caption{Kendall-Tau correlations on the NeuCRaB experiments with different $N/K$, where $K$ is the number of classes, $N$ the number of examples for computing the metric.}
  \label{tab:vtab}
  \vspace{-0.3in}
 \end{table}
\vspace{-0.1in}
\subsubsection{Results and Analysis}
Table~\ref{tab:vtab} reports the Kendall-Tau correlations of the various transferability metrics. 
Table~\ref{tab:vtab10-gflops} reports the GFLOPS per metric for each task as well as those of the feature extraction stage from the pretrained checkpoints. 
For LINEAR and PT-Gauss$_{grid}$, we include the GFLOPS for all hyperparameter runs as well as hyperparameter selection for LINEAR-VALID.

Although LEEP is the fastest algorithm, its averaged performance is worse than most other metrics.
All other metrics employ more expensive components (PCA and GMM for $\Ncal$-LEEP and $\Ncal$-PT-Dir/Gam, SVD for Hscore and LogME, and L-BFGS for LINEAR and PT-Gauss) but are still 1-2 orders of magnitude faster to compute than feature extraction from the penultimate layer.

$\Ncal$-LEEP, which obtains the source class assignments by applying GMM on the penultimate layer outputs, performs much better than LEEP on average. In addition, PACTran-Dirichlet and PACTran-Gamma with the GMM assignments (denoted as $\Ncal$-PT-Dir and $\Ncal$-PT-Gam) perform similarly to the $\Ncal$-LEEP algorithm, which indicates that the EEP estimator is surprisingly close to Bayesian optimal based on the GMM assignments of the VTAB tasks. 

Among the three algorithms that use the L-BFGS optimizer, LINEAR-VALID performs better than LINEAR for large $K$. However, for small $K$ LINEAR-VALID becomes worse, probably because the training and validation splits are too small.
%, which makes the learned classifier overfit to the training split.
% and the validation set unreliable. 
In contrast, the PT-Gauss metrics are consistently among the best metrics across all settings, 
which provides clear evidence that the 3rd ``flatness" term (Eq.\eqref{eq:pac-gauss-2}) plays a crucial role in predicting generalization error. 
For example, they are the only metrics with correlation 0.4 or higher on 2-9 classes.

In comparing between PT-Gauss$_{grid}$ and PT-Gauss$_{fix}$, we find that PT-Gauss$_{grid}$ usually performs well whenever LINEAR-VALID's does (since it depends on it for hyperparameter selection). 
On the other hand, when LINEAR-VALID is worse ($K$ is small), PT-Gauss$_{fix}$ outperforms PT-Gauss$_{grid}$.

 \begin{table}[!ht]
\centering
\vspace{-0.1in}
  \begin{tabular}{c | c | c | c }
GFLOPS & 100+ classes & 10-99 classes & 2-9 classes  \\\hline
LEEP & 6.40E-1 & 2.00E-2 & 1.55E-3\\
$\Ncal$-LEEP & 2.09E2 &1.40E0 &7.25E-2  \\
Hscore & 1.33E2 &1.30E2 &1.29E2 \\
LogME & 1.34E2 & 1.30E2 & 1.29E2 \\
LINEAR & 2.89E2 & 9.03E0 & 7.07E-1 \\
LINEAR-VALID & 9.64E1 & 3.02E0  & 2.36E-1 \\ \hline
$\Ncal$-PT-Dir &2.09E2 &1.40E0 &7.25E-2  \\
$\Ncal$-PT-Gam &2.09E2 &1.40E0 &7.25E-2 \\
PT-Gauss$_{grid}$ & 2.90E2 & 9.07E0 & 7.10E-1 \\
PT-Gauss$_{fix}$ & 6.45E1 & 2.02E0 & 1.58E-1 \\ \hline
Penultimate Feature &3.88E3 & 6.84E2 & 1.90E2
  \end{tabular}
  \caption{GFLOPS of running each metrics and the penultimate-layer feature extraction stage on the subsampled datasets, when $N/K=10$. }
  \label{tab:vtab10-gflops}
\vspace{-0.3in}
 \end{table}
 
\subsection{Visual Question Answering}
We further conduct experiments over the multi-modal VQA task.
Following common practice (\cite{vqa1,vqa2}), we treat VQA as a classification task (vocab-based VQA). That is, we construct a vocabulary based on the top answers in the training sets and classify into one of those labels.

\vspace{-0.1in}
\subsubsection{Pretrained Checkpoints}
We apply the state-of-art VQA model architecture, which fuses image and question representations in a multimodal Transformer model~\cite{transformers} (see C.1). %\ref{sec:vqa_arch}
We pretrain the VQA models over 9 different datasets, including: VQA-v2~\cite{vqa2}, GQA~\cite{gqa}, V7W~\cite{visual7w}, CNETVQA, TP-COLOR-COCO, TP-COLOR-CC3M, TP-COLOR-CC12M, VQ2A-COCO, VQ2A-CC3M~\cite{vq2a}. 
The detailed descriptions of the datasets are provided in C.2. %\ref{sec:vqa-pretrain-datasets}

For each pretraining dataset, we consider 3 different model sizes and 4 different finetuning hyperparameter settings.
For model sizes, the number of layers $t$ of the text-encoder and $m$ of the multimodal-encoder is varied from $(t, m) \in \cbr{(6, 3), (9, 5), (12, 7)}$. 
For hyperparameters, dropout ratios are varied from $\cbr{0, 0.1}$. We use two learning schedules: a constant learning rate of 0.0005 and a decay learning rate starting at 0.2. 
For each of these 12 settings, we set batch size to 128, and save a checkpoint after 100,000 iterations.
\vspace{-0.1in}
\subsubsection{Downstream Task}
We chose the OKVQA dataset~\cite{okvqa} because the task requires additional knowledge beyond its own training set, and it has been shown that proper pretraining brings significant benefits to performance~\cite{okvqa,vq2a}. 

\subsubsection{Experimental Settings}
Finetuning details are available in C.3. %\ref{sec:finetune-vqa} 
The hyperparameter settings match the NeuCRaB experiments.
Otherwise, we vary the number of data examples for metric computation from $N \in \cbr{40, 100, 200}$ and restrict the examples from the top 20 answers such that $N/K \in \cbr{2, 5, 10}$.
For each $N$, we create 5 subsamples of the OKVQA train set.
Each metric is then evaluated on the 5 splits and the average correlation score is reported.

\vspace{-0.1in}
\subsubsection{Results and Analysis}
In total, there are $108 = 9 \times 12$ checkpoints that span 9 different pretraining datasets, and 12 different model configurations. 
In Table \ref{tab:vqa}, we report their results in 3 different ways: (1) ``CD'' (cross pretraining data sources), reports the averaged correlation of metrics across the 9 different pretraining datasets for each of the 12 model configurations;
(2) ``CM'' (cross models), reports the averaged correlation of metrics cross the 12 model configurations for each of the 9 pretraining datasets; and (3) ``Total'', reports the correlation over all 108 checkpoints. 

As can be seen, when the pretraining tasks are classification based, LEEP performs much better compared to the ``mixed supervision'' tasks in the previous section.
On the other hand, PACTran-Gamma outperforms LEEP and PACTran-Dirichlet, which indicates that an unnormalized weight transfer matrix is more helpful in these setting.
LINEAR-VALID is a strong baseline, especially as more data examples are provided. 
Finally, we see that PACTran-Gauss (with $\beta = 0.1 N$ and $\sigma_0^2 = \frac{1}{D}$ from the grid search) provides competitive performance in all cases, and is consistently among the best in evaluating transferability from different pretraining datasets (``CD'').

\begin{table}[!t]
\centering
  \begin{tabular}{c | c  c  c | c c c | c c c}
N & & 40 & & & 100 & & & 200 & \\ \hline 
 & CD & CM & Total & CD & CM & Total & CD & CM & Total \\\hline
LEEP & 0.420 & 0.337 &0.471 & 0.430 & 0.373 &0.492& 0.435 & 0.402 &0.508  \\
$\Ncal$-LEEP & 0.309 & 0.077 &0.295 & 0.452 & 0.232 &0.427 & 0.503 & 0.329 &0.480\\
Hscore & 0.220 & 0.048 &0.198 & 0.253 & 0.079 &0.222 & 0.233 & 0.116 &0.243 \\
LogME & 0.350 & 0.141 &0.402 & 0.343 & 0.154 &0.395 & 0.357 & 0.160 &0.397\\
LINEAR & 0.355 & 0.137 &0.410 & 0.351 & 0.167 &0.407& 0.382 & 0.209 &0.423  \\
LINEAR-VALID & {\bf 0.488} &0.118 &0.430 & 0.526 &0.172 &0.474  & 0.579 &0.360 &0.528 \\ \hline
PT-Dir & 0.253 &0.329 &0.301& 0.449 &{\bf 0.418} &0.480 & 0.460 &{\bf 0.469} &0.503 \\
PT-Gam & 0.453 & {\bf 0.348} & {\bf 0.490} & 0.518 &{\bf 0.411} &{\bf 0.544} & 0.522 &0.430 &0.532 \\
$\Ncal$-PT-Dir & 0.424 &0.093 &0.358  & 0.522 &0.277 &0.476 & 0.548 &0.335 &0.504  \\
$\Ncal$-PT-Gam & 0.421 &0.092 &0.353 & 0.524 &0.278 &0.474  & 0.547 &0.333 &0.504\\
PT-Gauss$_{grid}$  & {\bf 0.480} &0.272 &0.451& {\bf 0.566} &0.349 &{\bf 0.544}& {\bf 0.617} &0.391 &{\bf 0.582} \\
%PT-Gauss$_{fix}$  & {\bf 0.483} &0.266 &0.453 & {\bf 0.564} &0.351 &{\bf 0.537} & {\bf 0.616} &0.386 &{\bf 0.579} \\
  \end{tabular}
  \caption{Kendall-Tau correlations on the OKVQA experiments with different $N$.}
  \label{tab:vqa}
  \vspace{-0.3in}
 \end{table}
 
\section{Conclusion}
In this paper we presented PACTran, a PAC-Bayesian based framework for measuring the transferability of pretrained checkpoints to downstream tasks. 
%Three variant PACTran metrics were studied using different hypothesis spaces and priors. 
%We conducted experiments over a set of vision tasks (VTAB) and a vision-and-language task (OKVQA).
Our method significantly improves upon previous methods in that it is both theoretically sound as well as compatible with downstream classification tasks.
We instantiated three variant PACTran metrics using different hypothesis spaces and priors and conducted experiments over a set of vision tasks (VTAB) and a vision-and-language task (OKVQA).
We showed that some PACTran variants can provide theoretical justification for existing methods. For example, ($\Ncal$-)PT-Dir and ($\Ncal$-)PT-Gam metrics subsume ($\Ncal$-)LEEP, in which the finetuning head sits on top of the pretrained classification head (or a GMM).
Our experiments also showed that several of the baseline metrics are unable to measure checkpoint transferability better than a simple linear classification and validation baseline (LINEAR-VALID). 
On the other hand, the proposed PT-Gauss metric behaved well as a measure of transferability in a limited data setting and consistently exhibited high correlation with the test performance of models finetuned on the downstream tasks.
Possibly, this is because it more closely matches the setup of the finetuned model, where the finetuning head is placed directly on the penultimate layer and trained with a cross-entropy loss.

%has the same hypothesis space as practical finetuning procedures by putting the finetuning head directly on the penultimate layer and training with a cross-entropy loss, 
%Specifically, the PACTran-Gauss metric, which puts the finetuning head directly on the penultimate layer (as commonly done in practice) often exhibits the best performance.

% We demonstrated the versatility of the PACTran framework in different hypothesis spaces by describing three types of PACTran metrics. Using the PACTran framework, existing metrics can be transformed so as to have better theoretical justification. For example, the ($\Ncal$-)PT-Dir and ($\Ncal$-)PT-Gam metrics subsume ($\Ncal$-)LEEP, in which the finetuning head sits on top of the pretrained classification head (or a GMM).
% On the other hand, the common practice is to put the finetuning head directly on the penultimate layer and train with a cross-entropy loss. 
% The PACTran-Gauss setup exactly matches this widely used routine (i.e. having the same hypothesis space), and often leads to the best performance in practice. 
% Therefore, we recommend PACTran-Gauss for practical use.

\clearpage

\bibliographystyle{splncs04}
\bibliography{main}

\newpage
\appendix
\section{Additional Derivations and Discussions Regarding the PACTran Metrics}
\subsection{PACTran-Dirichlet}
\label{sec:derive-pacdir}
\subsubsection{Variational Inference Derivations}
In variational inference, we make use of a set of independent distributions, including multinomial distributions $q(z_i)$ and Dirichlet distributions $q(\wb_z; \tilde{\alphab}_z)$ and apply Jensen's inequality~\cite{blei2017variational} to Eq.\eqref{eq:pac-lda} such that,
{\small
\begin{align}
    \log Z(S) 
    &\ge H_q(\zb) + H_q(\Wb) + \sum_{z_1} \ldots \sum_{z_N} q(\zb) \int d\Wb q(\Wb) \log p(\yb, \zb, \Wb | \xb) \nonumber\\
    &= H_q(\zb) + H_q(\Wb) + \sum_{z_1} \ldots \sum_{z_N} q(\zb) \int d\Wb q(\Wb) \nonumber\\
    &\rbr{\sum_z \log \frac{\Gamma(\sum_y \alpha_y)}{\prod_y \Gamma(\alpha_y)} + \sum_z\sum_y(\alpha_y - 1)\log w_{yz} + \sum_i\sum_z \delta_{z_i=z} \log (M(\xb_i)_z w_{y_i z})}. \label{eq:elbo}
\end{align}
}

The variational inference seeks the optimal approximate distributions $q(z_i)$ and $q(\wb_z; \tilde{\alphab}_z)$ that maximize Eq.~\eqref{eq:elbo}.
%\begin{align*}
%     \sum_{\zb} q(\zb) \int \rbr{q(\Wb) \log p(\yb, \zb, \Wb | \xb) - q(\wb) \log q(\wb)} d \Wb - \sum_{\zb} q(\zb) \log q(\zb).
%\end{align*} 
Taking the functional derivative w.r.t. $q(\zb)$ and making it equal to 0, one gets
\begin{align*}
    \log q^*(z_i=z) =& \int q(\Wb) \log p(\yb, z_i=z, \Wb | \xb) d \Wb + C \\
    =& \log M(\xb_i)_z + \EE_{q(\Wb)} \log w_{y_iz} + C,
\end{align*}
where $C$ is a constant. Since $q(\Wb)$ are Dirichlet distributions, we have
\begin{align*}
    \EE_{q(\Wb)} \log w_{y_iz} = \Psi(\tilde{\alpha}_{y_i z}) - \Psi(\sum_y \tilde{\alpha}_{y z}).
\end{align*}

Next, taking the functional derivative w.r.t. $q(\Wb)$ and making it equal to 0, one gets
\begin{align*}
    \log q^*(w_{yz}) =& \sum_{\zb} q(\zb) \log p(\yb, \zb, w_z^y | \xb) + C\\
    =& \rbr{\alpha_y - 1 + \sum_i q(z_i=z) \delta(y_i=y)} \log w_{yz} + C,
\end{align*}
where C is a constant. Since $\log q^*(w_{yz}) = (\tilde{\alpha}_{yz} -1)\log w_{yz} + C$, we have
\begin{align*}
    \tilde{\alpha}_{yz} = \alpha_y + \sum_i q(z_i=z) \delta(y_i=y).
\end{align*}

\subsection{PACTran-Gamma}
\label{sec:derive-pacgam}
\subsubsection{Marginal Evidence}
Since the denominator of Eq.\eqref{eq:pac-gamma-pyx} creates difficulties for Bayesian inference, we introduce a set of augmented variables $R_i$ from the exponential distribution as in~\cite{archambeau2012plackett} to "cancel out" the denominator, such that
\begin{align}
p(y_i, z, R_i | \xb_i, \Vb) = M(\xb_i)_z v_{y_i z}\exp\rbr{-R_i \rbr{\sum_{y \in \Ycal} \sum_{z \in \Zcal} M(\xb_i)_z v_{yz}}},  \label{eq:pac-gamma-pyx-aug}
\end{align}
It is easy to verify that $\int_0^{+\infty} p(y_i, z, R_i | \xb_i, \Vb) dR_i = p(y_i, z | \xb_i, \Vb)$. 
Therefore, the marginal evidence $\log Z(S)$ becomes,
\begin{align*}
&\log \int \prod_y \prod_z P(v_{yz}) \prod_i\rbr{\sum_z p(y_i, z| \xb_i, \Vb)} d \Vb \nonumber\\
=&\log \int \prod_y \prod_z P(v_{yz}) \prod_i\rbr{\sum_z p(y_i, z, R_i | \xb_i, \Vb)} dR_i d \Vb \nonumber\\
=&\log \int \prod_y \prod_z \frac{b^{a_y}}{\Gamma(a_y)}v_{yz}^{a_y-1} \exp(-bv_{yz}) \prod_i \nonumber\\
&\quad \rbr{\sum_z M(\xb_i)_z v_{y_i z}\exp\rbr{-R_i \rbr{\sum_{y \in \Ycal} \sum_{z \in \Zcal} M(\xb_i)_z v_{yz}}}} dR_i d \Vb \nonumber\\
=&\log \sum_{z_1} \ldots \sum_{z_N} \int \prod_y \prod_z \frac{b^{a_y}}{\Gamma(a_y)}v_{yz}^{a_y-1} \exp(-bv_{yz}) \prod_i \nonumber\\
&\quad \rbr{M(\xb_i)_{z_i} v_{y_i z_i}\exp\rbr{-R_i \rbr{\sum_{y \in \Ycal} \sum_{z \in \Zcal} M(\xb_i)_z v_{yz}}}} dR_i d \Vb .
\end{align*}
Since the exact inference is infeasible, we again apply variational inference. 

\subsubsection{Variational Inference Derivations}
Similar to the PACTran-Dirichlet, we make use of a set of independent distributions $q(z_i)$, $q(v_{yz}; \tilde{a}_{yz}, b)$ and $q(R_i; \tilde{\lambda}_i)$ and write
{\small
\begin{align}
    &\log Z(S) \nonumber\\
    \ge& H_q(\zb) + H_q(\Vb) + H_q(\Rb) + \sum_{z_1} \ldots \sum_{z_N} q(\zb) \int d\Vb d \Rb q(\Vb) q(\Rb) \nonumber\\
    &+\sum_y \sum_z a_y \log b -\sum_y \sum_z \log \Gamma(a_y) + \sum_z\sum_y ((a_y - 1) \log v_{yz} - b v_{yz})+  \nonumber\\
    &\sum_i\sum_z \delta_{z_i=z} \log (M(\xb_i)_z v_{y_i z})-\sum_i R_i \rbr{\sum_{y \in \Ycal} \sum_{z \in \Zcal} M(\xb_i)_z v_{yz}}. \label{eq:elbo-gamma}
\end{align}
}
The PACTran-Gamma metric is the resulting negative ELBO after applying variational principles, and takes the following form (when $b=1$):
\begin{align}
& \sum_y \sum_z \rbr{\log \Gamma(a_y) - \log \Gamma(\tilde{a}_{yz})} + \sum_i \log \tilde{\lambda}_i \nonumber\\
+& \sum_i \sum_z q^*(z_i=z)\rbr{\log q^*(z_i=z)- \log M(x_i)_z}, 
\end{align}
where,
\begin{align*}
    &q^*(z_i=z) = \text{softmax}\rbr{\log M(\xb_i)_z + \Psi(\tilde{a}_{y_i z})},\\
    &\tilde{a}_{yz} = a_y + \sum_i q^*(z_i=z)\delta(y_i=y),\;\;
    \tilde{\lambda}_i = \sum_y \sum_z M(\xb_i)_z \tilde{a}_{yz} .
\end{align*}
The above equations are obtained in a similar way as the ones of the PACTran-Dirichlet metric in \ref{sec:derive-pacdir}. 
%Since the resulting PACTran-Gamma metric is an optimal ELBO, it is a valid PAC-Bayesian bound to the transfer error for the same reason as the PACTran-Dirichlet metric.

% updates b as well, appears to be unstable.
%\begin{align}
%& \sum_y \sum_z \rbr{\log \Gamma(a_y) - \log \Gamma(\tilde{a}_{yz}) + \tilde{a}_{yz} \log \tilde{b}_{yz} - a_y \log b} \nonumber\\
%+& \sum_i \sum_z q^*(z_i=z)\rbr{\log q^*(z_i=z)- \log M(x_i)_z} + \sum_i(\log \tilde{\lambda}_i - 1), \label{eq:pac-gamma-v}
%\end{align}
%where,
%\begin{align*}
%    &q^*(z_i=z) = \text{softmax}\rbr{\log M(\xb_i)_z + \Psi(\tilde{a}_{y_i z}) - \log(\tilde{b}_{y_i z})},\\
%    &\tilde{a}_{yz} = a_y + \sum_i q^*(z_i=z)\delta(y_i=y),\;\; 
%    \tilde{b}_{yz} = b + \sum_i M(\xb_i)_z / \tilde{\lambda}_i, \\
%    &\tilde{\lambda}_i = \sum_y \sum_z M(\xb_i)_z \tilde{a}_{yz} / \tilde{b}_{yz}.
%\end{align*}

\subsection{PACTran-Gaussian}
\label{sec:derive-pacgau}

\subsubsection{The optimal Gaussian Posterior}
To obtain the optimal parameters $\sigma_*^2$ and $\thetab_*$ of the Gaussian posterior, first take the derivative of Eq.\eqref{eq:pac-gauss-1} w.r.t. $\sigma_q^2$ and make it zero, 
\begin{align*}
    \text{Tr}(\nabla^2\hat{L}(\thetab_q, S)) + \frac{KD}{\lambda}\rbr{\frac{1}{\sigma_0^2} - \frac{1}{\sigma_q^2} } = 0.
\end{align*}
After rearrangement, one gets
\begin{align*}
    \frac{\sigma_0^2}{\sigma_q^2} = 1 + \frac{\beta}{KD}\text{Tr}(\nabla^2 \hat{L}(\thetab_q, S)),
\end{align*}
where $\beta=\lambda \sigma_0^2$. Now plugging this into Eq.\eqref{eq:pac-gauss-1} yields
\begin{align}
    \hat{L}(\thetab_q, S)+ \frac{\|\thetab_q\|_F^2}{2\beta} + \frac{KD\sigma_0^2}{\beta} \log \frac{\sigma_0^2}{\sigma_q^2}, \label{eq:pac-gauss-2'}
\end{align}
and the optimal $\thetab_*$ is the one which minimizes the above objective function. 
Strictly speaking, the objective function with respect to $\thetab_q$ should consider the last term of Eq.\eqref{eq:pac-gauss-2'}. However, this would make the objective non-convex and hard to optimize. Therefore, we approximate the solution by ignoring the last term and only optimize $\thetab_q$ over the first two terms, 
which is a strongly convex objective and can be solved efficiently with an off-the-shelf optimizer (e.g. L-BFGS). 

\subsubsection{2nd-Order Derivative of the Cross-Entropy Loss.}
\label{sec:2nd-derivative}
For a given dataset $S$, assuming $\Xb$ is the feature matrix of size $N \times D$, where $N$ is the number of examples and $D$ is the feature dimension. $\Yb$ is a binary matrix of size $N \times K$ representing the labels, where $K$ is the number of classes. Then the logits can be represented as
\begin{align*}
    \Gb = \Xb \Wb + \bb,
\end{align*}
where $\Wb$ is $D \times K$ and $\bb$ is a $K$-dim bias vector. For $\thetab_q = (\Wb, \bb)$, 
its cross-entropy loss on the dataset $S$ is
\begin{align*}
    \hat{L}(\thetab_q, S) = \frac{1}{N}\sum_i \rbr{-\sum_k y_{ik} g_{ik} + \log \sum_k \exp(g_{ik})}.
\end{align*}
Its first derivative w.r.t. $w_{jk}$ is
\begin{align*}
\frac{\partial \hat{L}}{\partial w_{jk}} = \frac{1}{N}\sum_i x_{ij} \frac{\partial \hat{L}}{\partial g_{ik}} = \frac{1}{N}\sum_i x_{ij} \rbr{\text{softmax}(g_{ik}) - y_{ik}},
\end{align*}
and the second derivative w.r.t. $w_{jk}$ is 
\begin{align}
\frac{\partial^2 \hat{L}}{\partial w_{jk}^2} &= \frac{1}{N}\sum_i x_{ij}^2 \rbr{\text{softmax}(g_{ik}) - \text{softmax}(g_{ik})^2}. \label{eq:2nd_w}
\end{align}
Similarly, its first derivative w.r.t. $b_{k}$ is
\begin{align*}
\frac{\partial \hat{L}}{\partial b_{k}} = \frac{1}{N}\sum_i \rbr{\text{softmax}(g_{ik}) - y_{ik}},
\end{align*}
and the second derivative is
\begin{align}
\frac{\partial^2 \hat{L}}{\partial b_{k}^2} &= \frac{1}{N}\sum_i \rbr{\text{softmax}(g_{ik}) - \text{softmax}(g_{ik})^2}. \label{eq:2nd_b}
\end{align}
Given Eq.\eqref{eq:2nd_w} and Eq.\eqref{eq:2nd_b}, the trace of the Hessian can be written as
\begin{align*}
    \text{Tr}(\nabla^2 \hat{L}(\thetab_q, S)) = \sum_{jk} \frac{\partial^2 \hat{L}}{\partial w_{jk}^2} + \sum_k \frac{\partial^2 \hat{L}}{\partial b_{k}^2},
\end{align*}
which is used as the "flatness regularizer" (in the 3rd term of Eq.\eqref{eq:pac-gauss-2'}).

\subsubsection{Differences between LogME and PACTran-Gaussian}
\label{sec:logme_pt_gauss}
Although both LogME~\cite{you2021logme} and PACTran-Gaussian apply the Gaussian priors on the top-layer parameters $\thetab$, they differ in the following two aspects: 
(1) LogME models the data distribution using the Gaussian likelihood, which corresponds to the squared loss in its logarithm form for optimization. 
On the other hand, PACTran applies the cross-entropy loss, which is more natural for classification tasks and is universally applied in practical downstream finetunings. 
(2) LogME optimizes all adjustable hyper-parameters along with the parameters $\thetab$ which results with a highly complex optimization problem. In contrast, PACTran-Gaussian only optimizes the parameters $\thetab$ which is convex, 
while heuristically setting the hyper-parameters $\beta$ and $\sigma_0$ separately (Section \ref{sec:hparams-pac-gauss}).

\subsection{Complexity of the PACTran Metrics}
Overall, the complexity of the PACTran metrics is $O(NKDt)$, where $t$ is either the number of variational inference updates, or the number of L-BFGS steps.
\subsubsection{PACTran-Dirichlet}
The PACTran-Dirichlet metric in Eq.~\eqref{eq:pac-lda-v} involves two sums, the sum over the $D$ source classes of $z$ and the sum of $N$ examples. Each $C(\alpha)$ involves $K$ classes of $y$. So the overall complexity is $O(ND + KD)$. To compute $q^*$ and $\tilde{\alpha}$, it involves $t \le 10$ iterations of variational updates. There are $ND$ of $q^*$ terms and the overall complexity is $O(ND + KD)$. There are $KD$ terms of $\tilde{\alpha}$ and the overall complexity is $O(NKD)$. Therefore, the overall complexity of PACTran-Dirichlet is $O(NKDt)$.

\subsubsection{PACTran-Gamma}
The PACTran-Gamma metric has similar complexity to the PACTran-Dirichlet. The complexity of Eq.~\eqref{eq:pac-gamma-v} is $O(ND + KD)$. There are $ND$ of $q^*$ terms and the overall complexity is $O(ND)$. There are $KD$ terms of $\tilde{\alpha}$ and the overall complexity is $O(NKD)$. There are $N$ terms of $\tilde{\lambda}$ and the overall complexity is $O(NKD)$. Therefore, the overall complexity of PACTran-Gamma is also $O(NKDt)$ where $t$ is the number of variational updates.

\subsubsection{PACTran-Gaussian}
The PACTran-Gaussian metric according to Eq.~\eqref{eq:pac-gauss-2} involves three terms. 
Evaluating the first two terms has complexity $O(NKD)$. 
The third term involves the 2nd-order derivative of the loss which has a closed form solution as shown in Section \ref{sec:2nd-derivative} and evaluation complexity $O(NKD)$\footnote{It is worth noting that our complexity is significantly lower than the one of the classical Laplacian approximation, which involves the determinant of a Hessian with $O(NK^3D^3)$ complexity.}. To obtain $\thetab_*$, we call L-BFGS which requires computing the derivative, which is also of complexity $O(NKD)$. Therefore, the overall complexity of PACTran-Gaussian is $O(NKDt)$, where $t$ is the number of L-BFGS function/derivative evaluations.

\section{Additional Details of the Neural Checkpoint Ranking Benchmark (NeuCRaB) Experiments}
\subsection{Pretraining Checkpoint Descriptions}
All checkpoints were based on the ResNet50-v2 architecture and pretrained on Imagenet using various approaches. 
\begin{enumerate}
    \item Jigsaw: trained with the self-supervised jigsaw-puzzle loss~\cite{jigsaw_noroozi2017}.
    \item Relative Patch Location: trained with the relative path location prediction self-supervised loss~\cite{relative_doersch2016}.
    \item Exemplar: trained with the Exemplar loss~\cite{exemplar}.
    \item Rotation: representation obtained by predicting image rotations~\cite{rotation}.
    \item Sup-Rotation: trained with a supervised loss and an auxiliary Rotation loss~\cite{rotation}.
    \item WAE-UKL: encoder obtained by training a Wasserstein Autoencoder using the RAM-MC method of~\cite{wae_ukl} as a distribution matching penalty that upper bounds the KL divergence (UKL stands for Upper-bound KL).
    \item WAE-GAN: encoder obtained by training a Wasserstein Auto-Encoder using GAN-based distribution matching loss~\cite{wae_gan_mmd}.
    \item WAE-MMD: encoder obtained by training a Wasserstein Auto-Encoder using the Maximum Mean Discrepancy (MMD) distribution matching loss~\cite{wae_gan_mmd}.
    \item Cond-BigGAN: representation obtained from the discriminator of a BigGAN trained for class-conditional image synthesis~\cite{biggan}.
    \item Uncond-BigGAN: representation obtained from the discriminator of an unconditional BigGAN model with auxiliary self-supervision.
    \item VAE: Encoder obtained by training a Variational Auto-Encoder~\cite{vae_2013}.
    \item Semi-Rotation-10\%: trained with a supervised loss on 10\% of the ImageNet examples and with an auxiliary Rotation loss~\cite{rotation} on all of the examples.
    \item Semi-Exemplar-10\%: trained with a supervised loss on 10\% of the ImageNet examples and with an auxiliary Exemplar loss~\cite{exemplar} on all of the examples.
    \item Sup-Exemplar-100\%: trained with a supervised loss and an auxiliary Exemplar loss~\cite{exemplar} on all of the examples.
    \item Sup-100\%: representation obtained by standard supervised training on ImageNet.
    \item Feature Vector: representation obtained by a ResNet50 model using the identity mappings as in~\cite{feature_vec} with supervised loss.
\end{enumerate}
\subsection{Downstreaming Task Descriptions}
In this section, we describe the VTAB downstream tasks used in Section~\ref{sec:neucrab}.
\begin{enumerate}
    \item \textbf{Caltech101}~\cite{caltech101_FeiFei2004} contains 101 classes, including animals, airplanes, chairs, etc. The image size varies from 200 to 300 pixels per edge. \item \textbf{Flowers102}~\cite{flowers102_Nilsback08} contains 102 classes, with 40 to 248 training images (at least 500 pixels) per class. 
    \item \textbf{Patch Camelyon}~\cite{patch_camelyon} contains 327,680 images of histopathologic scans of lymph node sections with image size of 96x96, which is collected to predict the presence of metastatic tissue. 
    \item \textbf{Sun397}~\cite{sun397_Xiao:2010} is a scenery benchmark with 397 classes, including cathedral, staircase, shelter, river, or archipelago. 
    \item \textbf{Cifar-10}~\cite{cifar10_Krizhevsky09} consists of 60,000 32x32 colour images in 10 classes, with 6,000 images per class. There are 50,000 training images and 10,000 test images. 
    \item \textbf{Oxford-IIIT Pet}~\cite{oxford_iiit_pet_parkhi12a} is a 37-class pet image dataset with roughly 200 images for each class. The images have large variations in scale, pose and lighting. All images have an associated ground truth annotation of breed. 
    \item \textbf{Smallnorb}~\cite{smallnorb_LeCun2004} is a dataset intended for experiments in 3D object recognition from shape. It contains images of 50 toys belonging to 5 generic categories: four-legged animals, human figures, airplanes, trucks, and cars. We used the azimuth angle as the label, which has 18 classes (0 to 340 every 20 degrees).
    \item \textbf{DMLAB}~\cite{zhai2019large} is a dataset for evaluating the ability of a visual model to reason about distances from the visual input in 3D environments. It has 100,000 360x480 color images in 6 classes. The classes are \{close, far, very far\} x \{positive reward, negative reward\} respectively.
    \item \textbf{CBIS-DDSM}~\cite{CBIS_DDSM_Citation} stands for Curated Breast Imaging Subset of Digital Database for Screening Mammography. It contains 65,130 patches with both calcification and mass cases, plus patches with no abnormalities. Designed as a traditional 5-class classification task.
\end{enumerate}

\subsection{Finetuning}
\label{sec:finetune}
Each pretrained checkpoint was finetuned on each downstream task in the following two ways: 
8 attempts were made by full-model finetuning of the checkpoints with batch size 512, weight-decay 0.0001, with an SGD-Momentum optimizer using a decaying learning schedule with different starting learning rate $lr$ and stopping iterations $iter$: $lr \in \cbr{0.1, 0.05, 0.01, 0.005}$ and $iter \in \cbr{10000, 5000}$; 
5 attempts were done with top-layer-only finetuning using an L-BFGS solver with weight decay $\frac{1}{B} \cdot \cbr{0.01, 0.1, 1., 10., 100.}$, where $B$ is the size of the training set. The ground-truth testing error was set to the lowest test error among all runs.

\subsection{Computation Platform}
The pretrained feature extraction and the pretrained model finetuning were done on the Google Cloud V1 2x2 TPUs. 
The transferability metrics were computed on the Google Cloud Intel Skylake CPU (2GHz per core) with 1 core and 10GB RAM per run.

\subsection{Hyperparameter Studies of the PACTran-Gaussian Metric}
\label{sec:hparams-pac-gauss}

Recall that in Eq.~\eqref{eq:pac-gauss-2} we decomposed the PACTran-Gaussian metric into two parts: the $l_2$-regularized empirical risk (RER) and the "flatness regularizer" (FR). There are two hyperparameters in the PACTran-Gaussian metrics: $\beta$ and $\sigma_0^2$. The $\beta$ hyperparameter is mainly responsible of adjusting the $l_2$ regularizer so that the magnitude of $\thetab_*$ would not get too large. The $\sigma_0^2$ hyperparameter is mainly responsible of balancing the weights between RER and FR. 
\begin{align*}
    \underbrace{\hat{L}(\thetab_*, S) + 
    \frac{\|\thetab_*\|_F^2}{2 \beta}}_{RER} + 
    \underbrace{\frac{KD \sigma_0^2}{2 \beta} \log \frac{\sigma_0^2}{\sigma_*^2}}_{FR}.  
\end{align*}

In the experiment, we performed a hyperparameter grid-search over $\beta \in a \cdot N$ and $\sigma_0^2 \in b \cdot \frac{1}{D}$, for various choices of $a$ and $b$. In particular, $a \in \cbr{0.1, 1, 10}$, and $b \in \cbr{1, 10, 100, 1000}$. The hyperparameter $(\beta, \sigma_0^2)$ that maximizes the Kendall correlation between the PT-Gauss metric and LINEAR-VALID was chosen for the PT-Gauss$_{grid}$ metric.

\begin{figure}[!ht]
    \centering
    \includegraphics[width=0.45\textwidth]{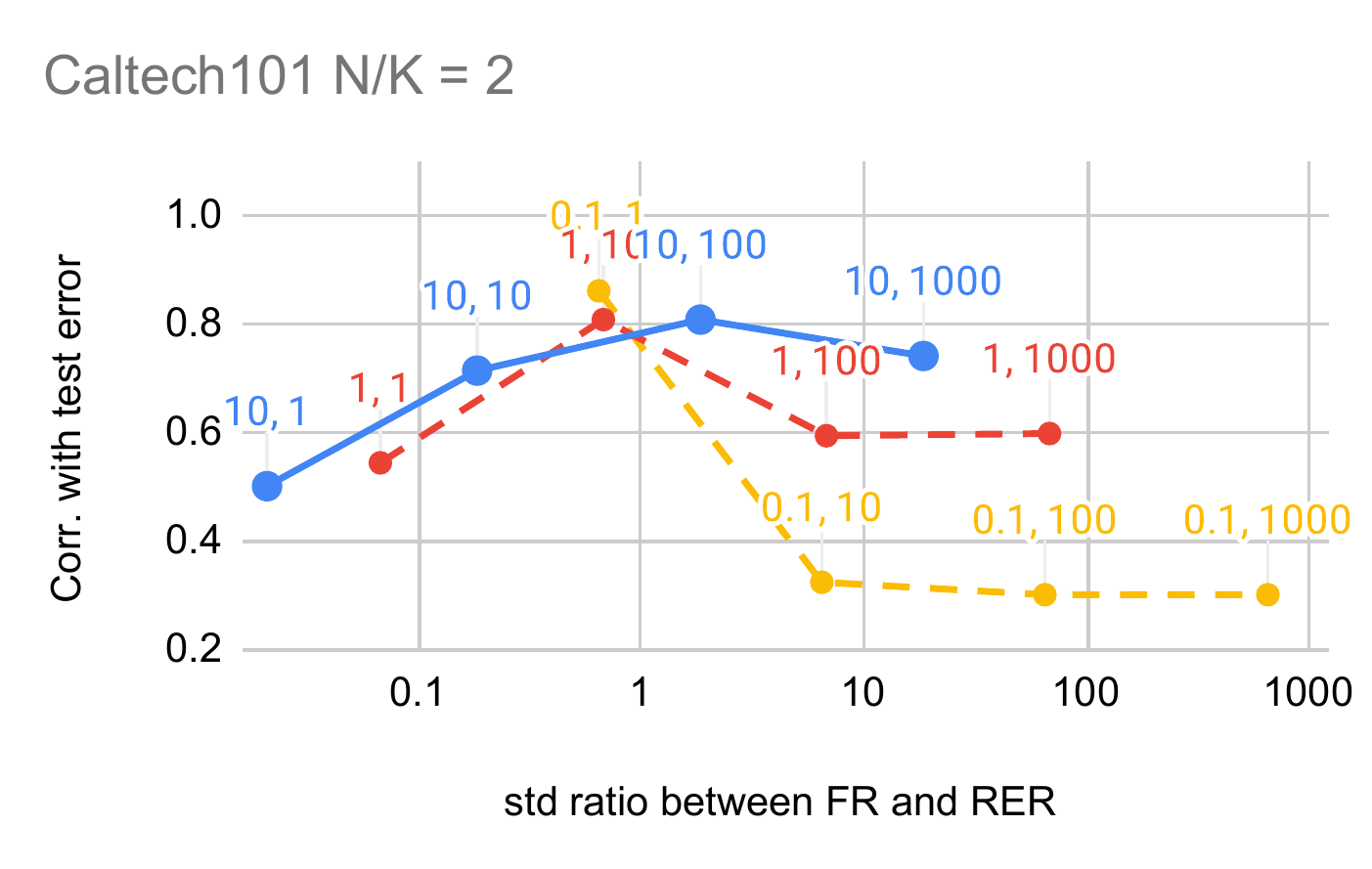}
    \includegraphics[width=0.45\textwidth]{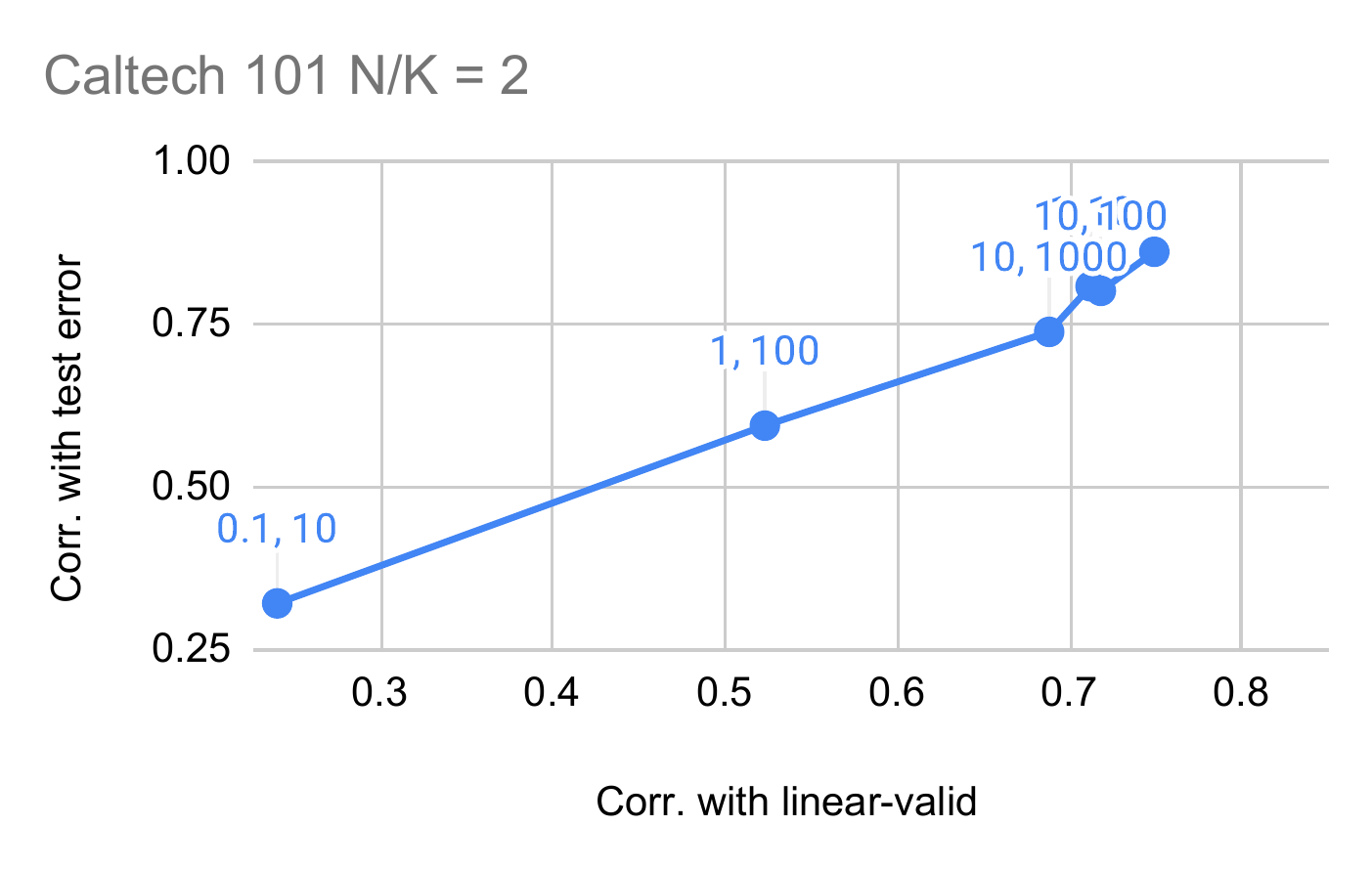}
    \includegraphics[width=0.45\textwidth]{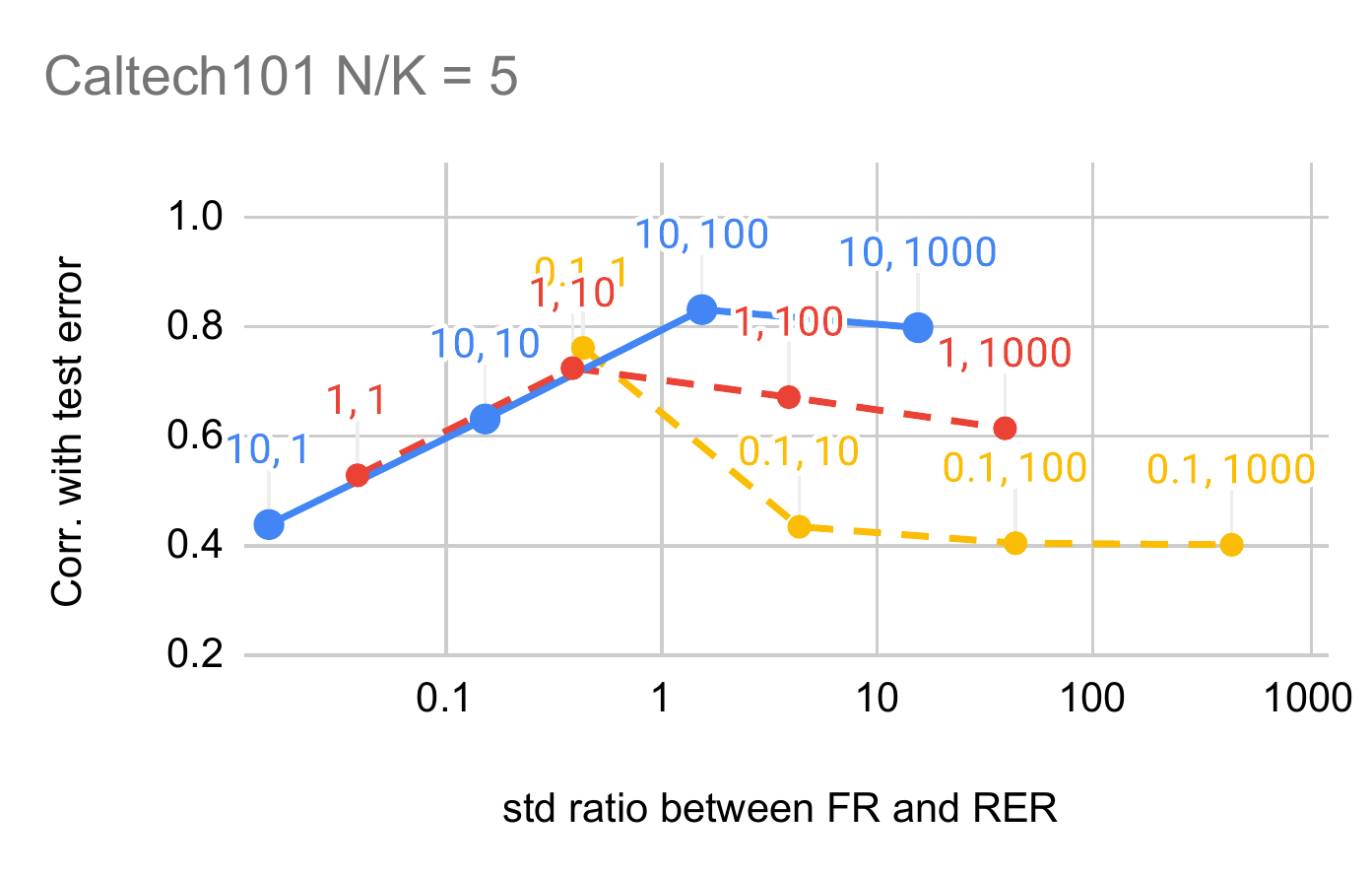}
    \includegraphics[width=0.45\textwidth]{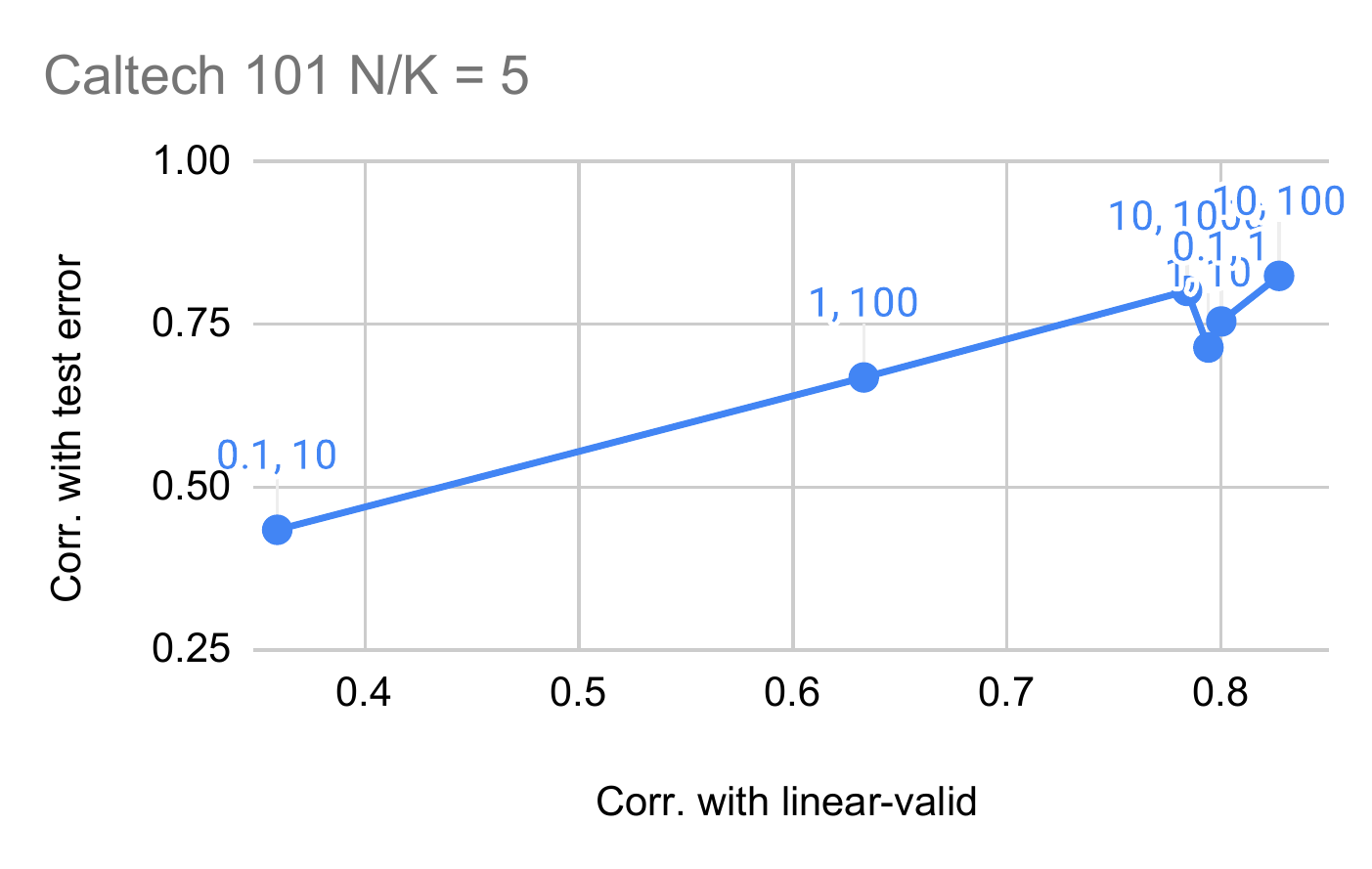}
    \includegraphics[width=0.45\textwidth]{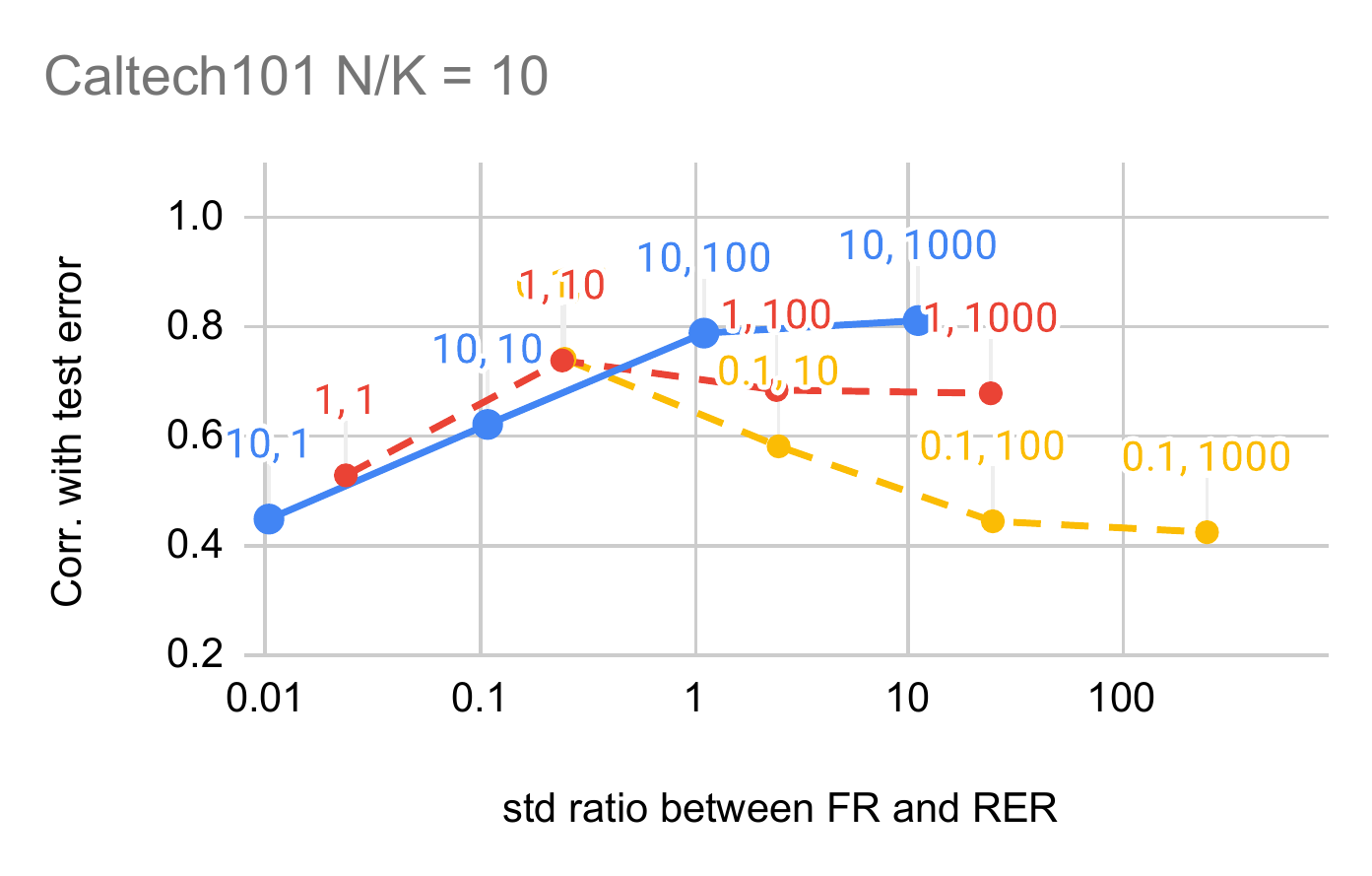}
    \includegraphics[width=0.45\textwidth]{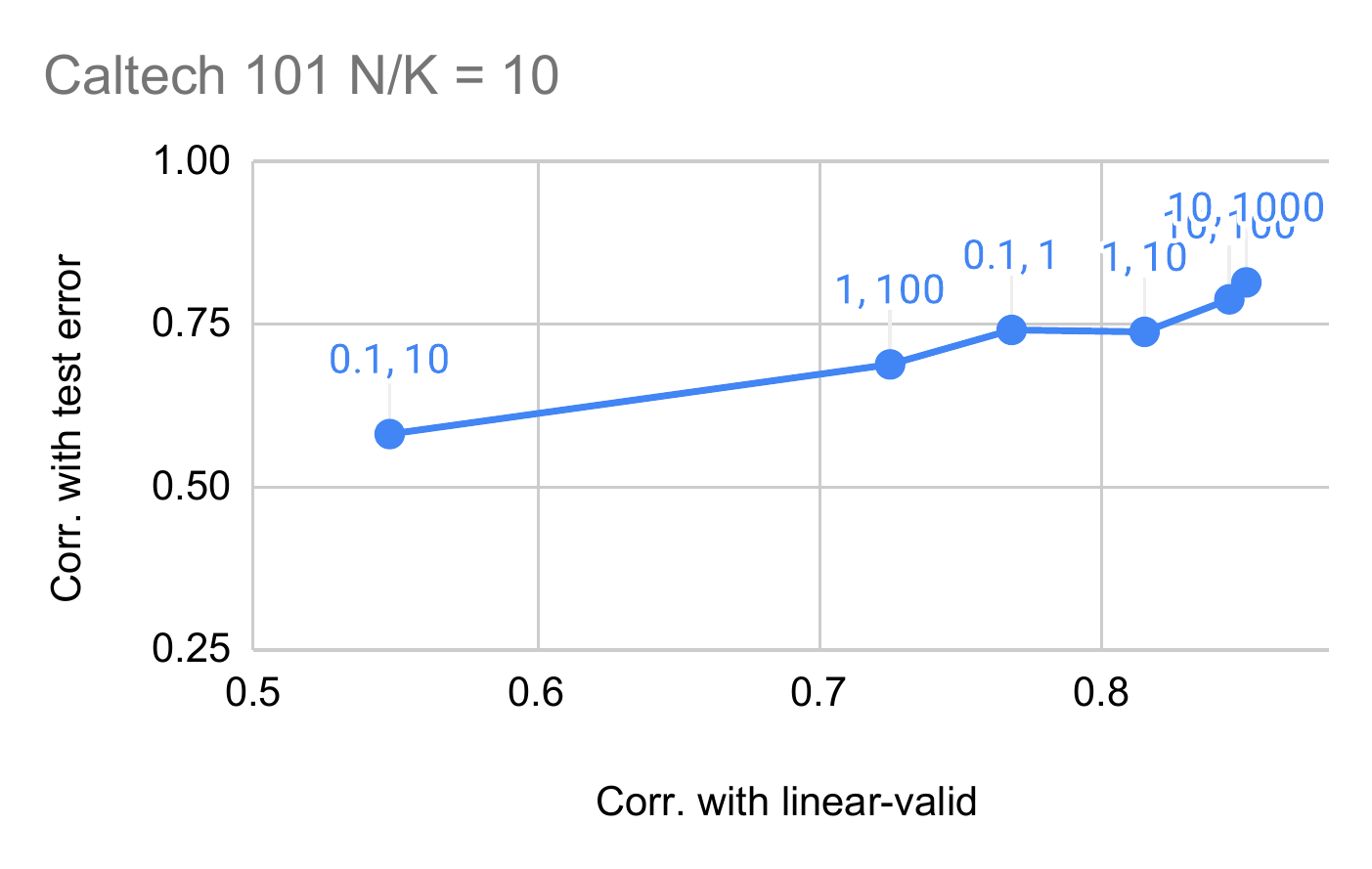}
    \caption{PACTran-Gaussian hyperparameter studies on Caltech101. Hyperparameters are labeled as $(a, b)$. High $y$-value indicates a good correlation with the downstream test error.}
    \label{fig:hp_caltech101}
\end{figure}

\begin{figure}[!ht]
    \centering
    \includegraphics[width=0.45\textwidth]{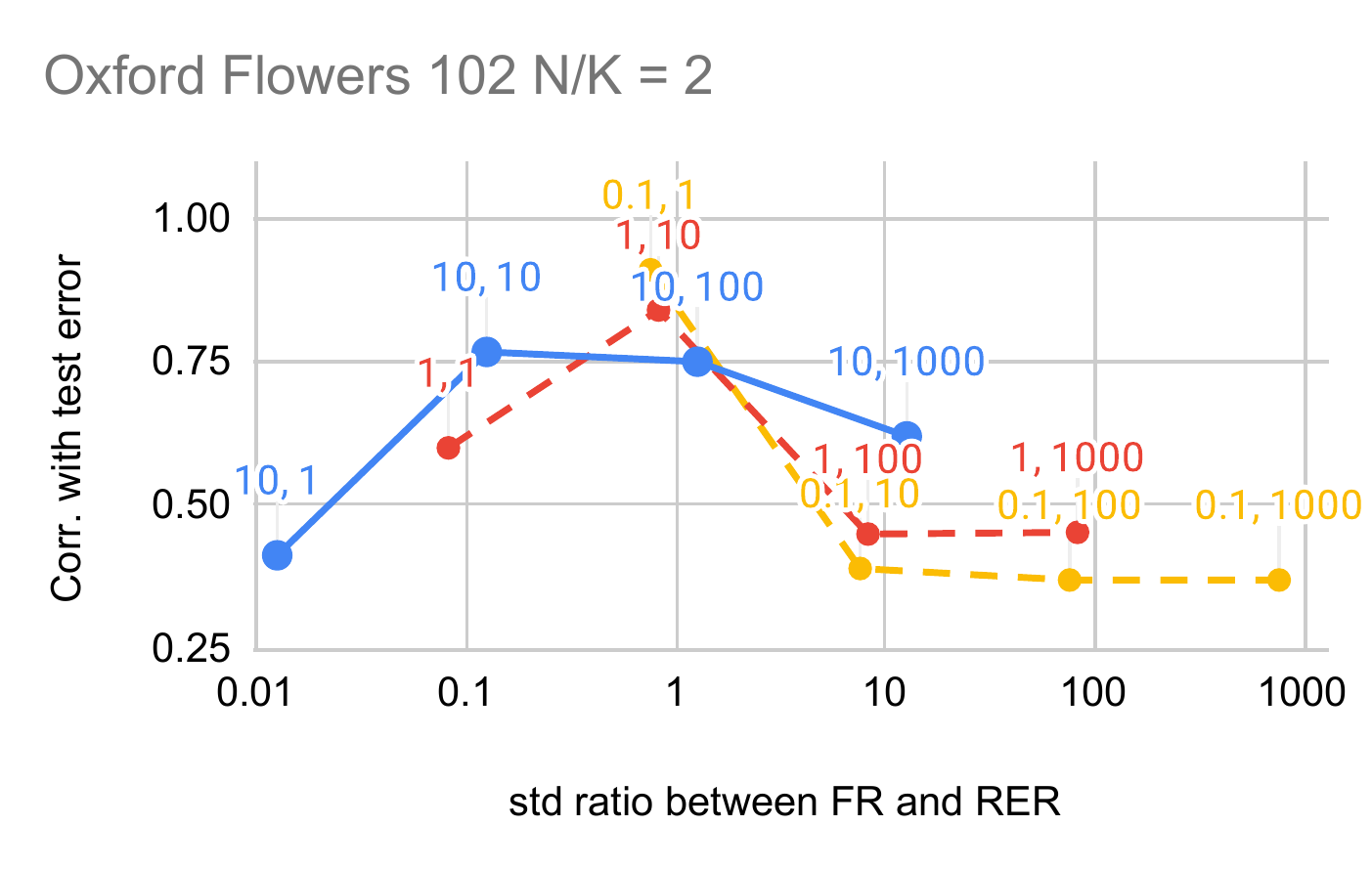}
    \includegraphics[width=0.45\textwidth]{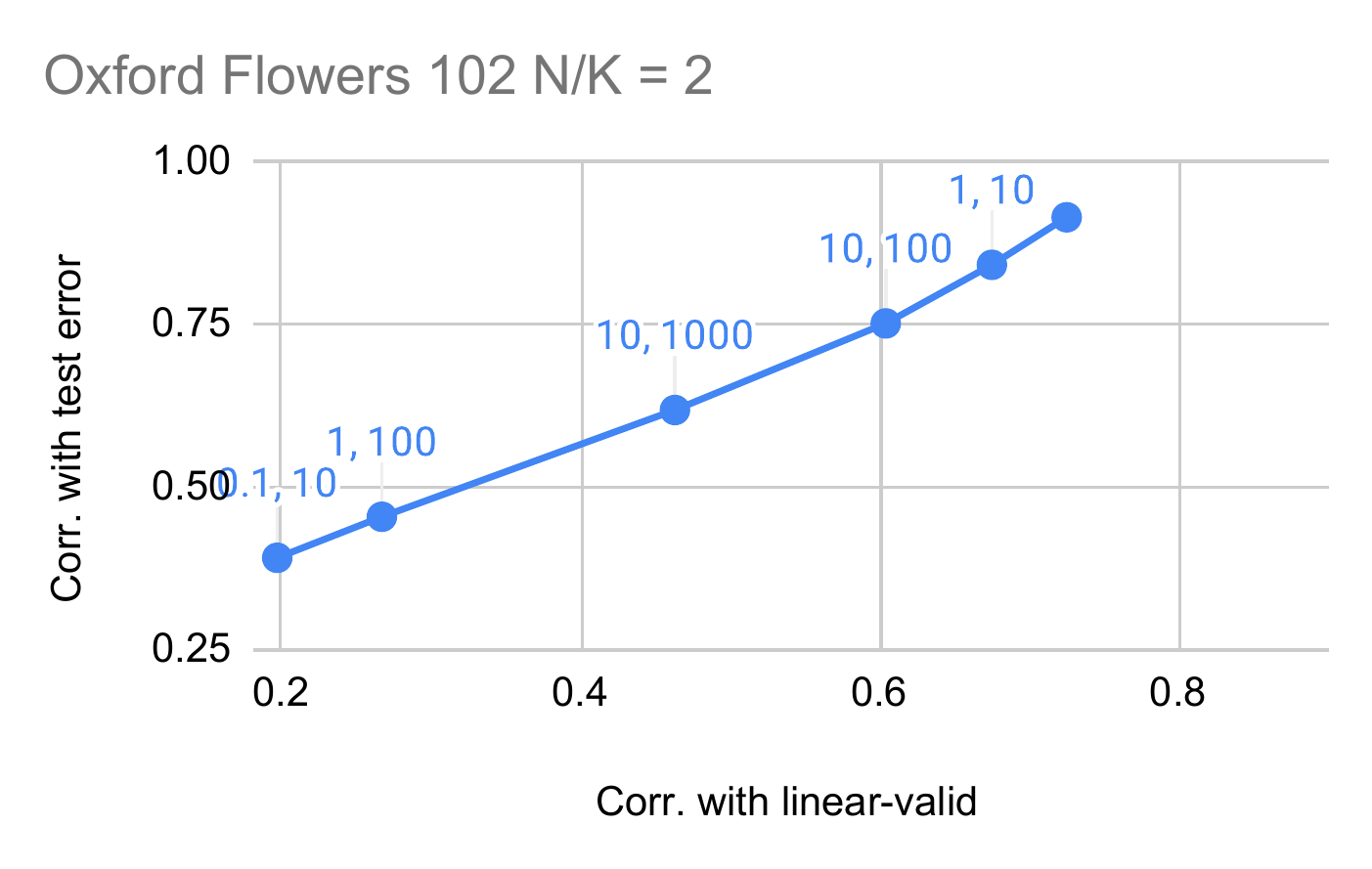}
    \includegraphics[width=0.45\textwidth]{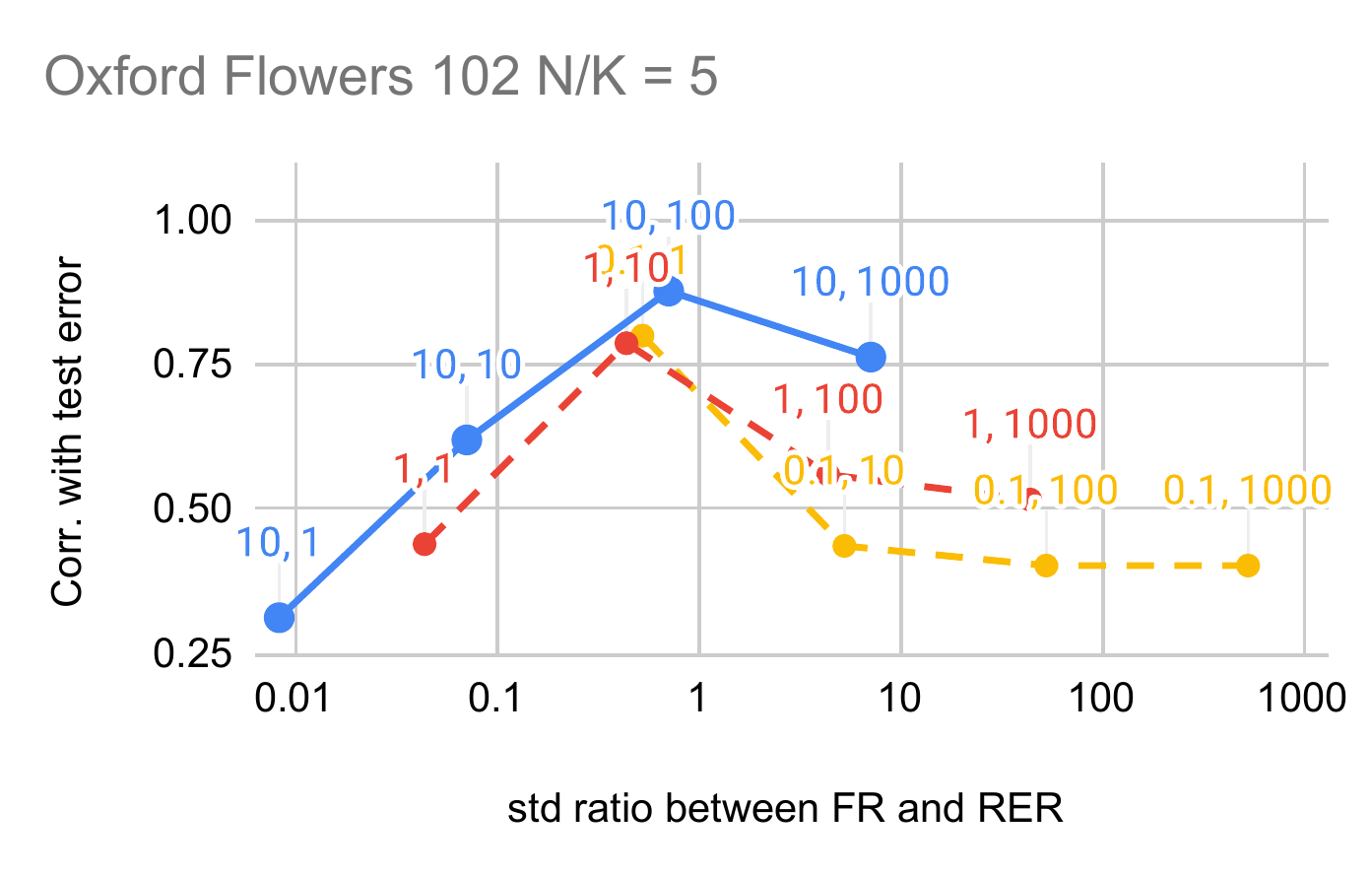}
    \includegraphics[width=0.45\textwidth]{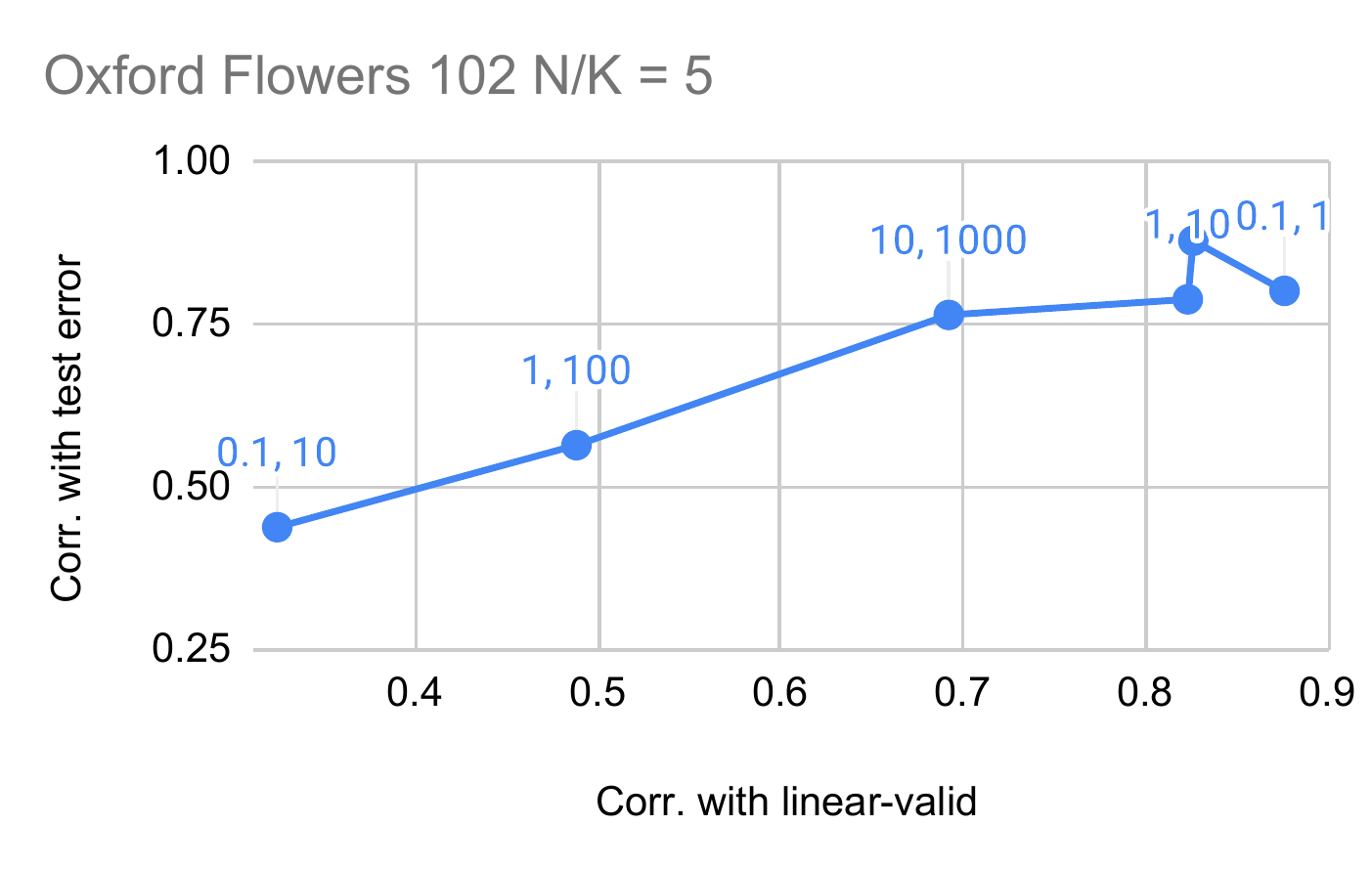}
    \includegraphics[width=0.45\textwidth]{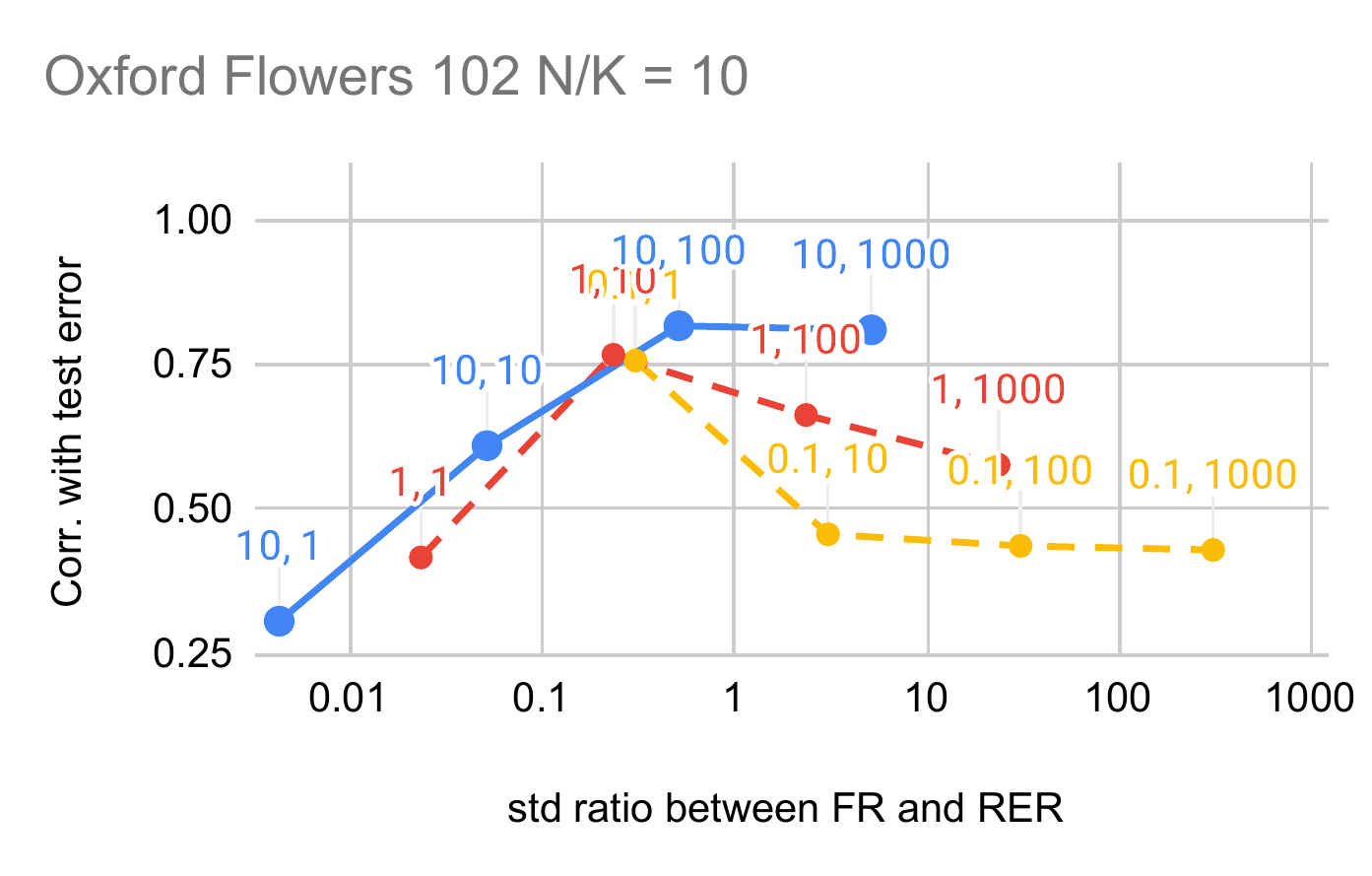}
    \includegraphics[width=0.45\textwidth]{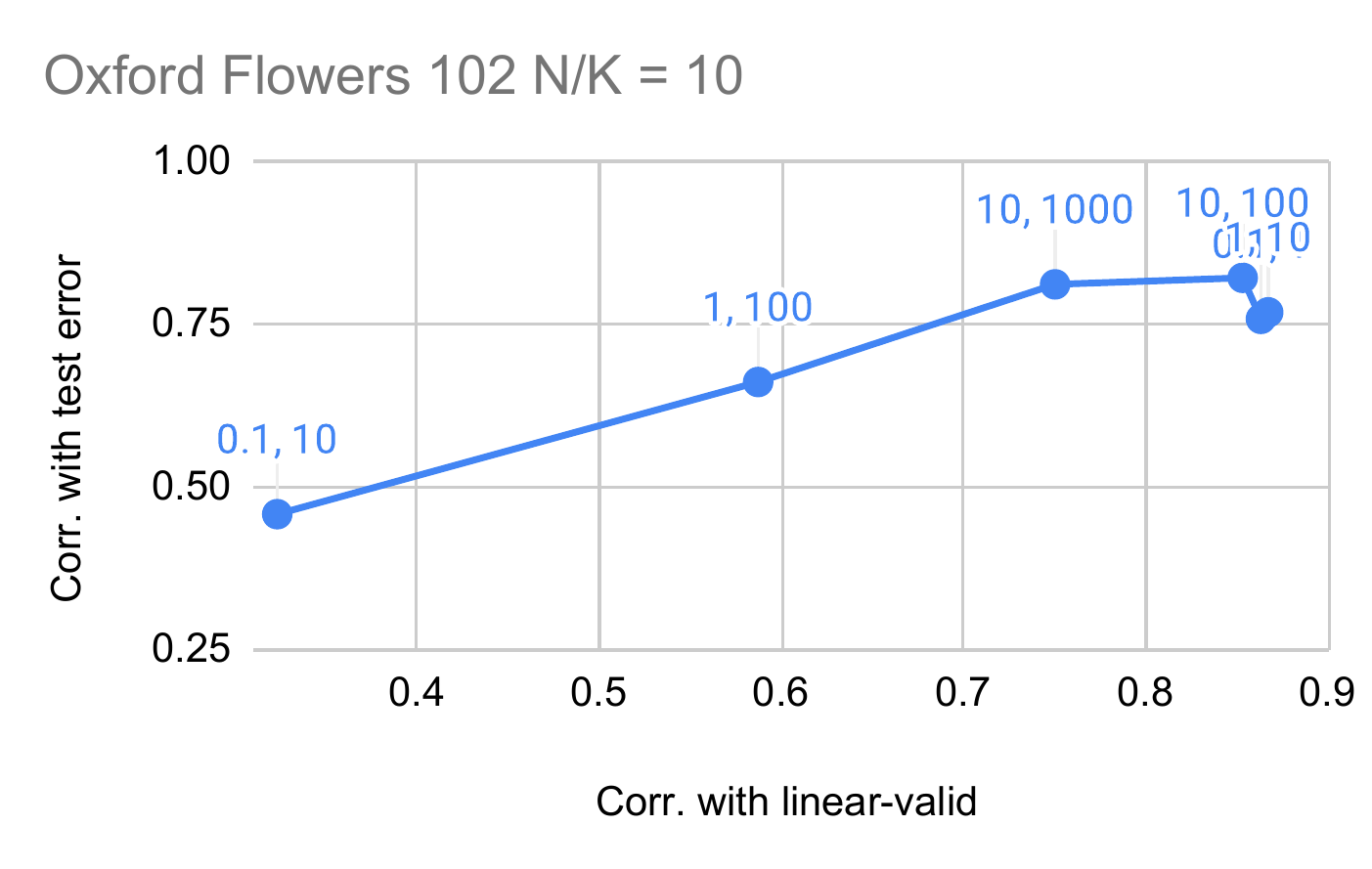}
    \caption{PACTran-Gaussian hyperparameter studies on Oxford Flowers102. Hyperparameters are labeled as $(a, b)$. High $y$-value indicates a good correlation with the downstream test error.}
    \label{fig:hp_flowers102}
\end{figure}

\begin{figure}[!ht]
    \centering
    \includegraphics[width=0.45\textwidth]{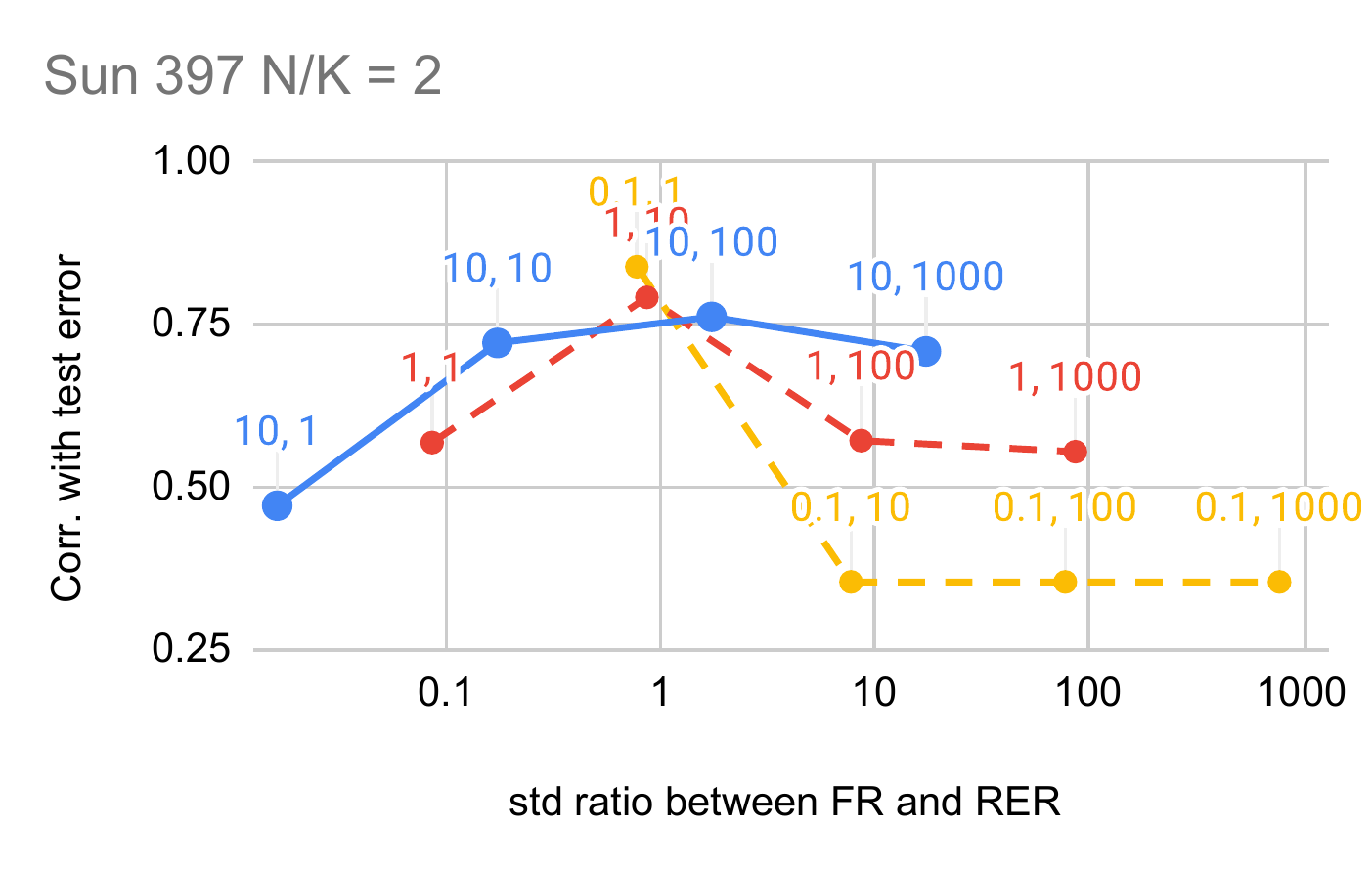}
    \includegraphics[width=0.45\textwidth]{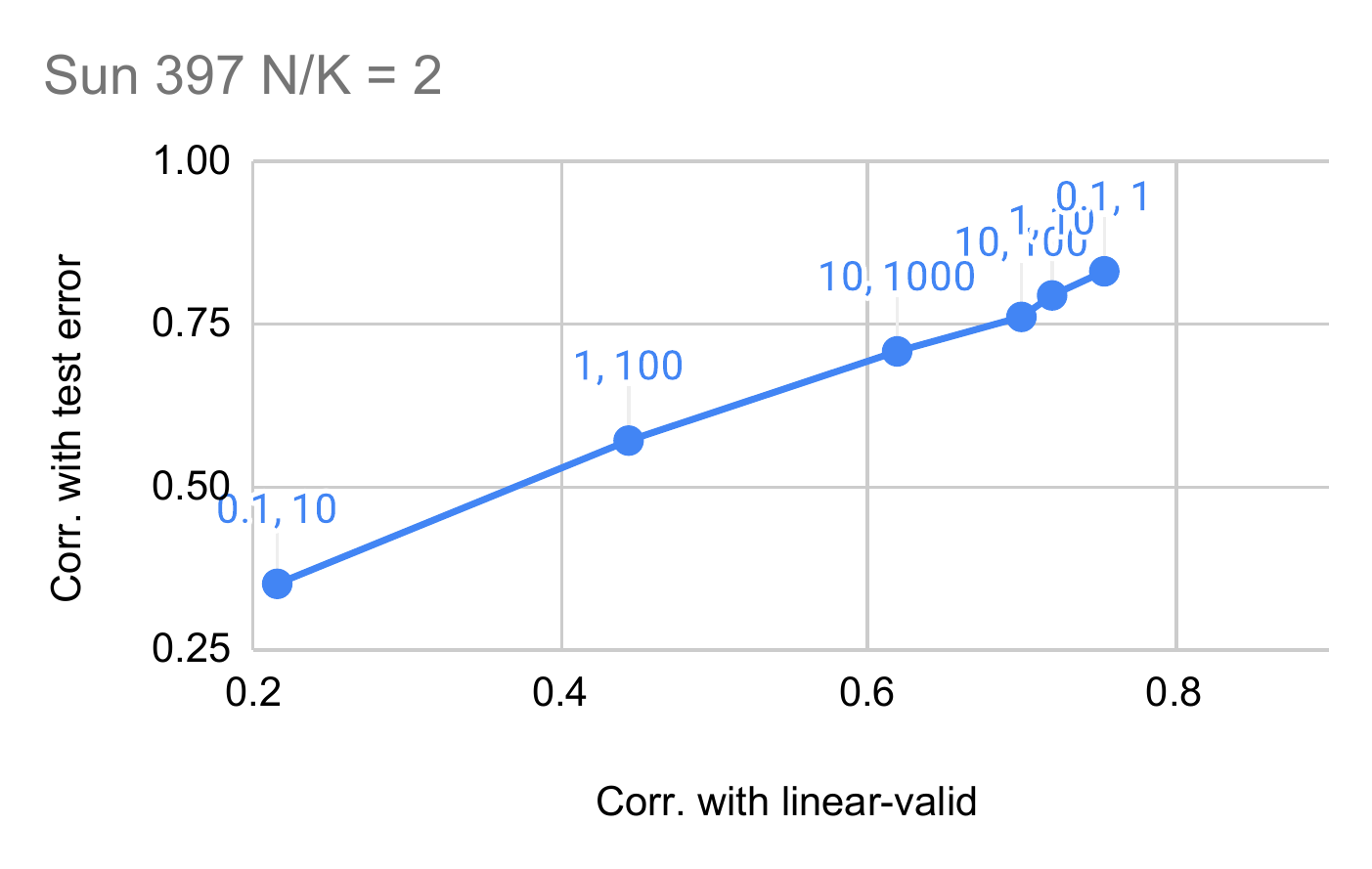}
    \includegraphics[width=0.45\textwidth]{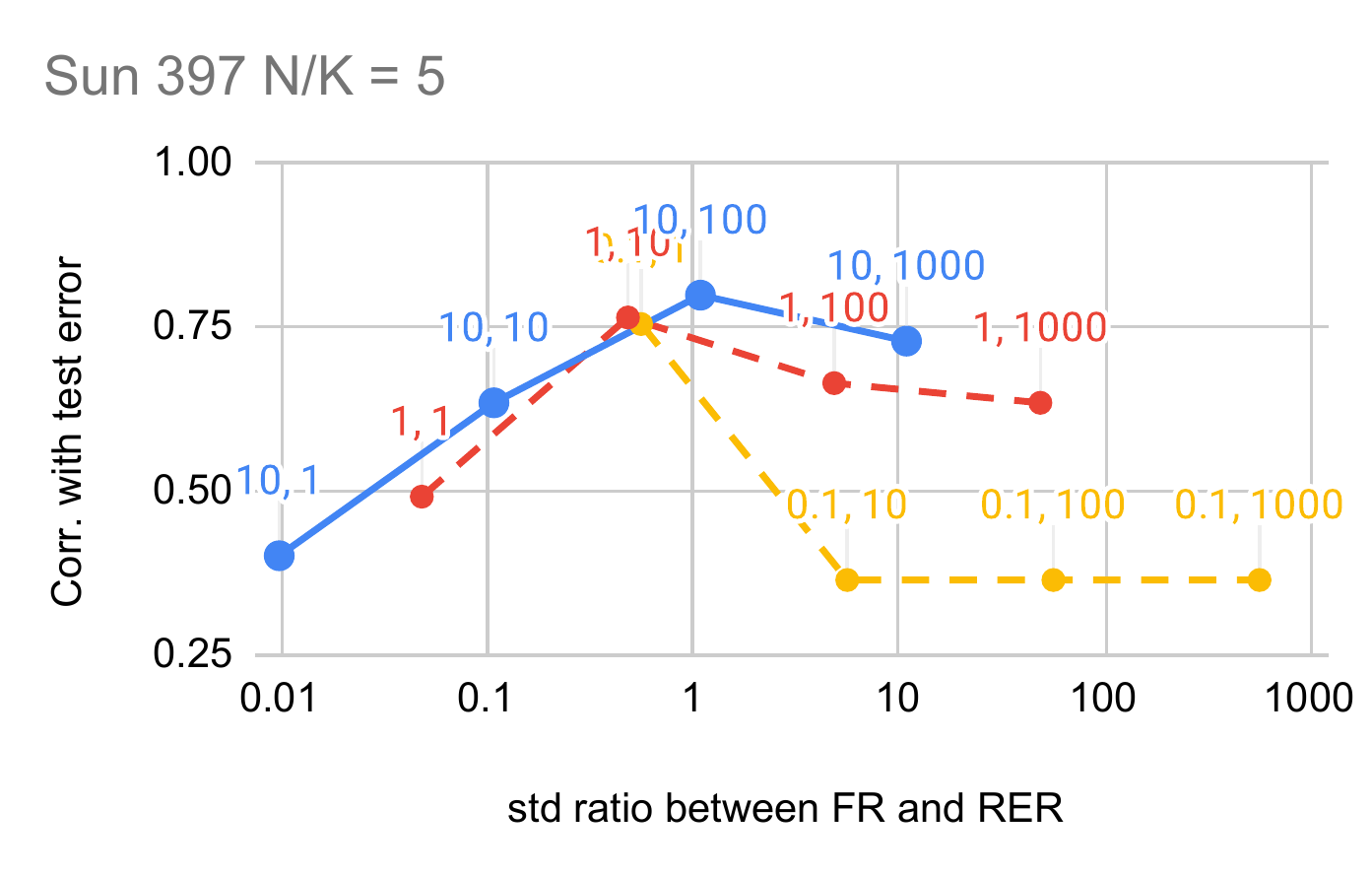}
    \includegraphics[width=0.45\textwidth]{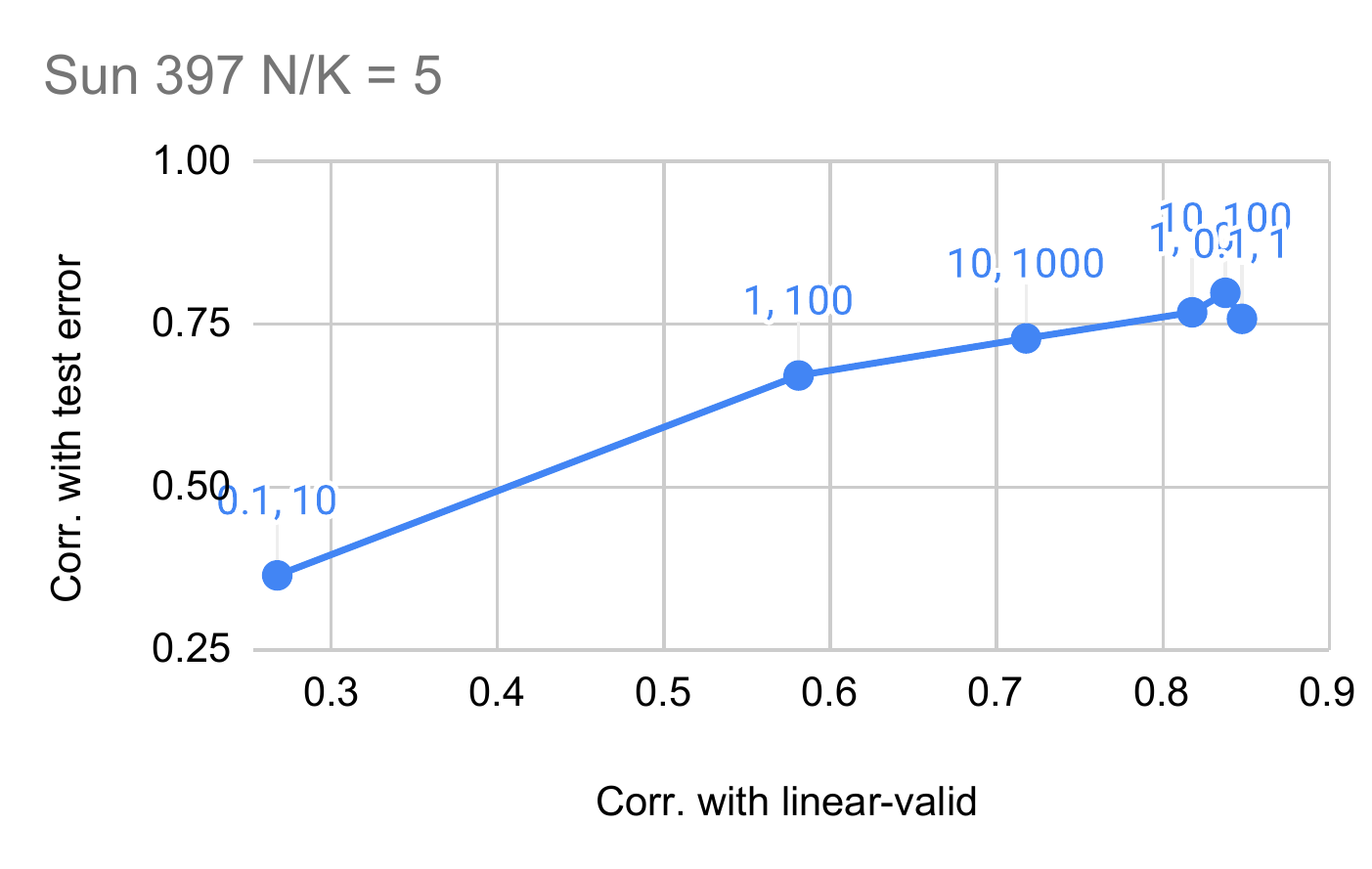}
    \includegraphics[width=0.45\textwidth]{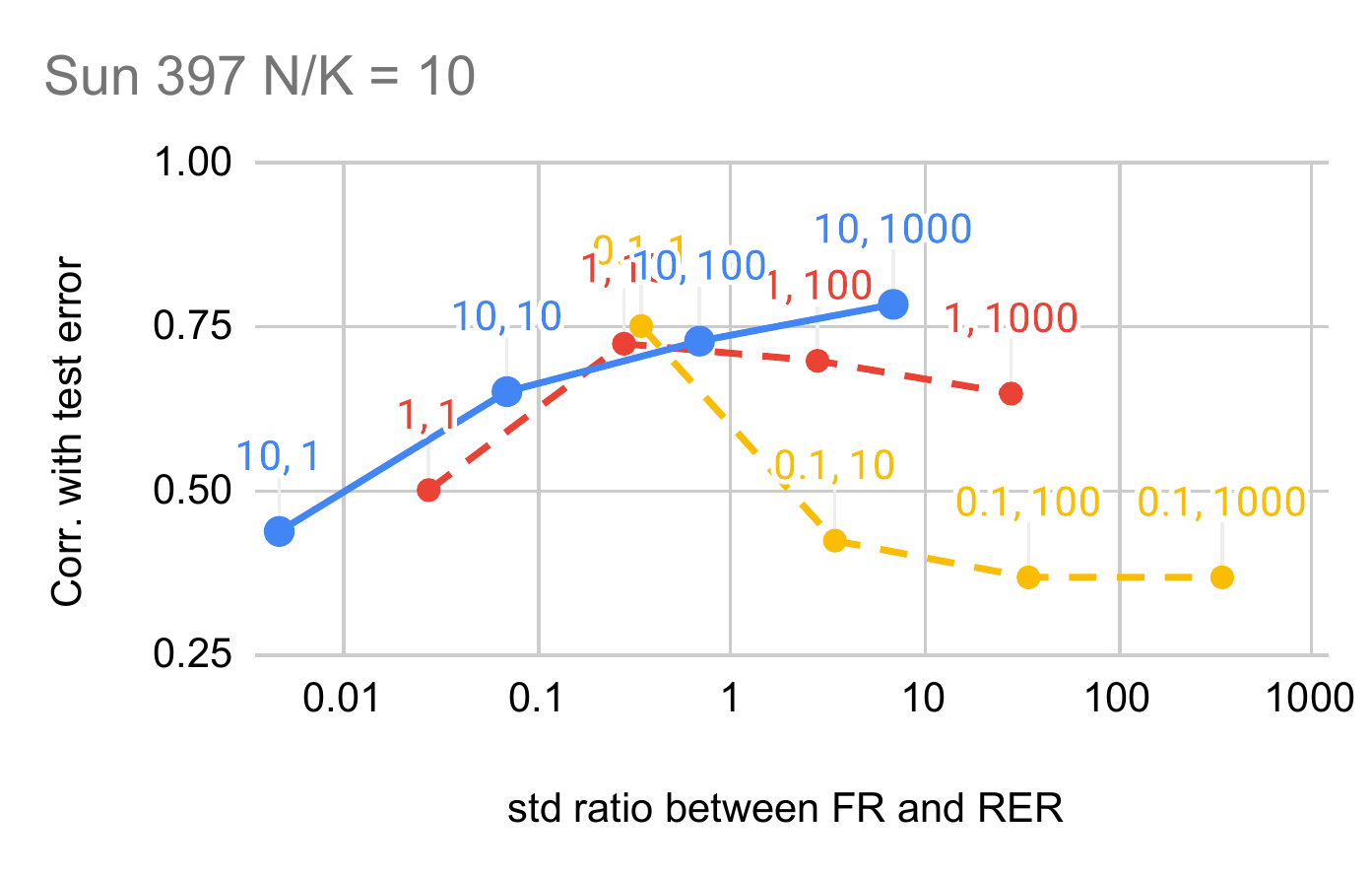}
    \includegraphics[width=0.45\textwidth]{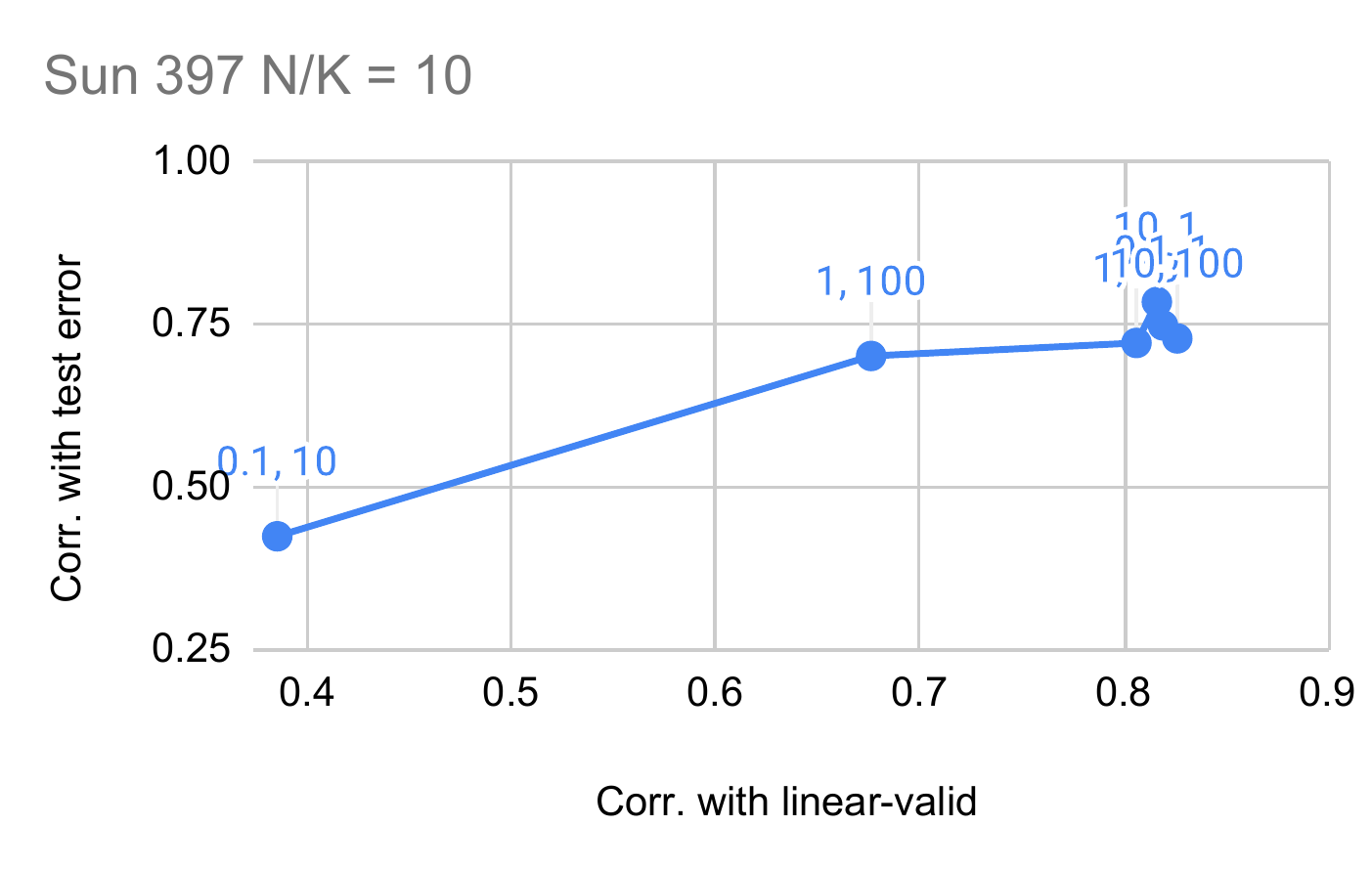}
    \caption{PACTran-Gaussian hyperparameter studies on Sun397. Hyperparameters are labeled as $(a, b)$. High $y$-value indicates a good correlation with the downstream test error.}
    \label{fig:hp_sun397}
\end{figure}

\begin{figure}[!ht]
    \centering
    \includegraphics[width=0.45\textwidth]{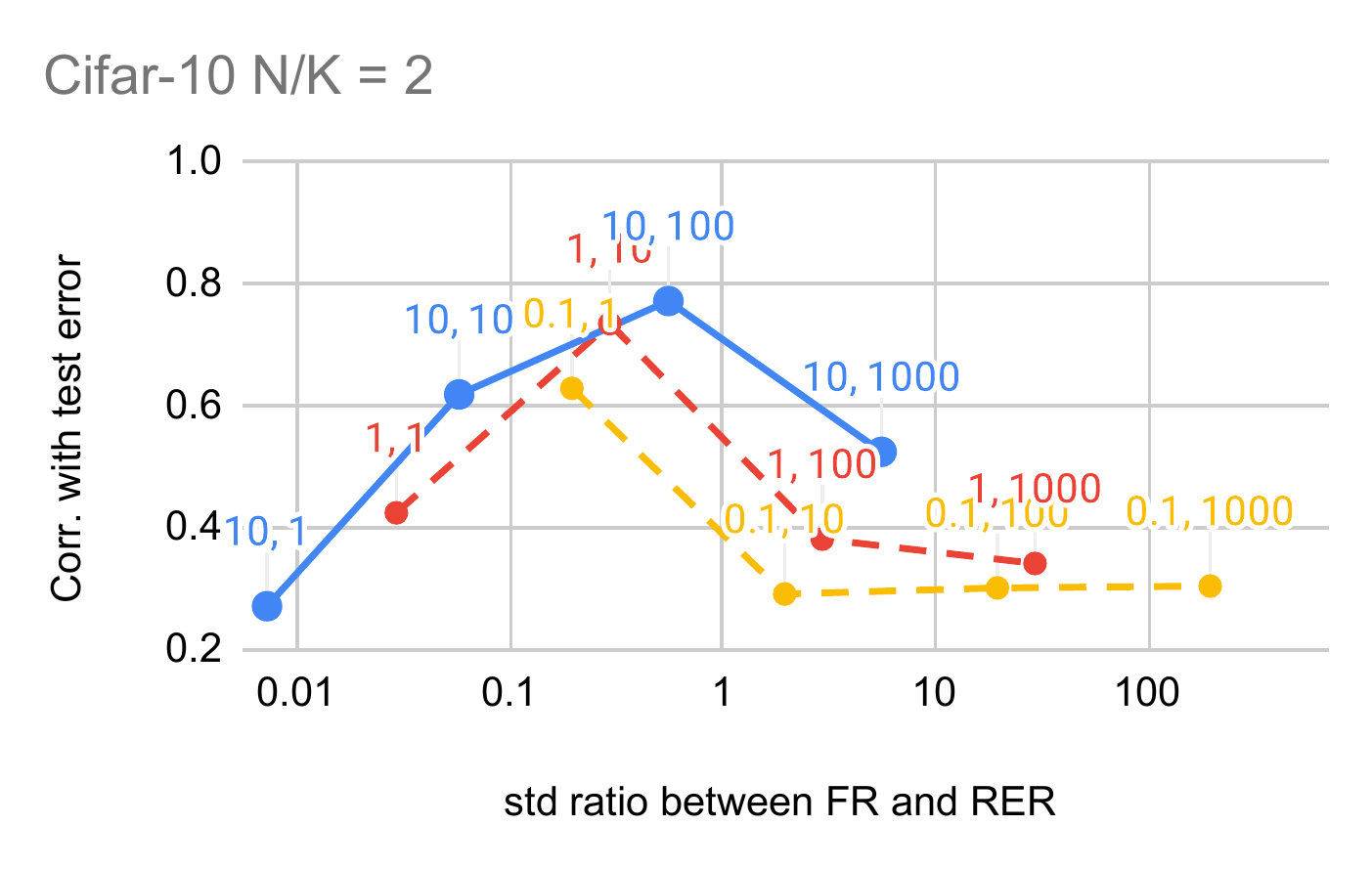}
    \includegraphics[width=0.45\textwidth]{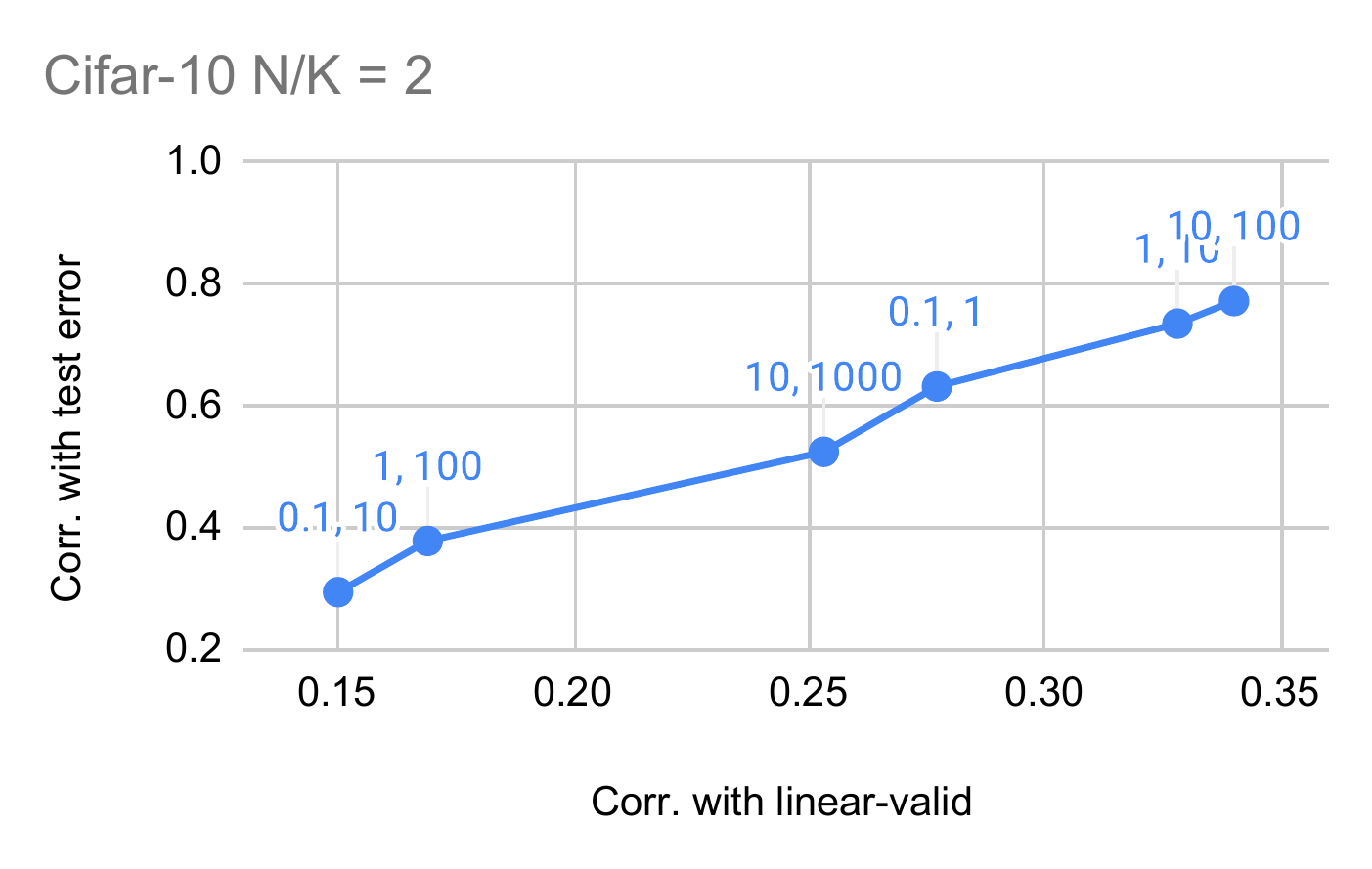}
    \includegraphics[width=0.45\textwidth]{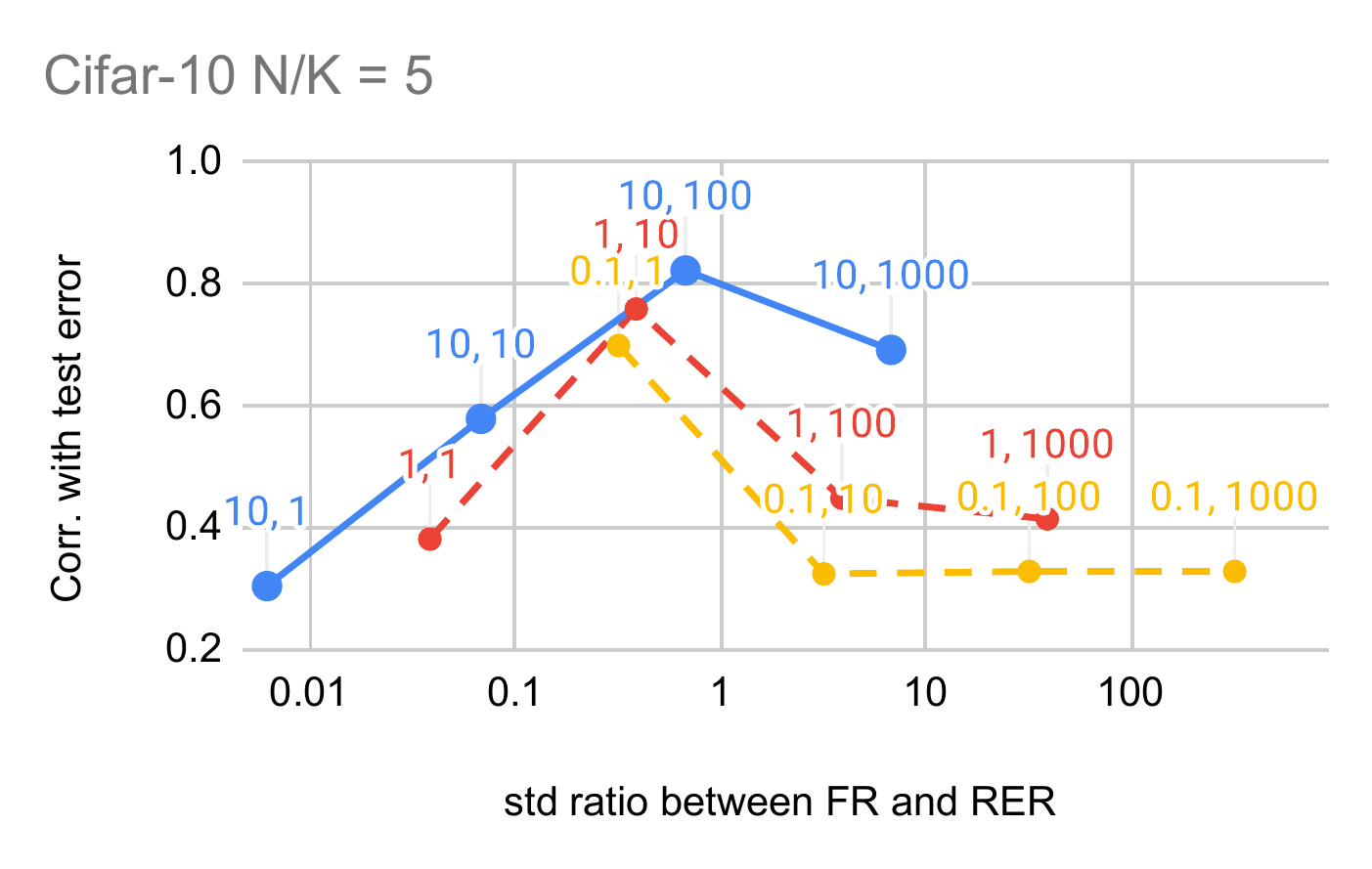}
    \includegraphics[width=0.45\textwidth]{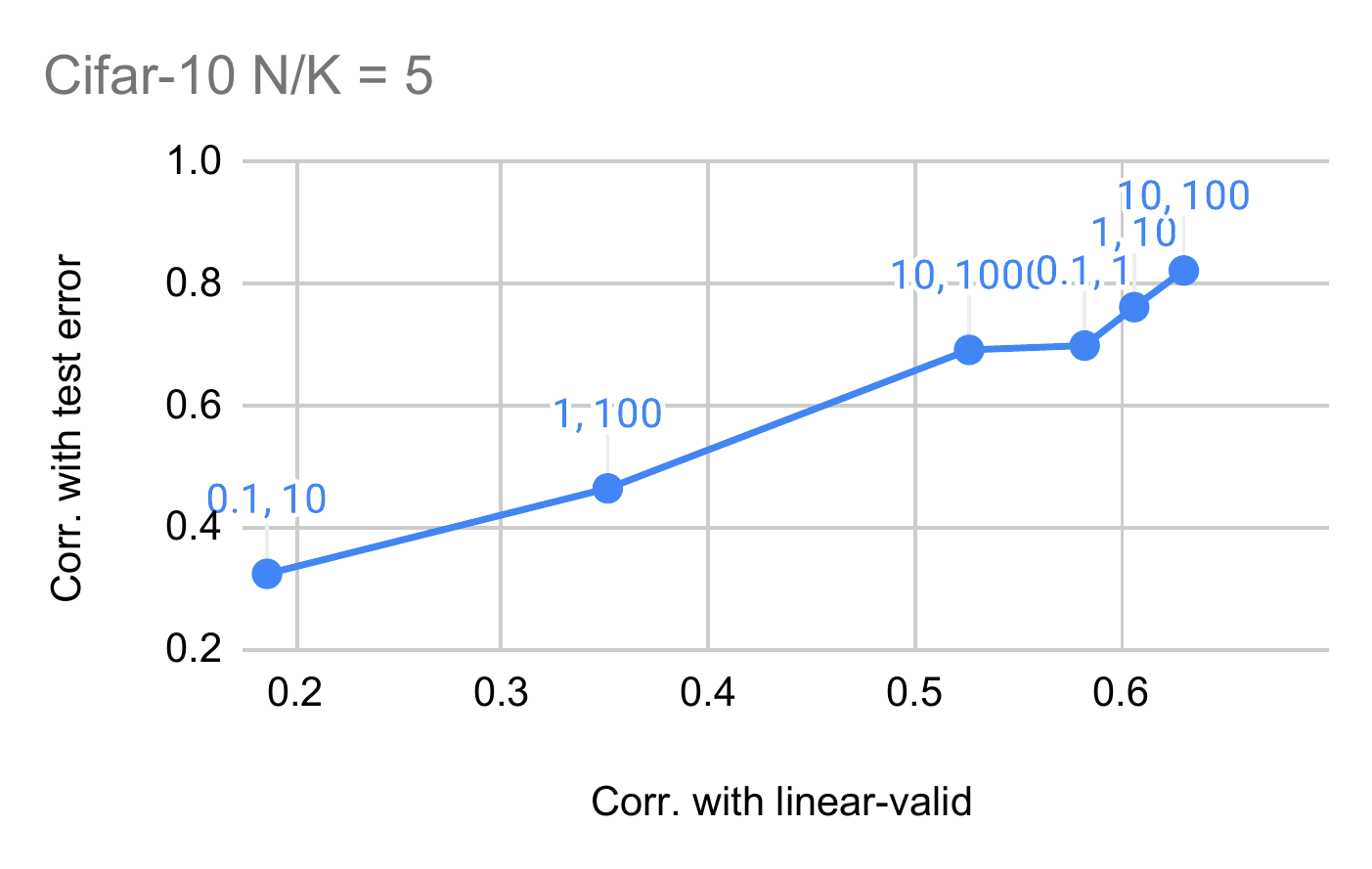}
    \includegraphics[width=0.45\textwidth]{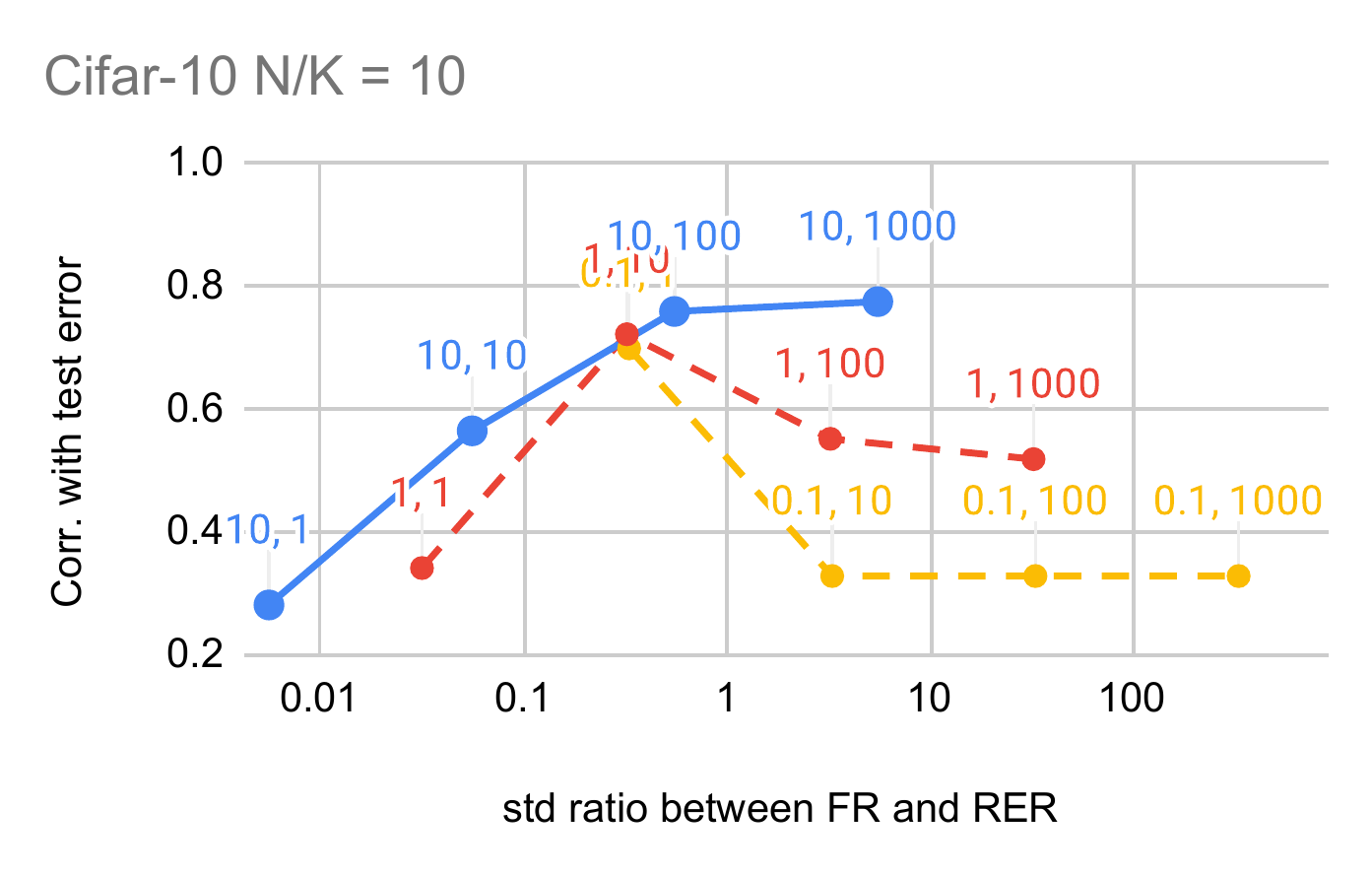}
    \includegraphics[width=0.45\textwidth]{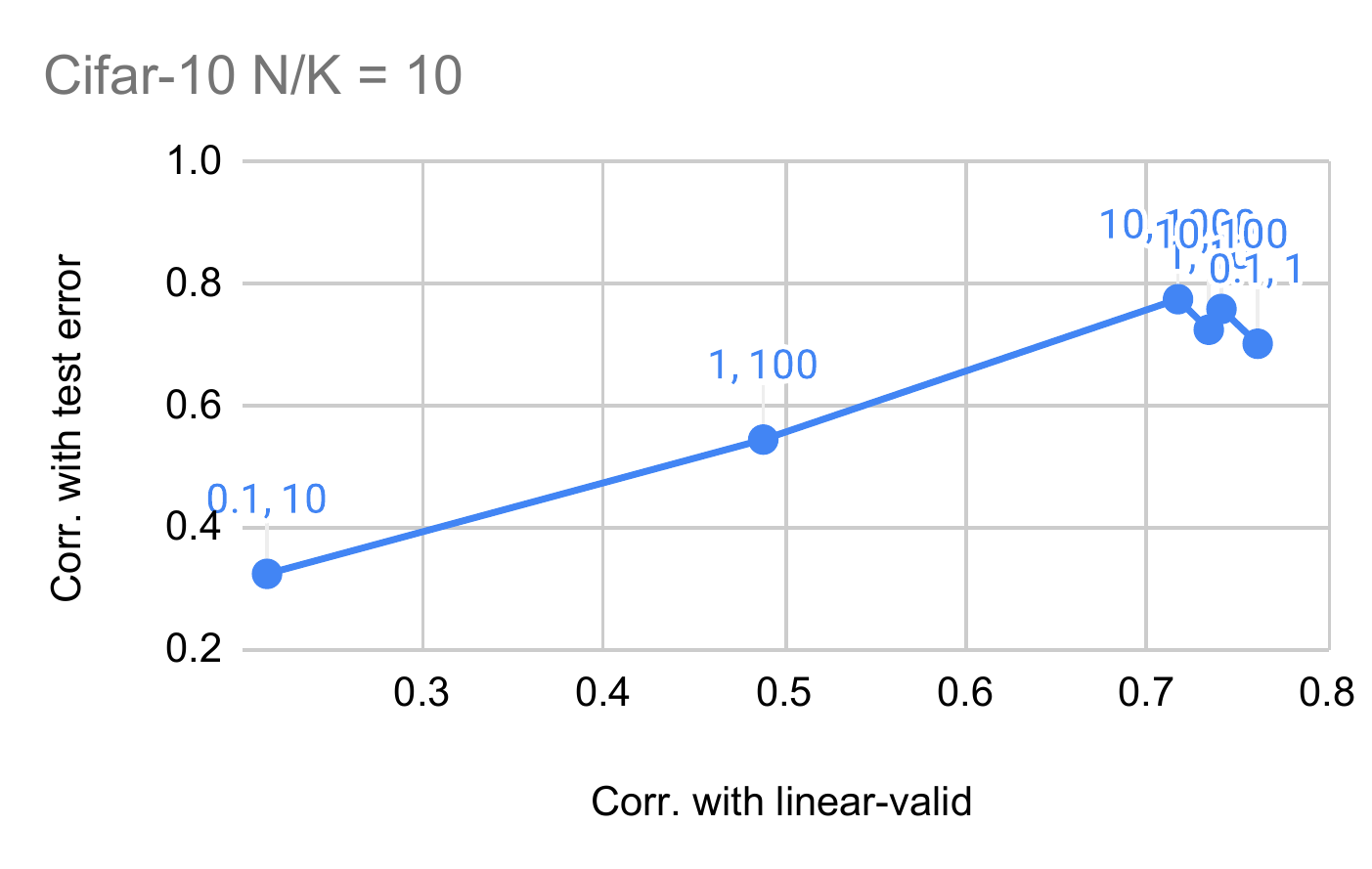}
    \caption{PACTran-Gaussian hyperparameter studies on Cifar-10. Hyperparameters are labeled as $(a, b)$. High $y$-value indicates a good correlation with the downstream test error.}
    \label{fig:hp_cifar_10}
\end{figure}

\begin{figure}[!ht]
    \centering
    \includegraphics[width=0.45\textwidth]{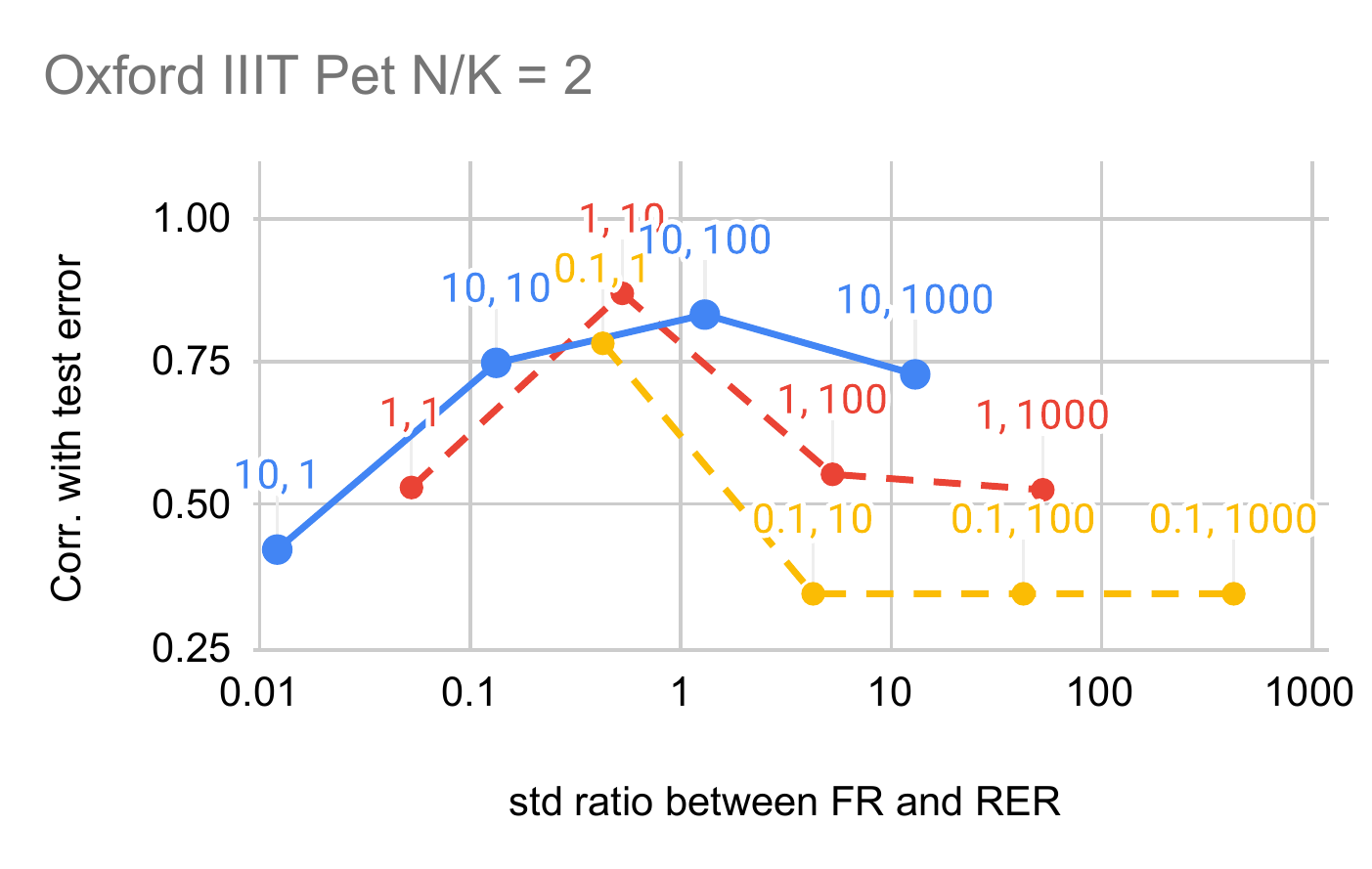}
    \includegraphics[width=0.45\textwidth]{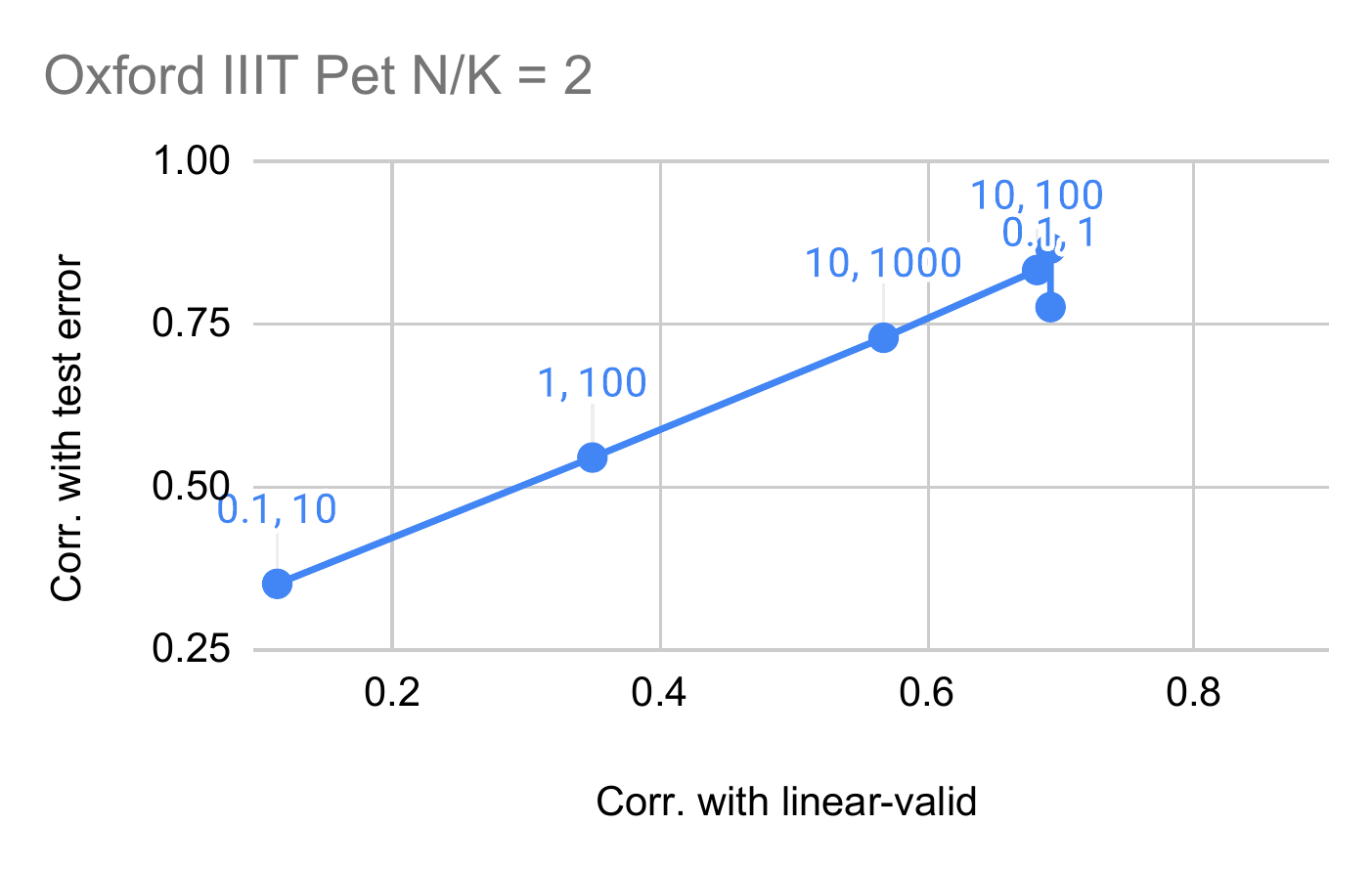}
    \includegraphics[width=0.45\textwidth]{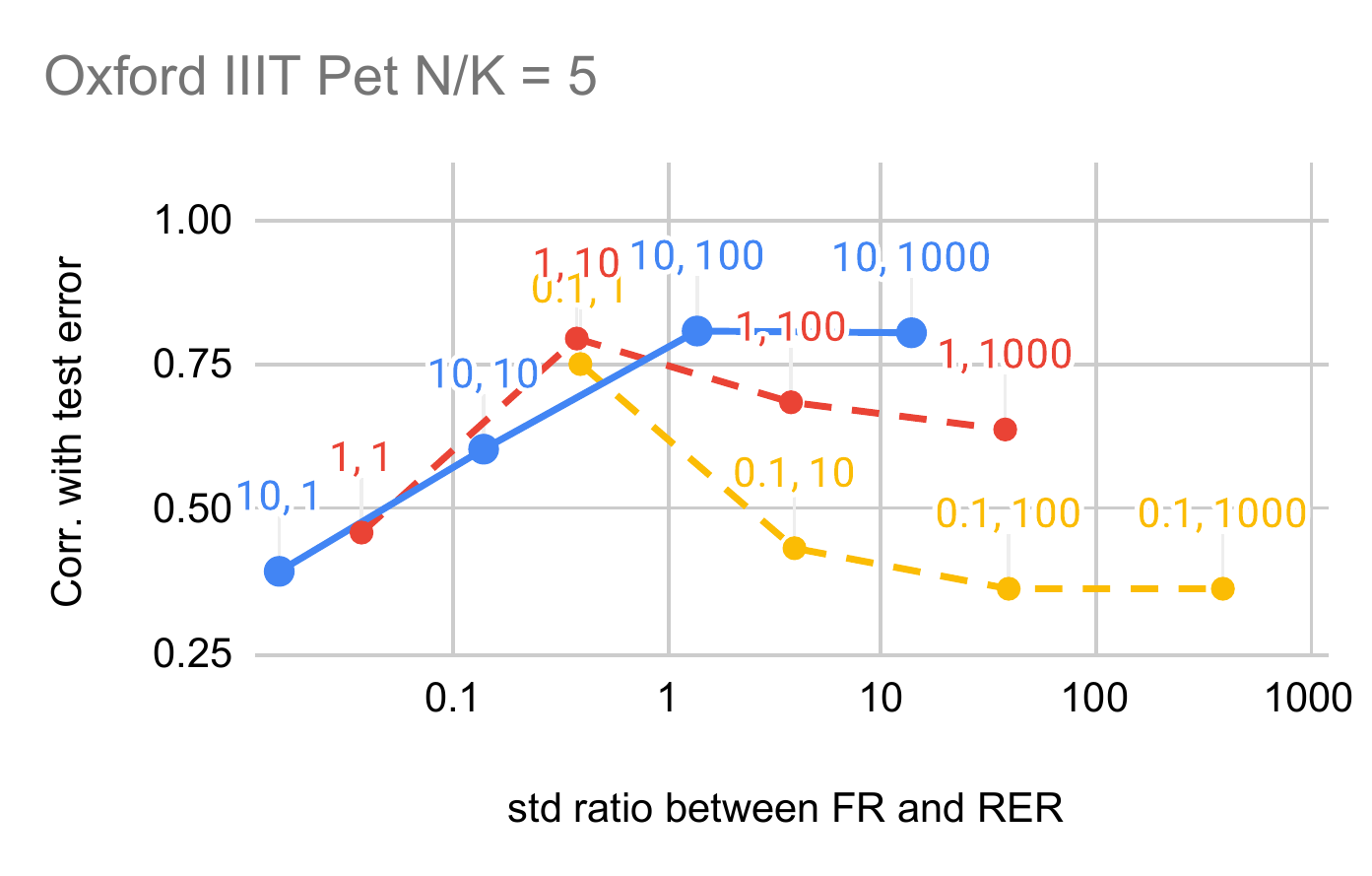}
    \includegraphics[width=0.45\textwidth]{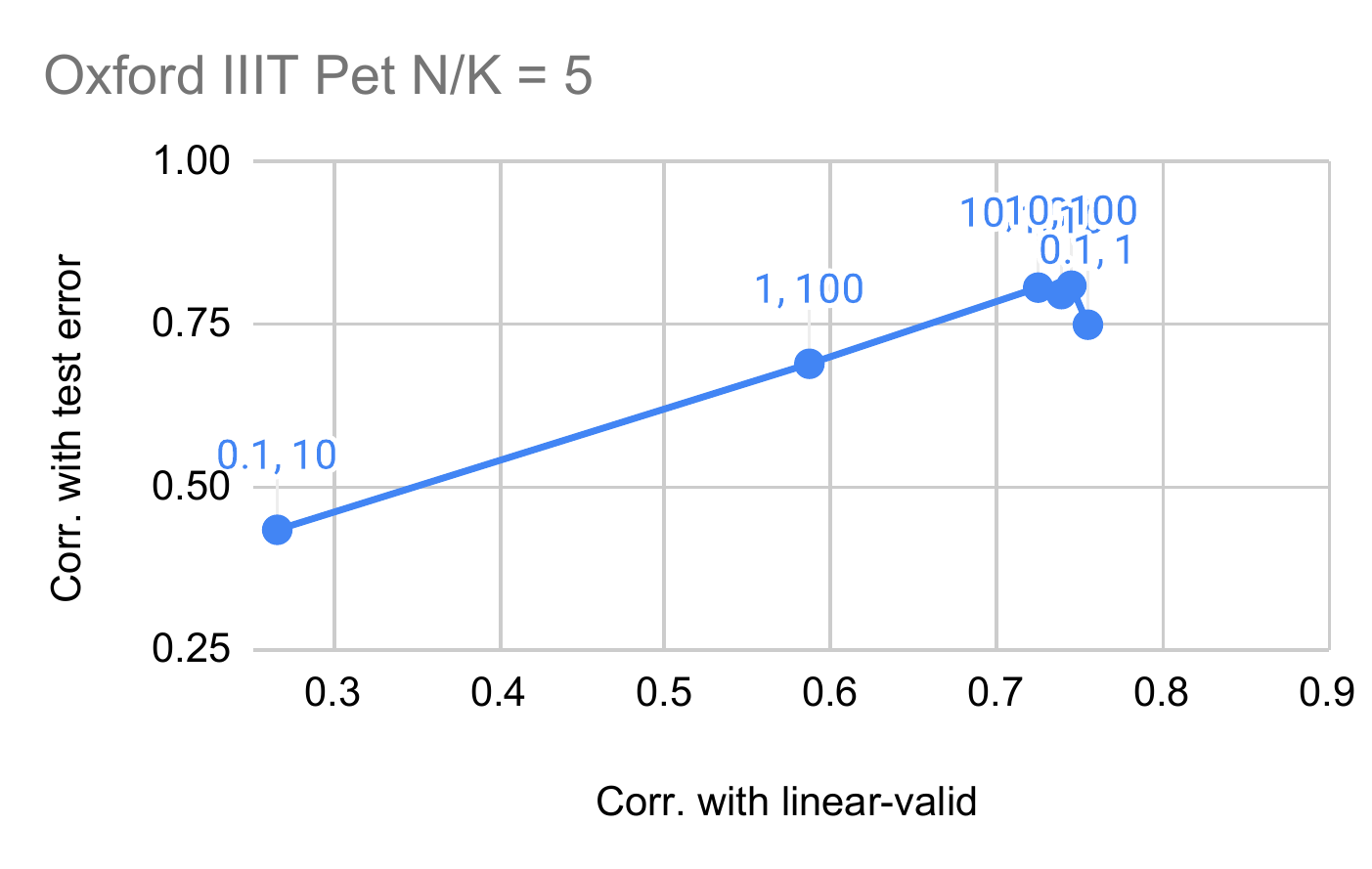}
    \includegraphics[width=0.45\textwidth]{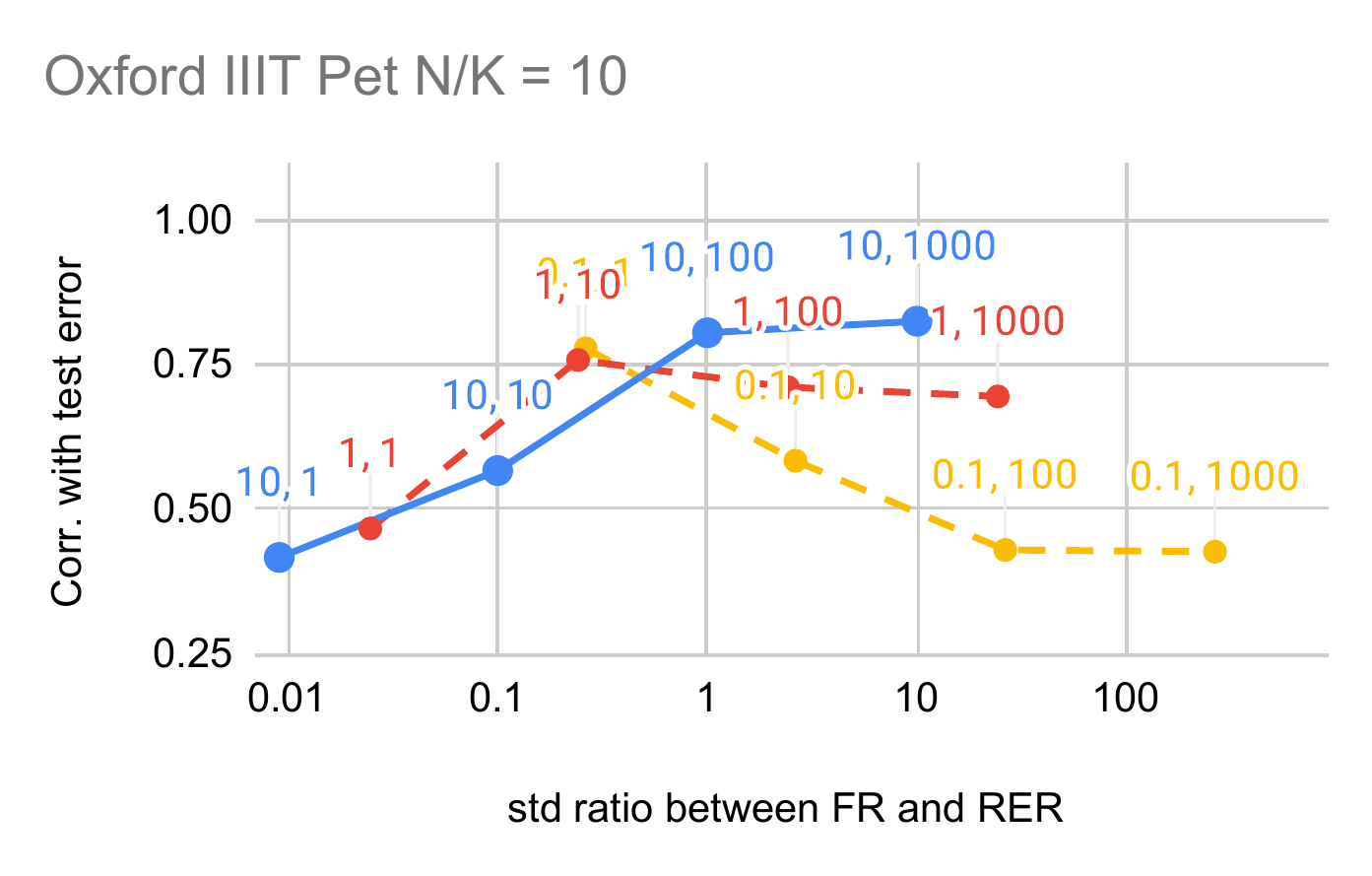}
    \includegraphics[width=0.45\textwidth]{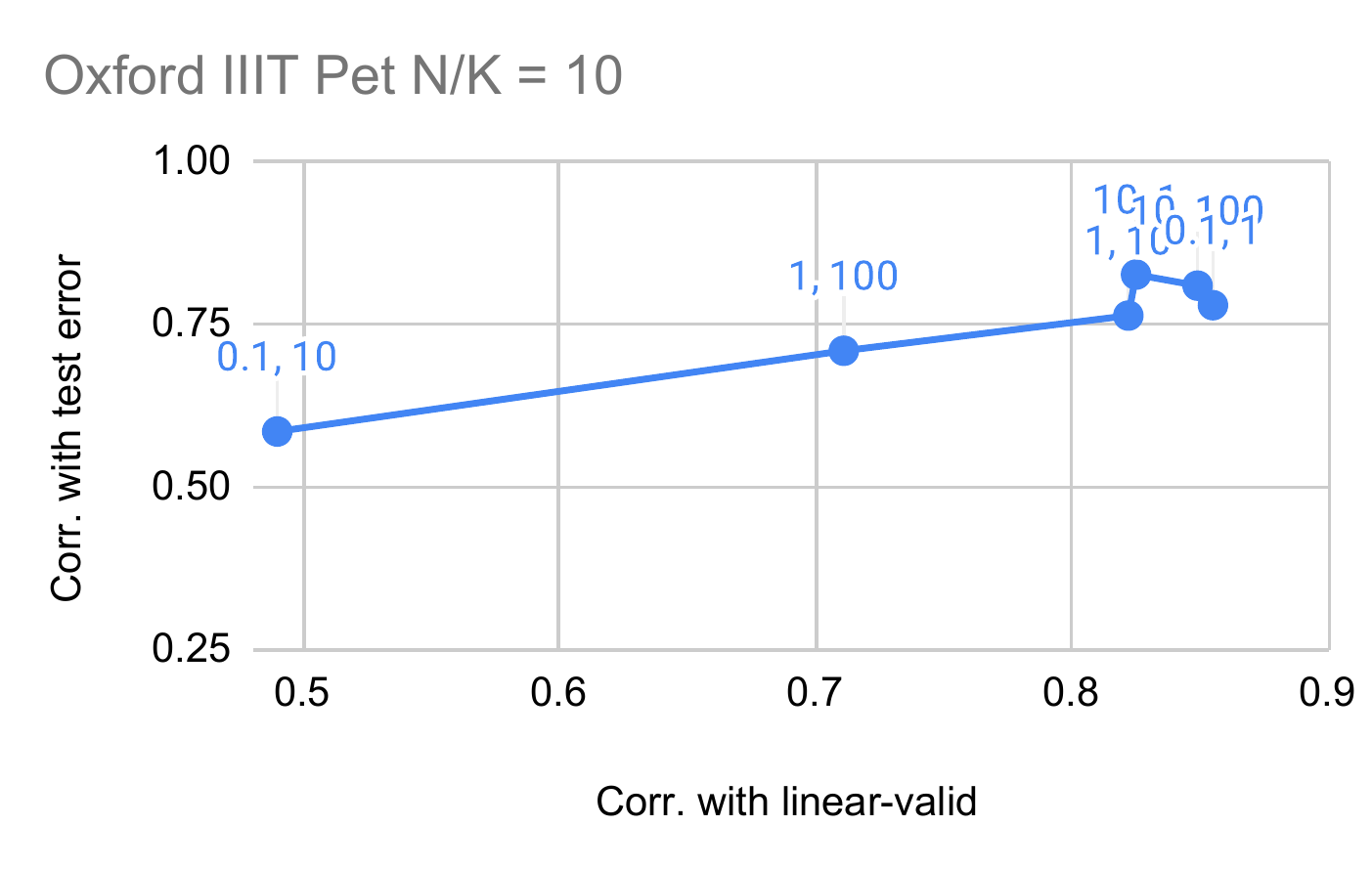}
    \caption{PACTran-Gaussian hyperparameter studies on Oxford IIIT Pet. Hyperparameters are labeled as $(a, b)$. High $y$-value indicates a good correlation with the downstream test error.}
    \label{fig:hp_oxford_iiit_pet}
\end{figure}

\begin{figure}[!ht]
    \centering
    \includegraphics[width=0.45\textwidth]{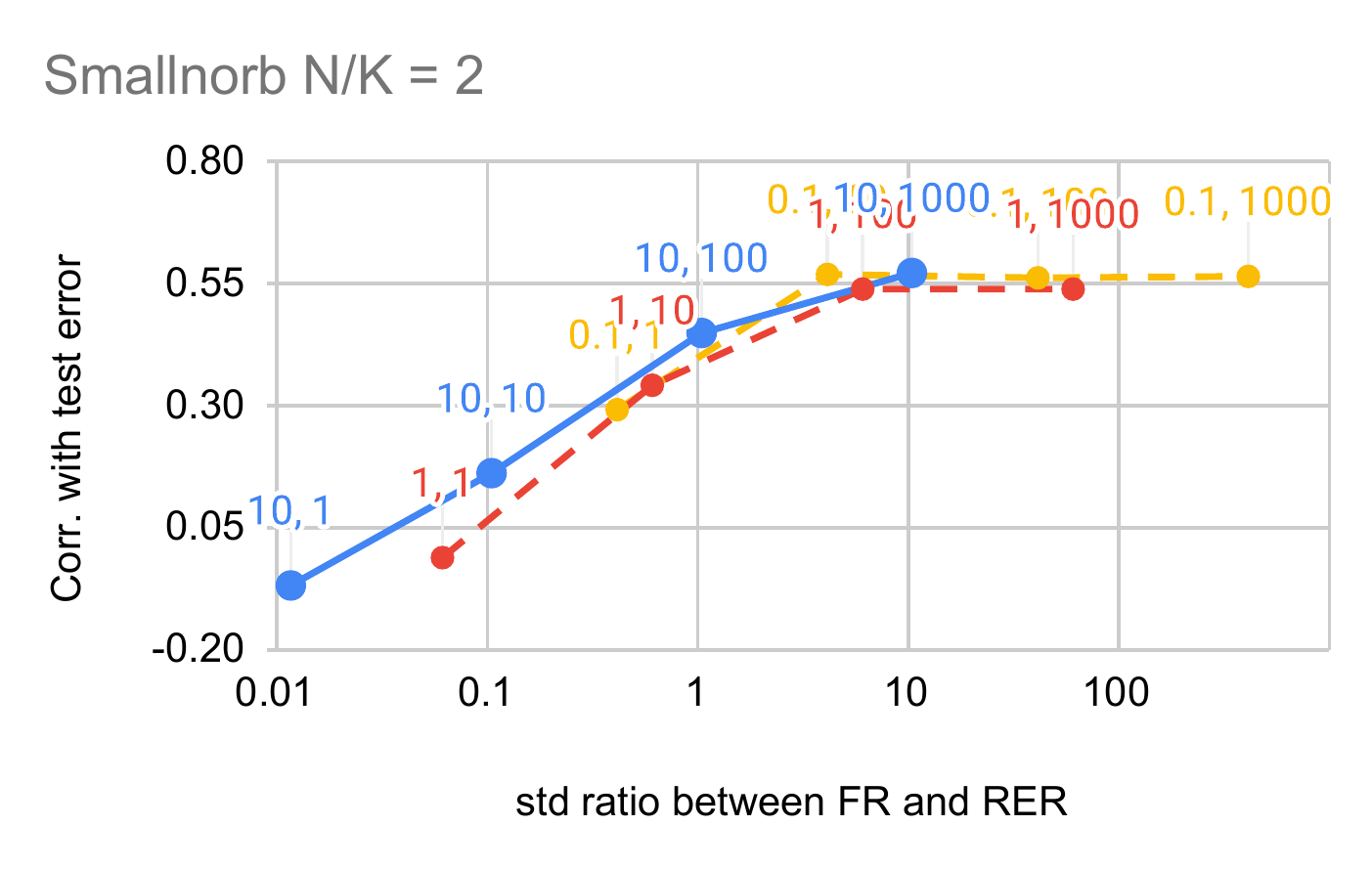}
    \includegraphics[width=0.45\textwidth]{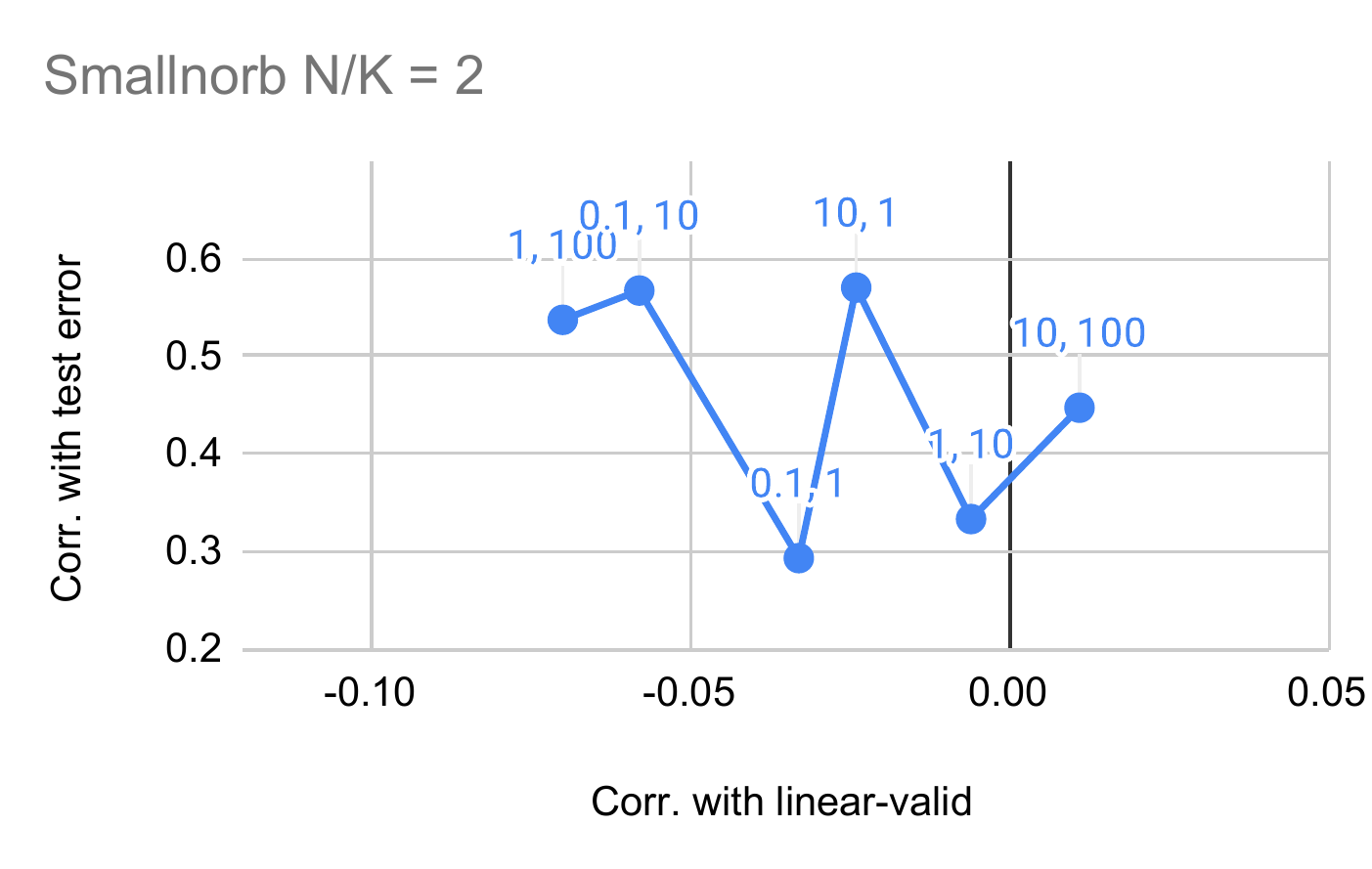}
    \includegraphics[width=0.45\textwidth]{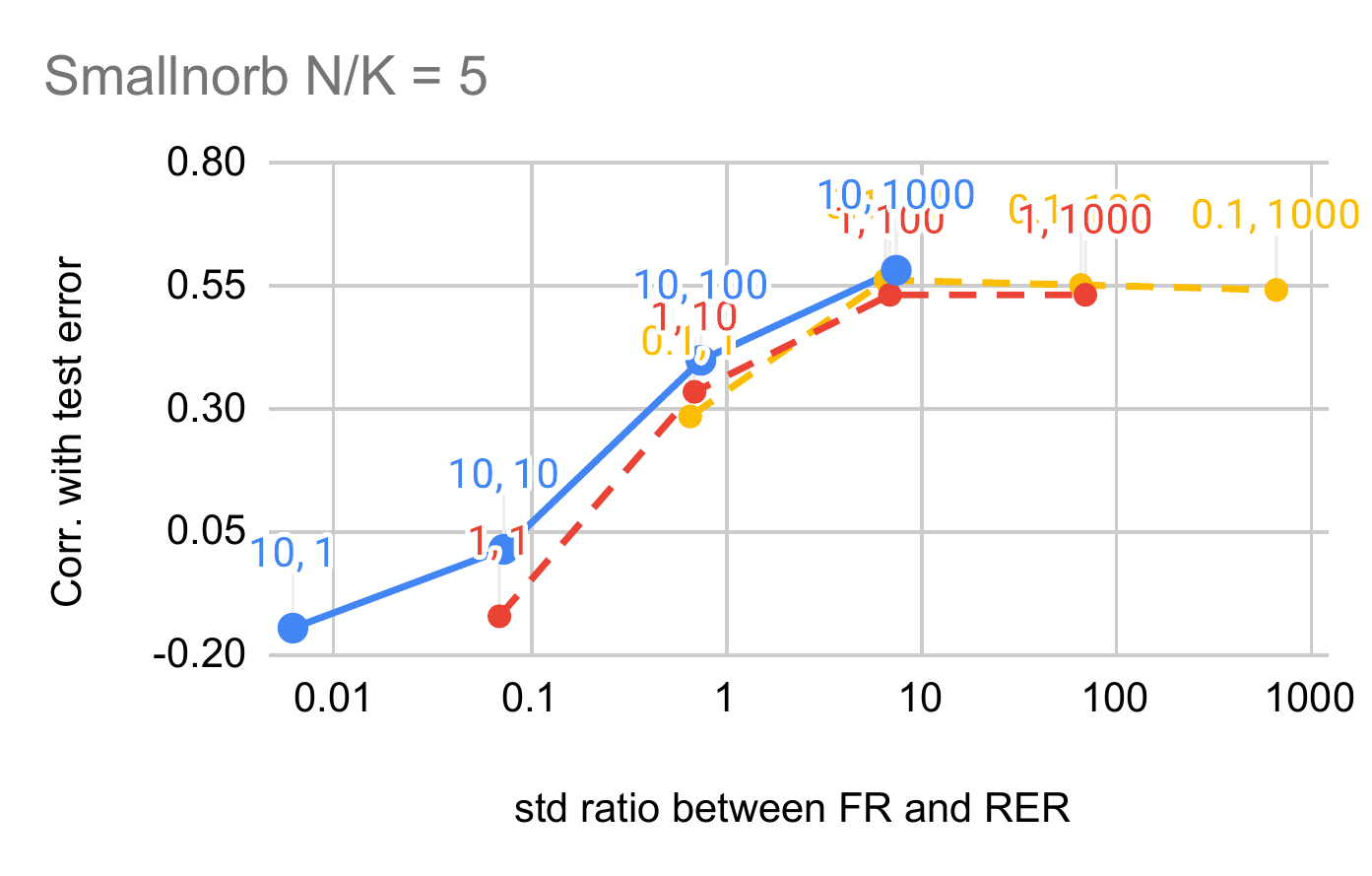}
    \includegraphics[width=0.45\textwidth]{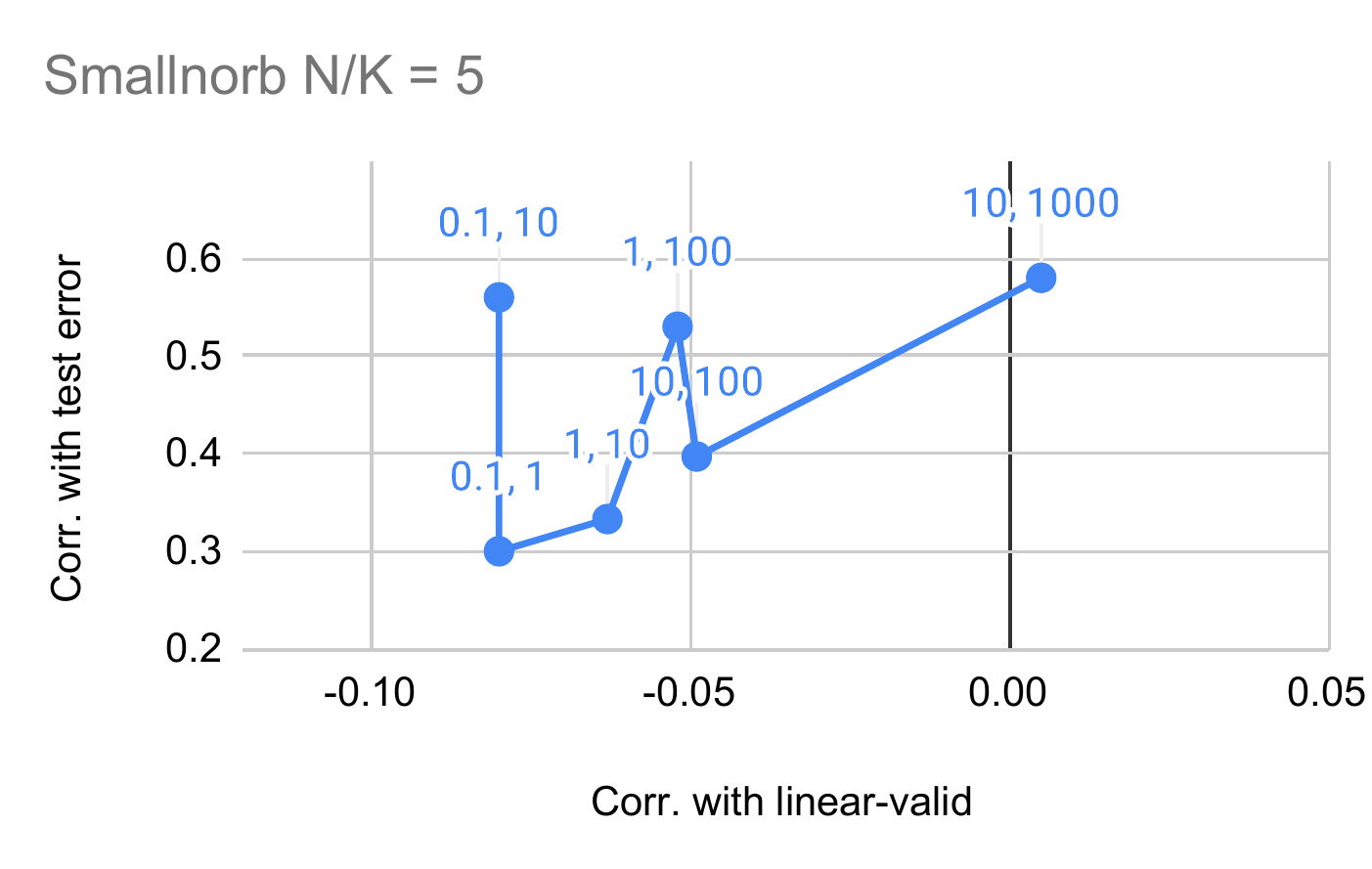}
    \includegraphics[width=0.45\textwidth]{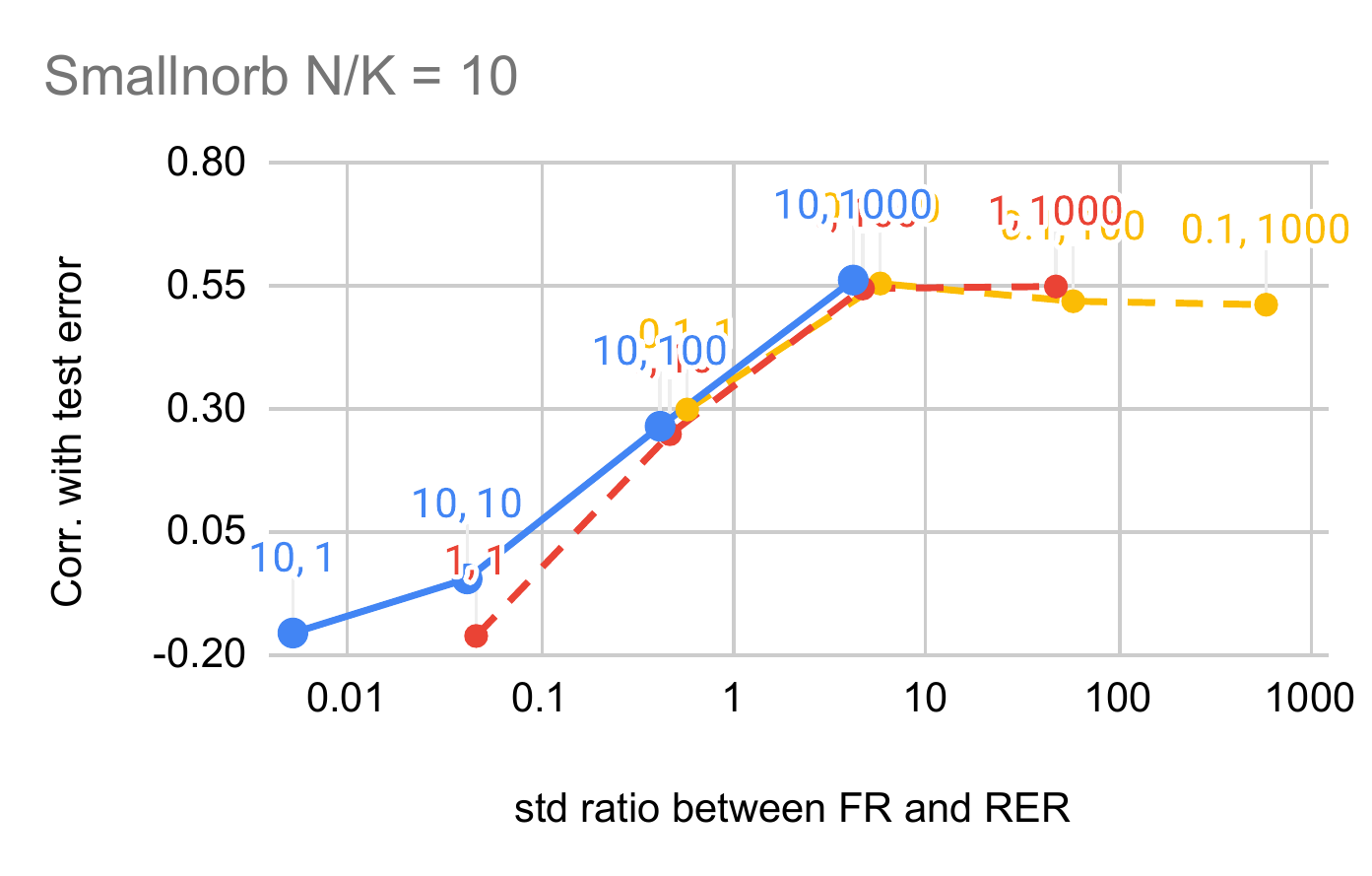}
    \includegraphics[width=0.45\textwidth]{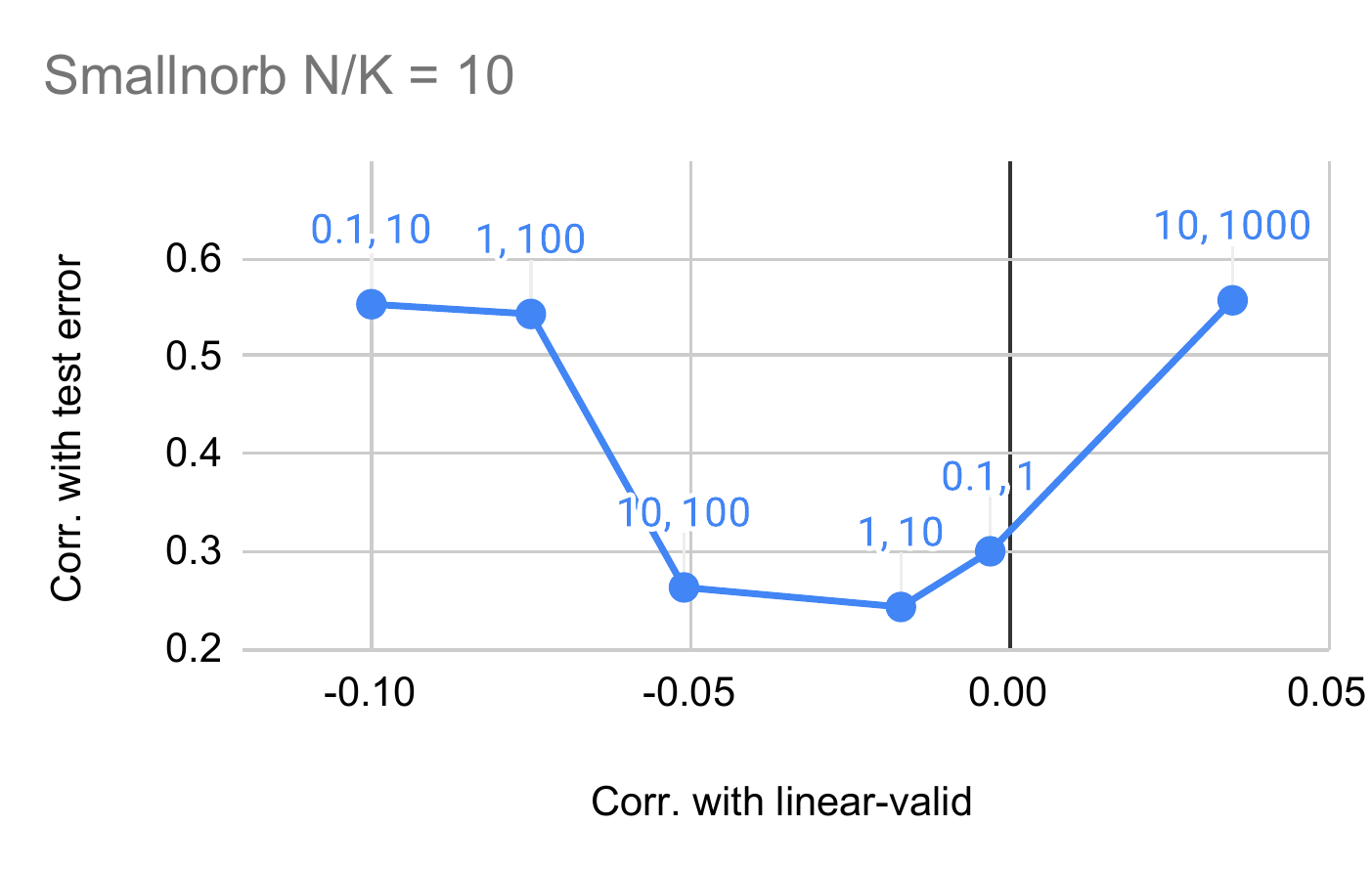}
    \caption{PACTran-Gaussian hyperparameter studies on Smallnorb. Hyperparameters are labeled as $(a, b)$. High $y$-value indicates a good correlation with the downstream test error.}
    \label{fig:hp_smallnorb}
\end{figure}

\begin{figure}[!ht]
    \centering
    \includegraphics[width=0.45\textwidth]{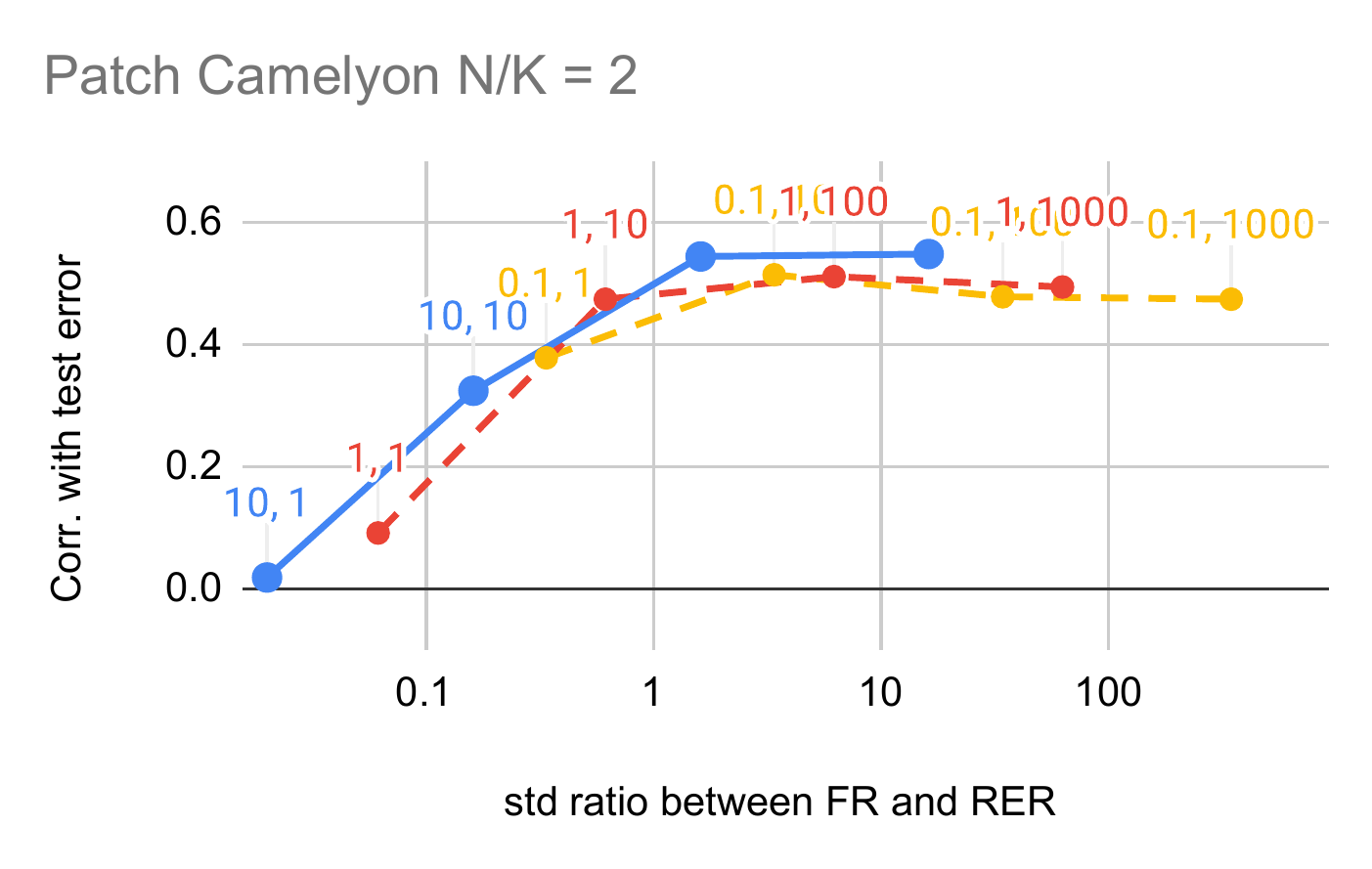}
    \includegraphics[width=0.45\textwidth]{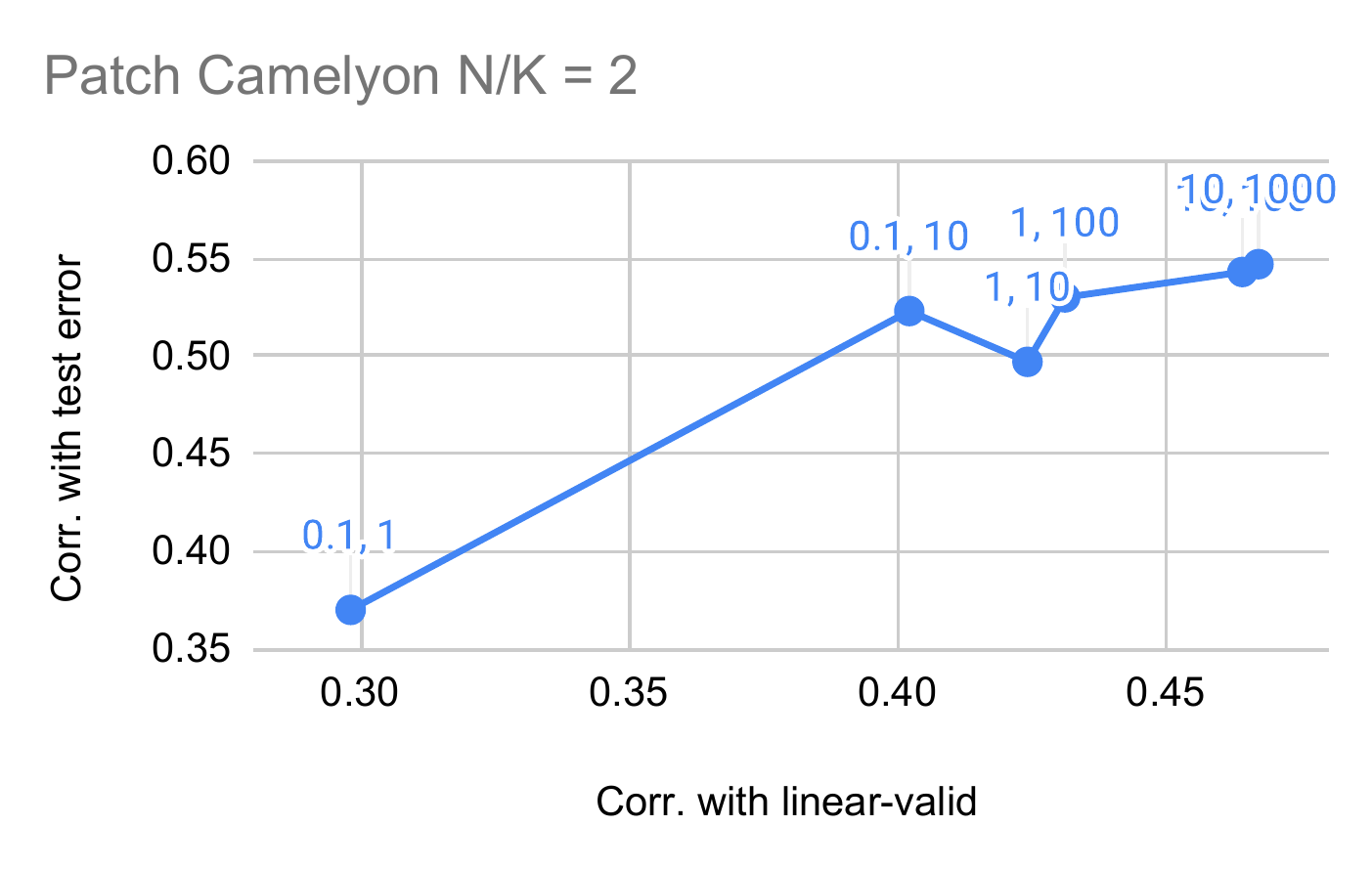}
    \includegraphics[width=0.45\textwidth]{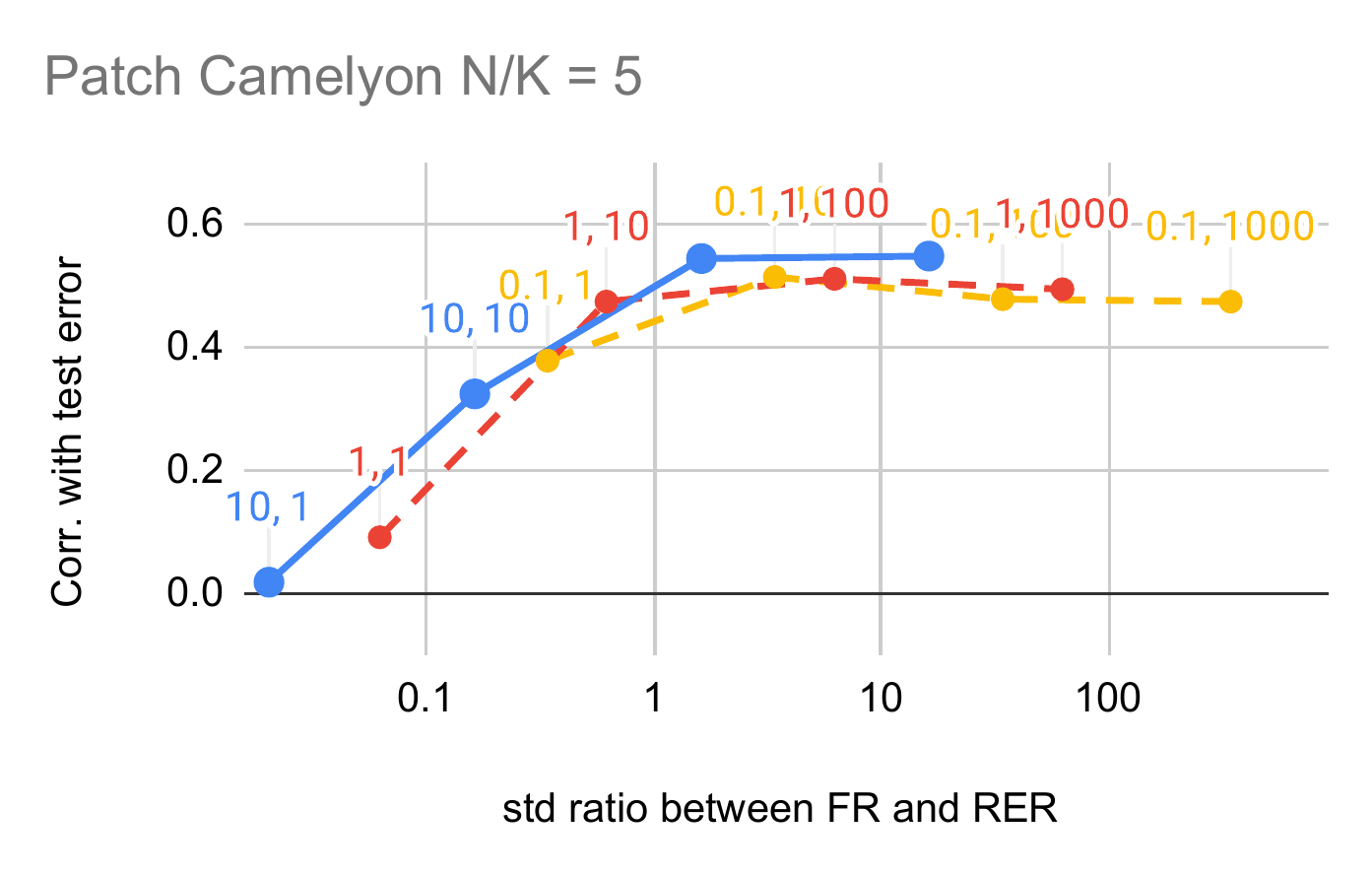}
    \includegraphics[width=0.45\textwidth]{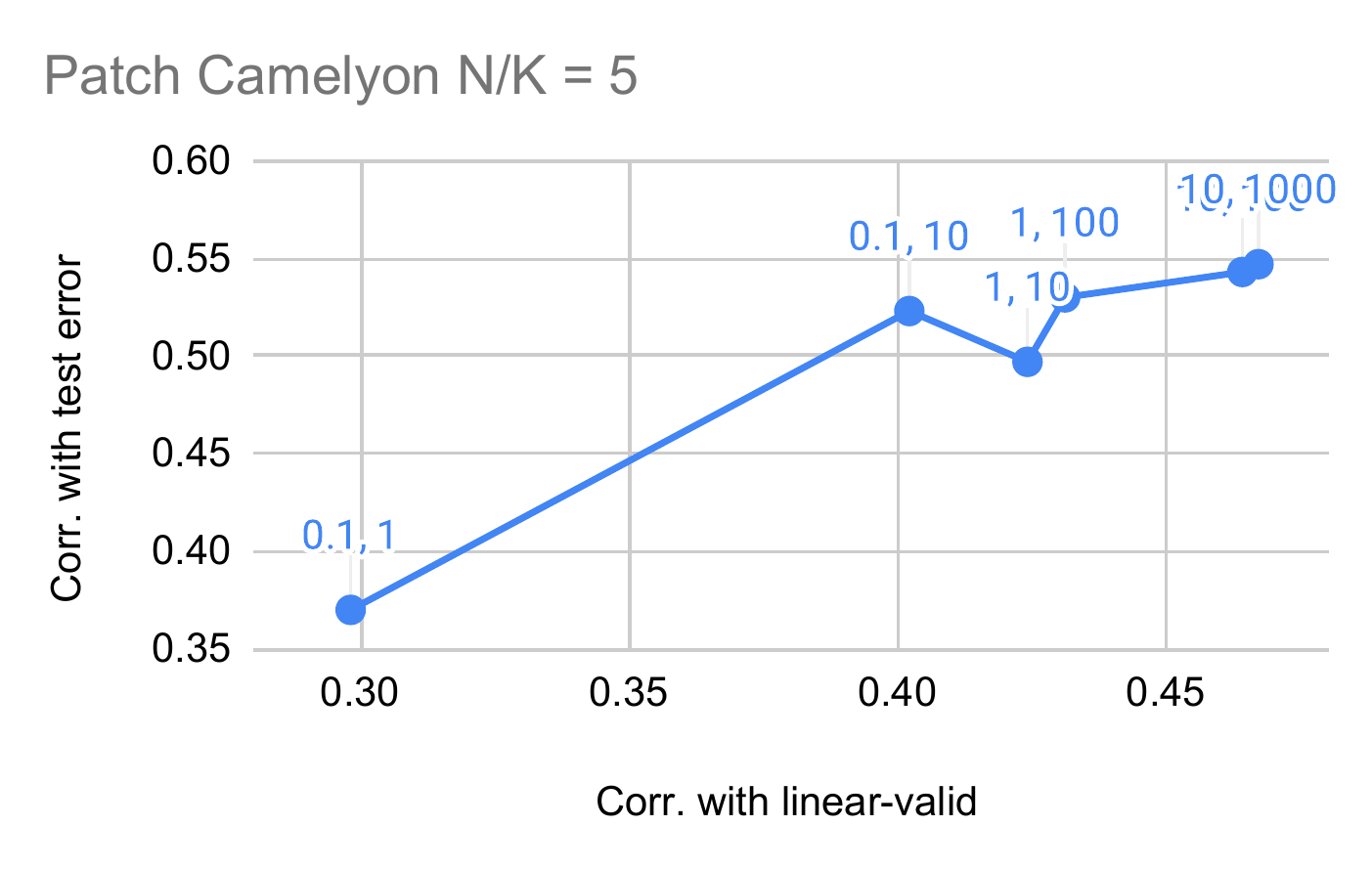}
    \includegraphics[width=0.45\textwidth]{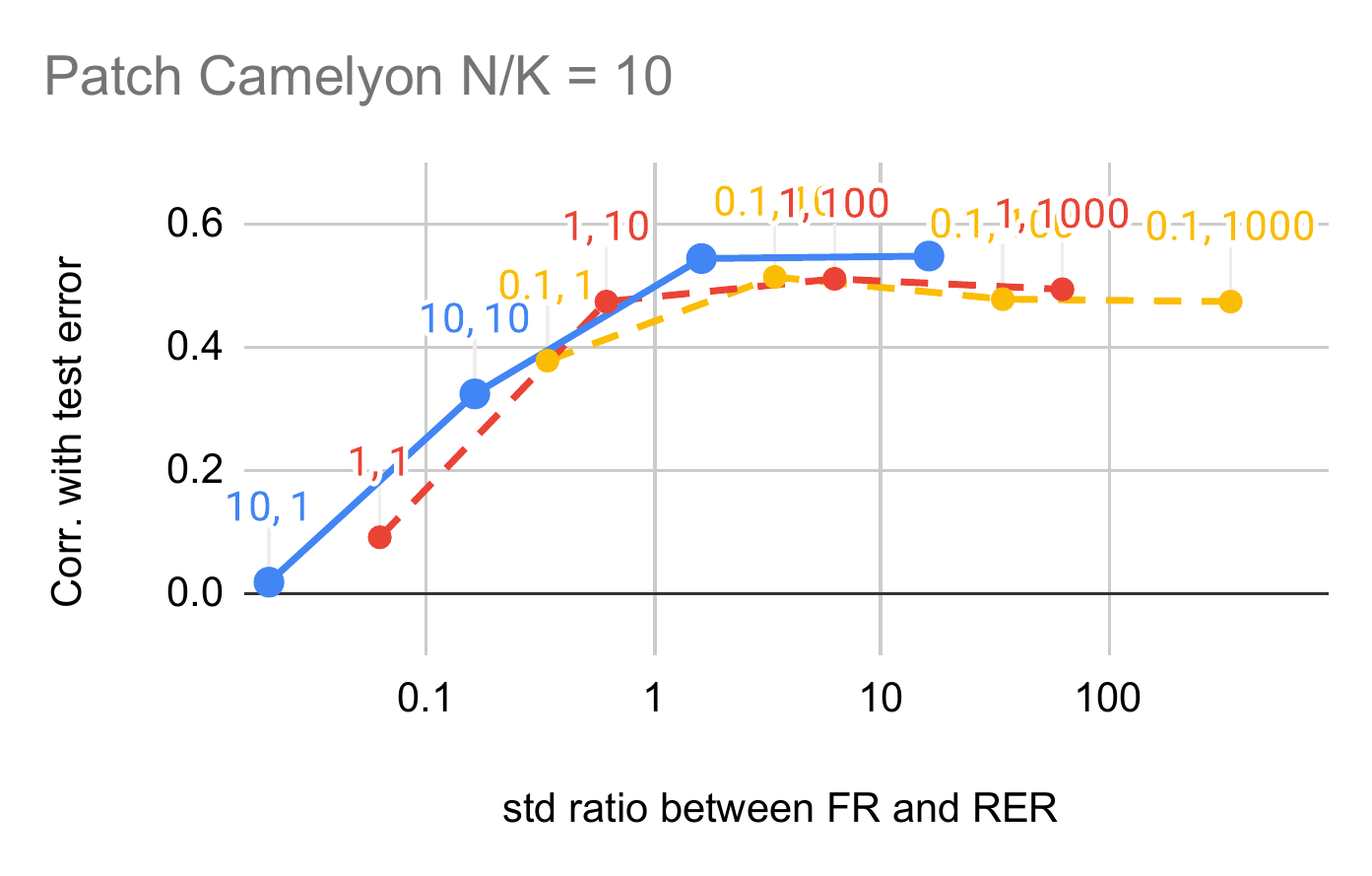}
    \includegraphics[width=0.45\textwidth]{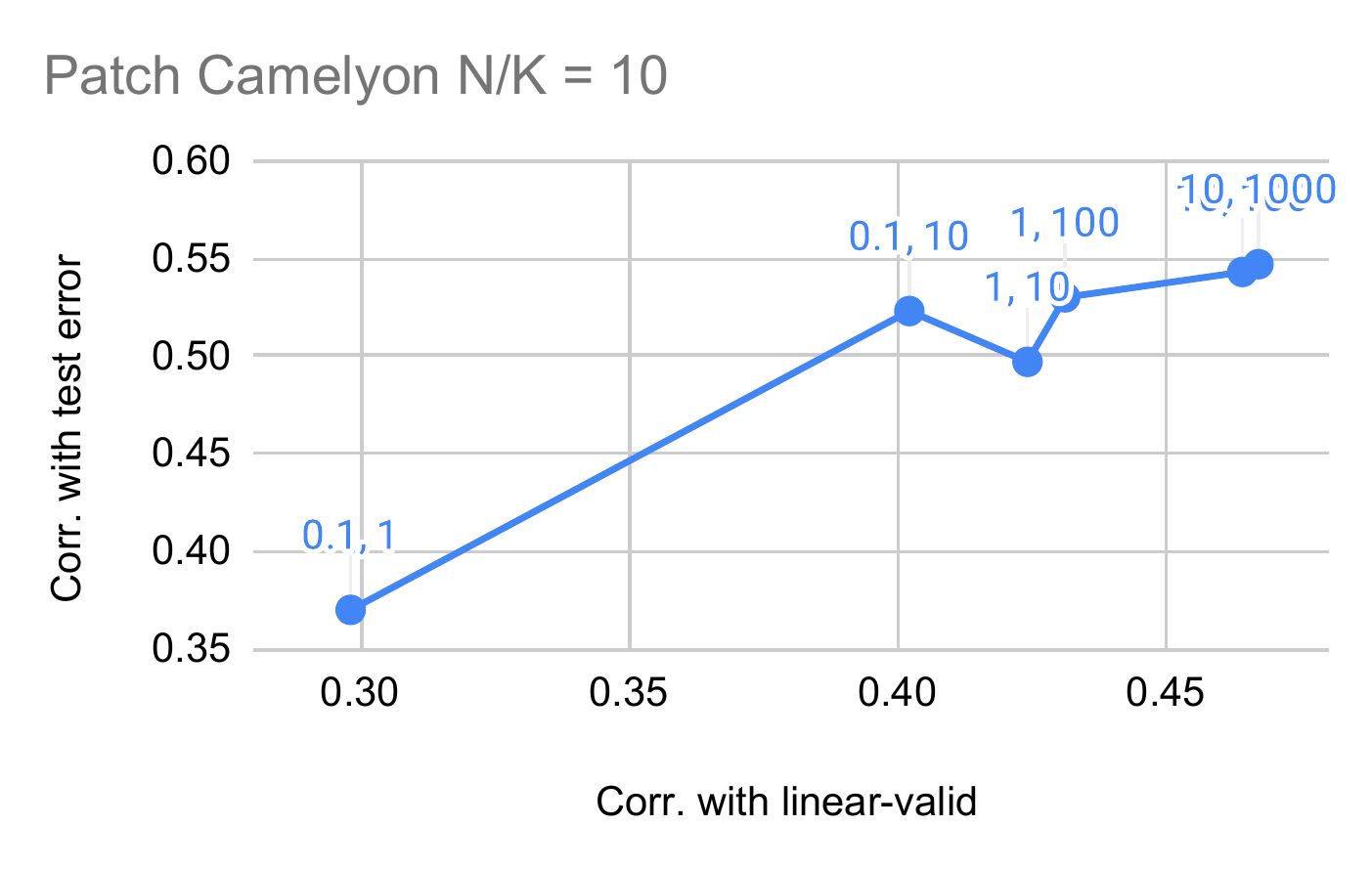}
    \caption{PACTran-Gaussian hyperparameter studies on Patch Camelyon. Hyperparameters are labeled as $(a, b)$. High $y$-value indicates a good correlation with the downstream test error.}
    \label{fig:hp_patch_camelyon}
\end{figure}

\begin{figure}[!ht]
    \centering
    \includegraphics[width=0.45\textwidth]{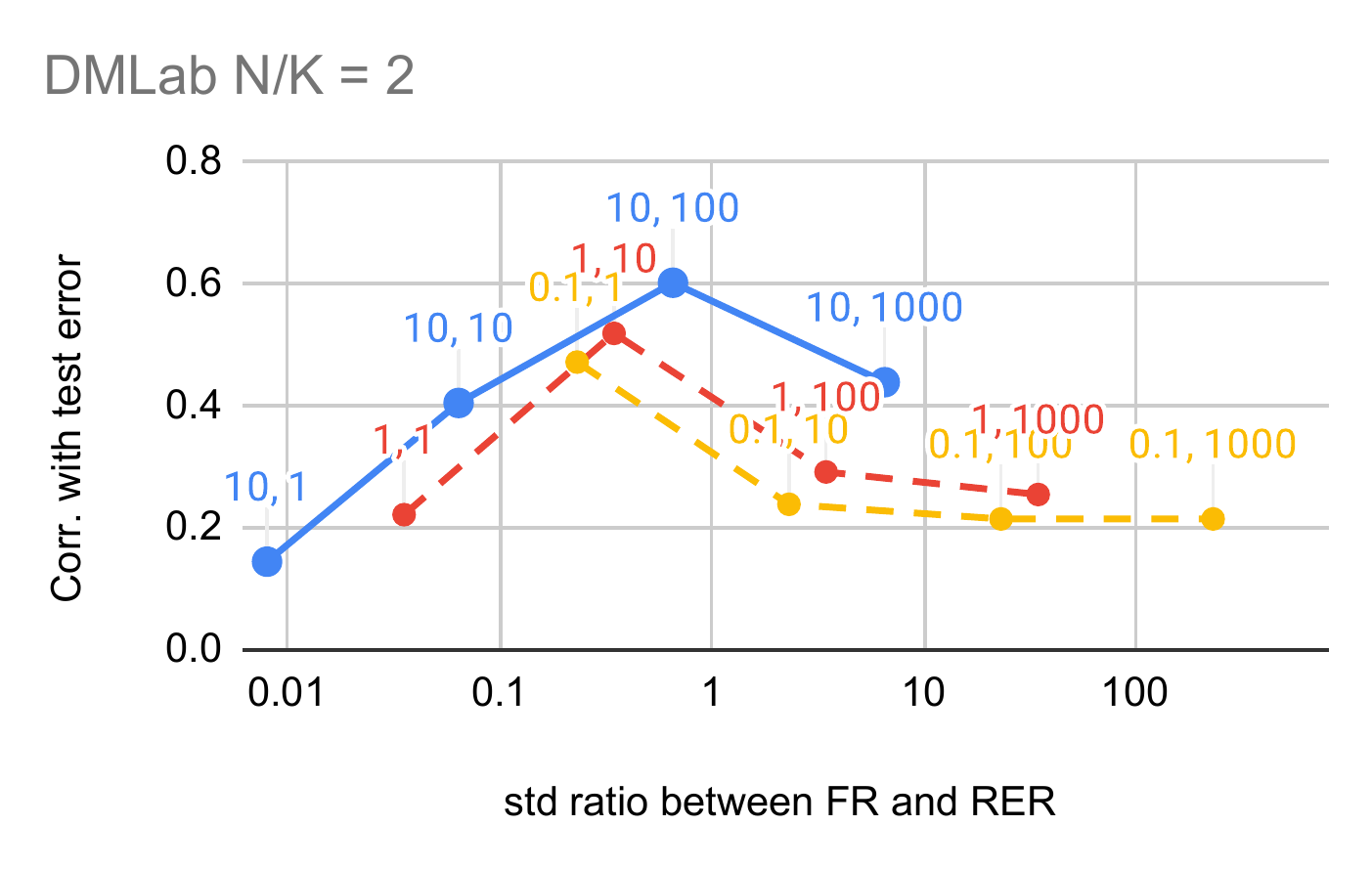}
    \includegraphics[width=0.45\textwidth]{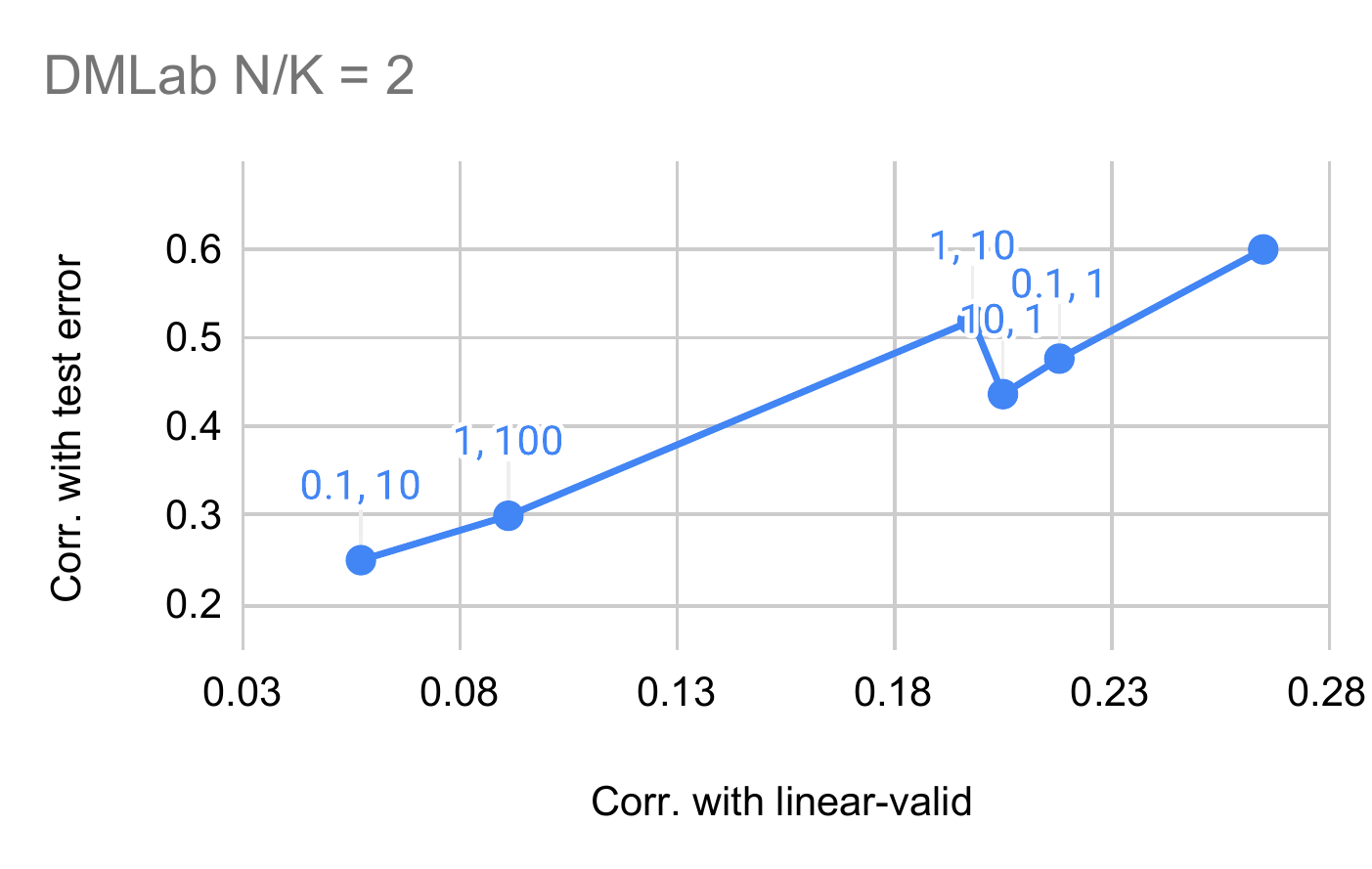}
    \includegraphics[width=0.45\textwidth]{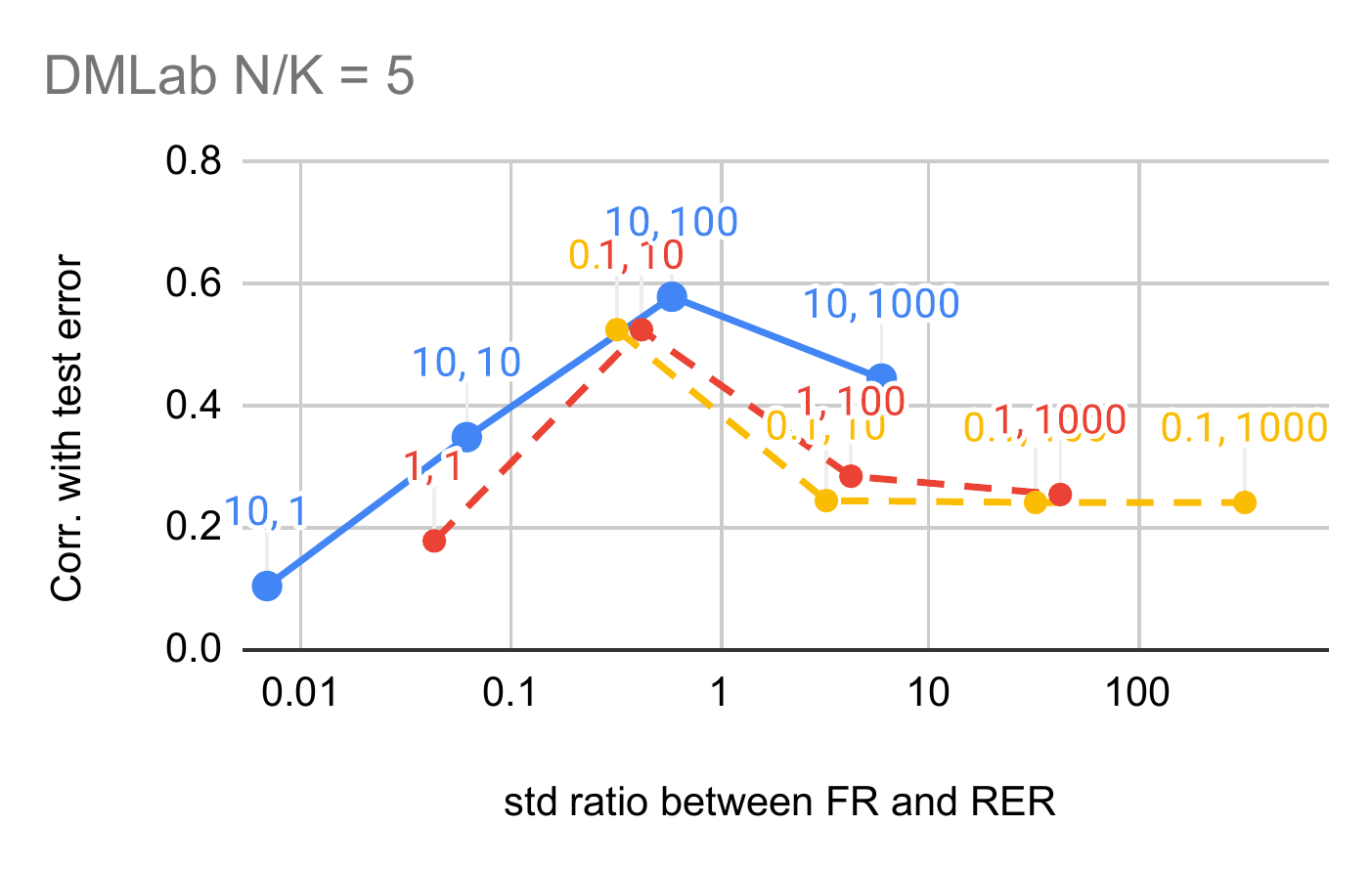}
    \includegraphics[width=0.45\textwidth]{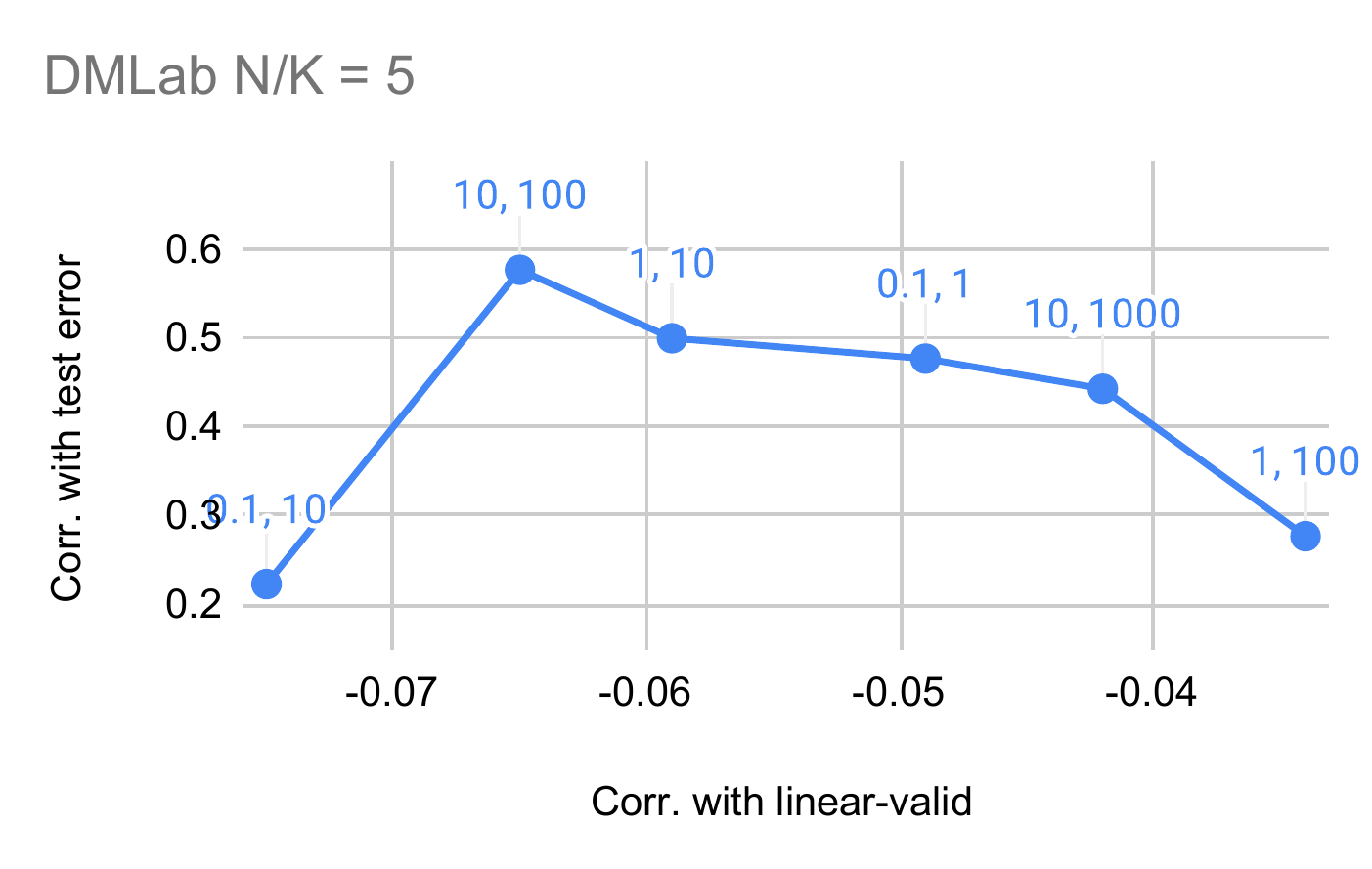}
    \includegraphics[width=0.45\textwidth]{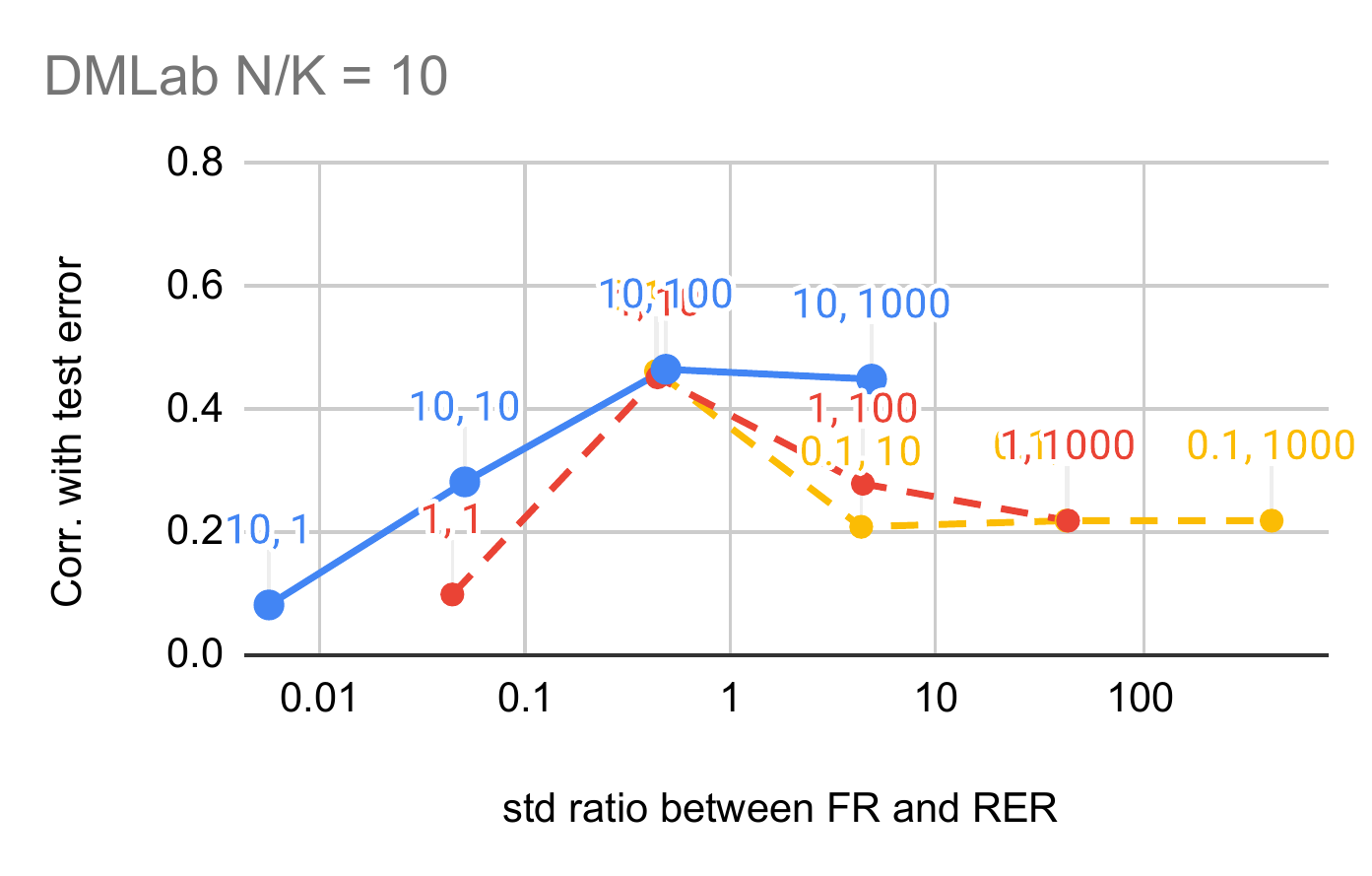}
    \includegraphics[width=0.45\textwidth]{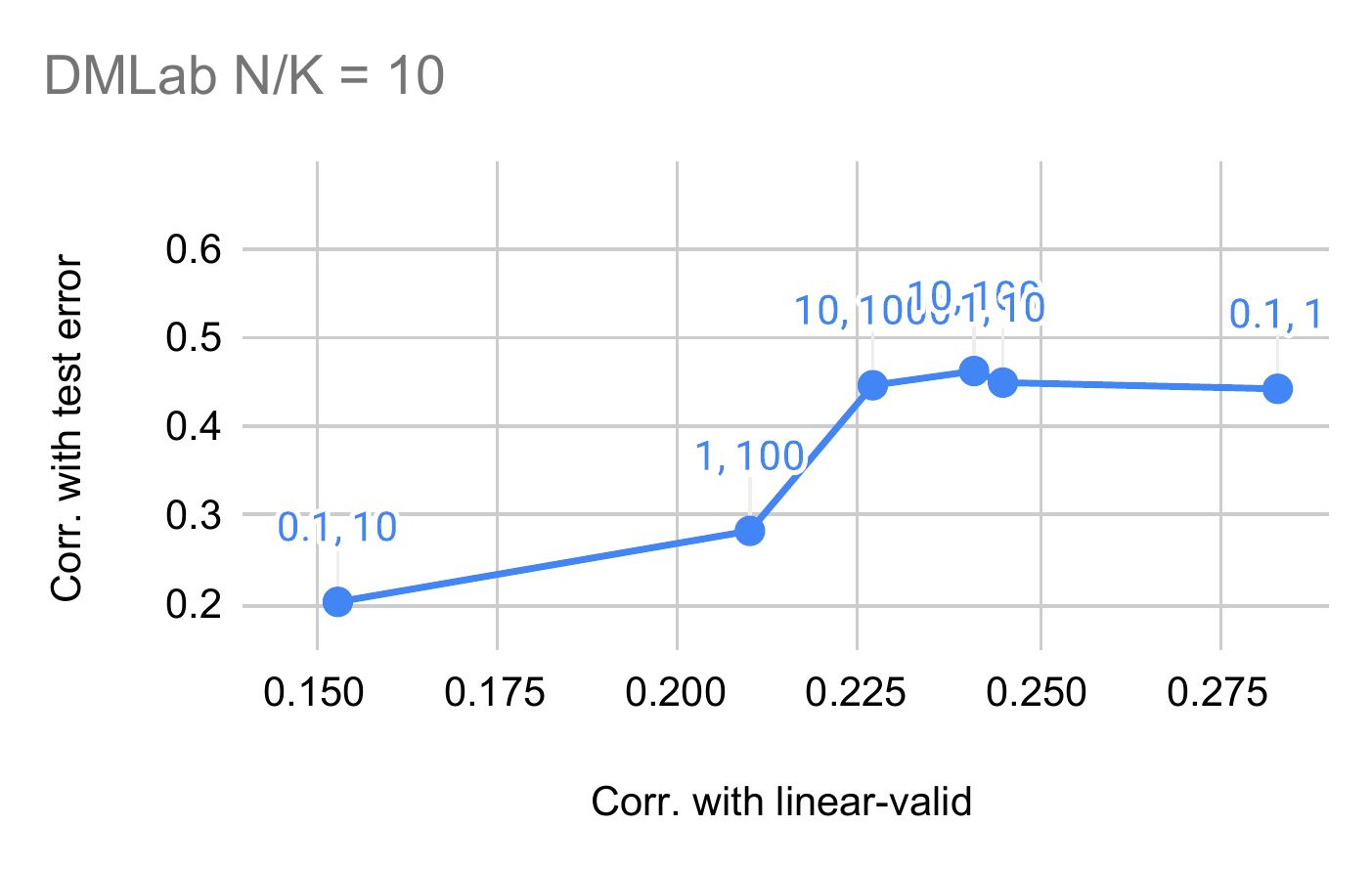}
    \caption{PACTran-Gaussian hyperparameter studies on DMLab. Hyperparameters are labeled as $(a, b)$. High $y$-value indicates a good correlation with the downstream test error.}
    \label{fig:hp_dmlab}
\end{figure}

\begin{figure}[!ht]
    \centering
    \includegraphics[width=0.45\textwidth]{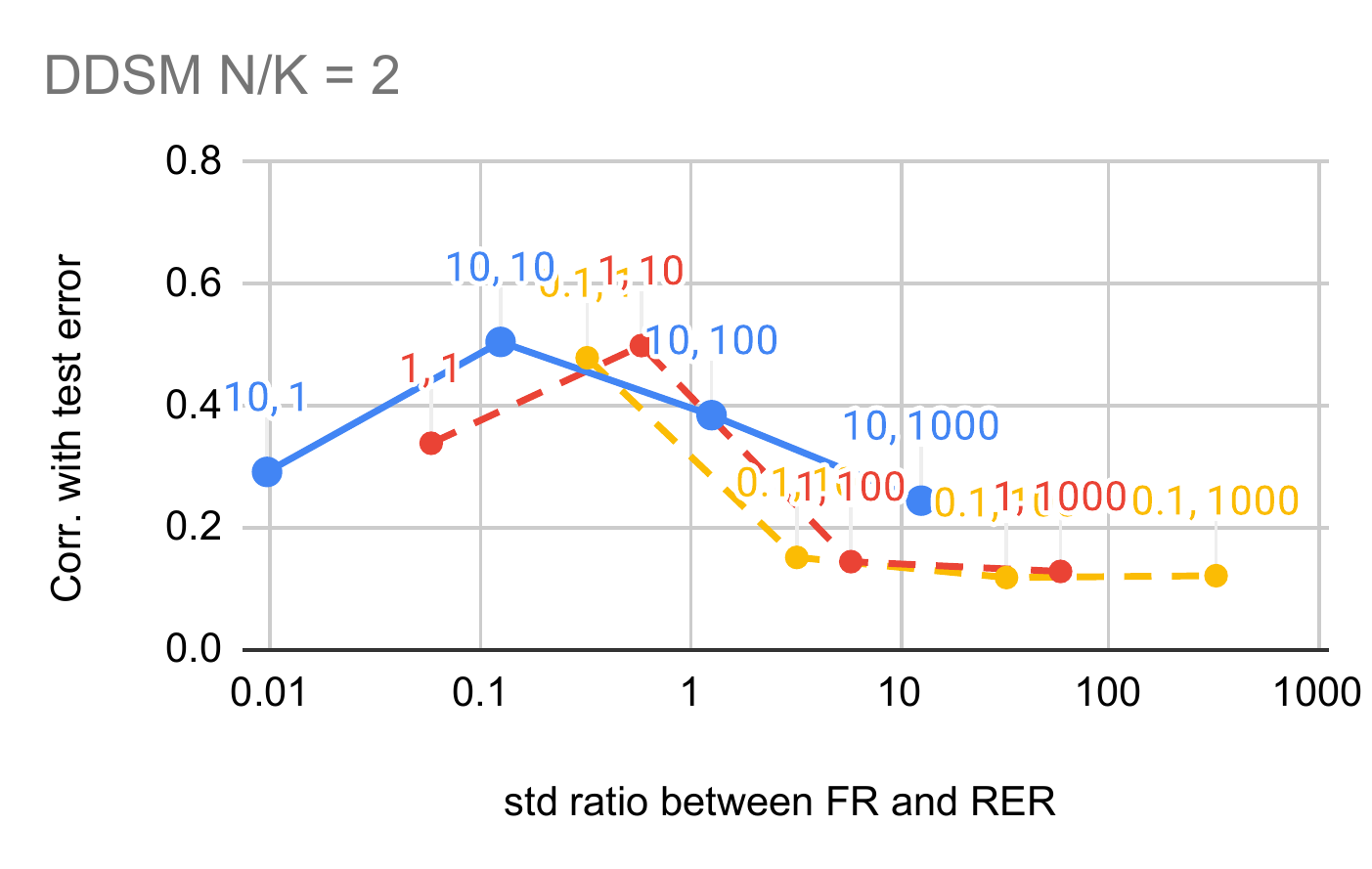}
    \includegraphics[width=0.45\textwidth]{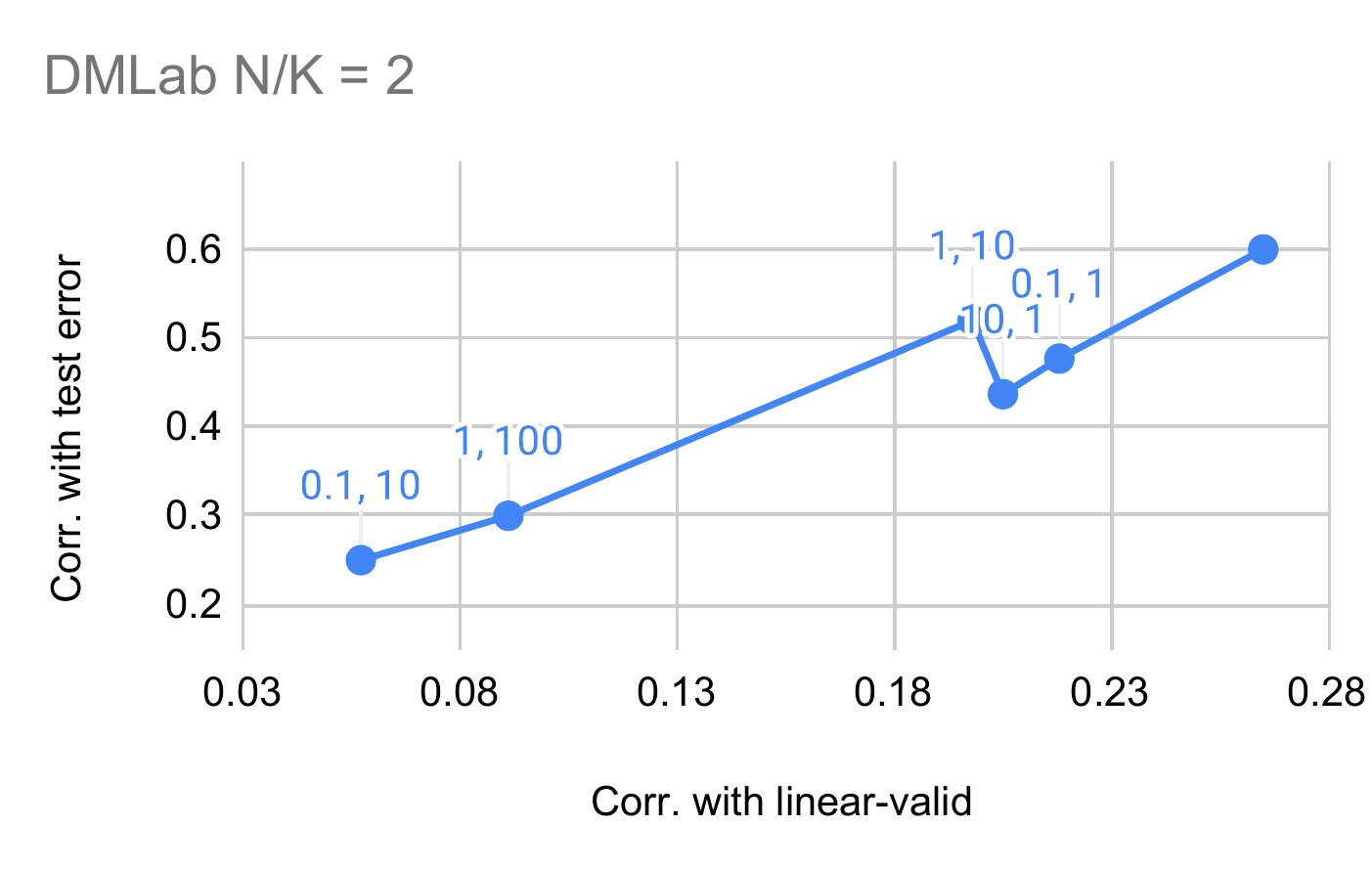}
    \includegraphics[width=0.45\textwidth]{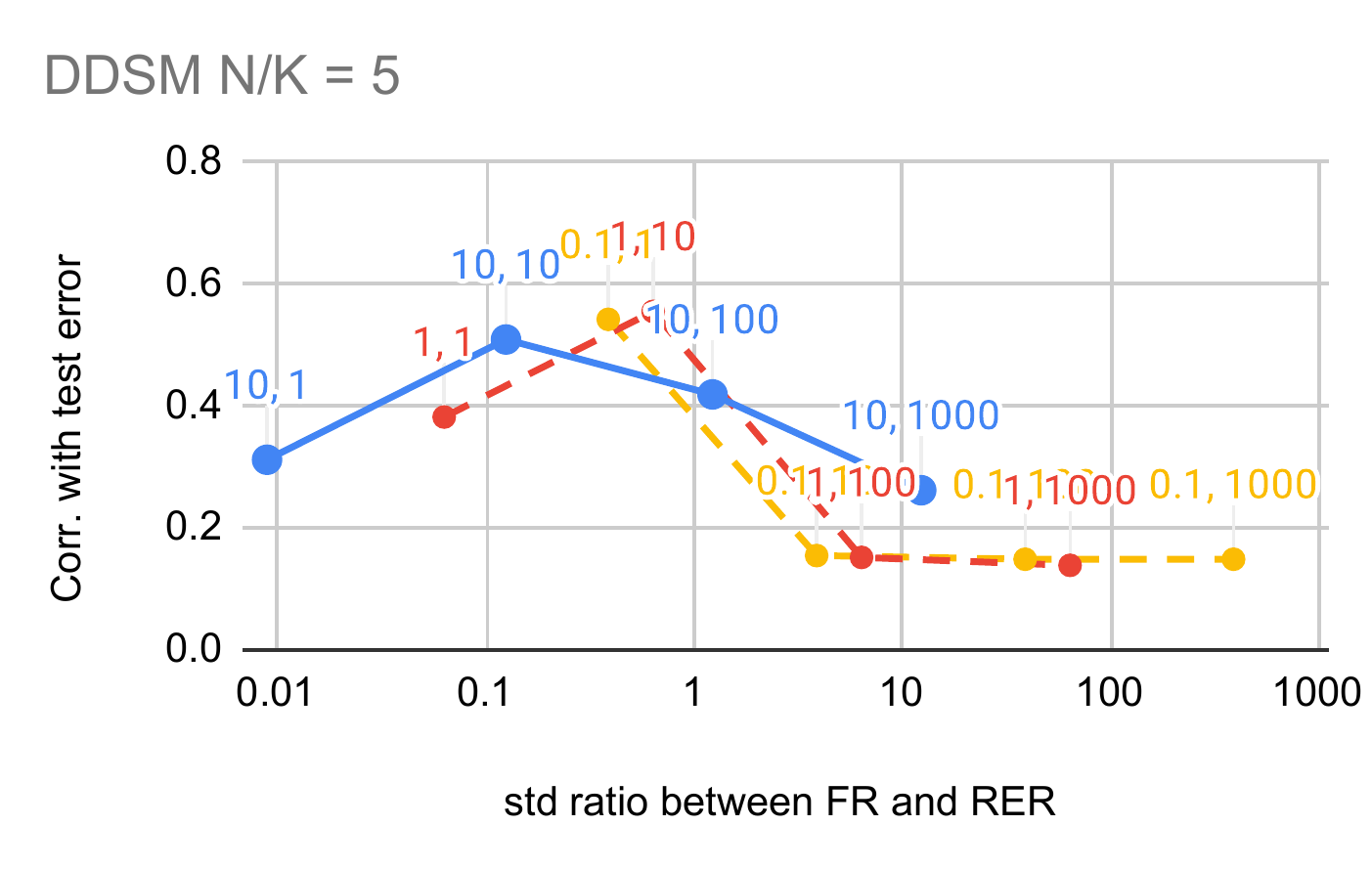}
    \includegraphics[width=0.45\textwidth]{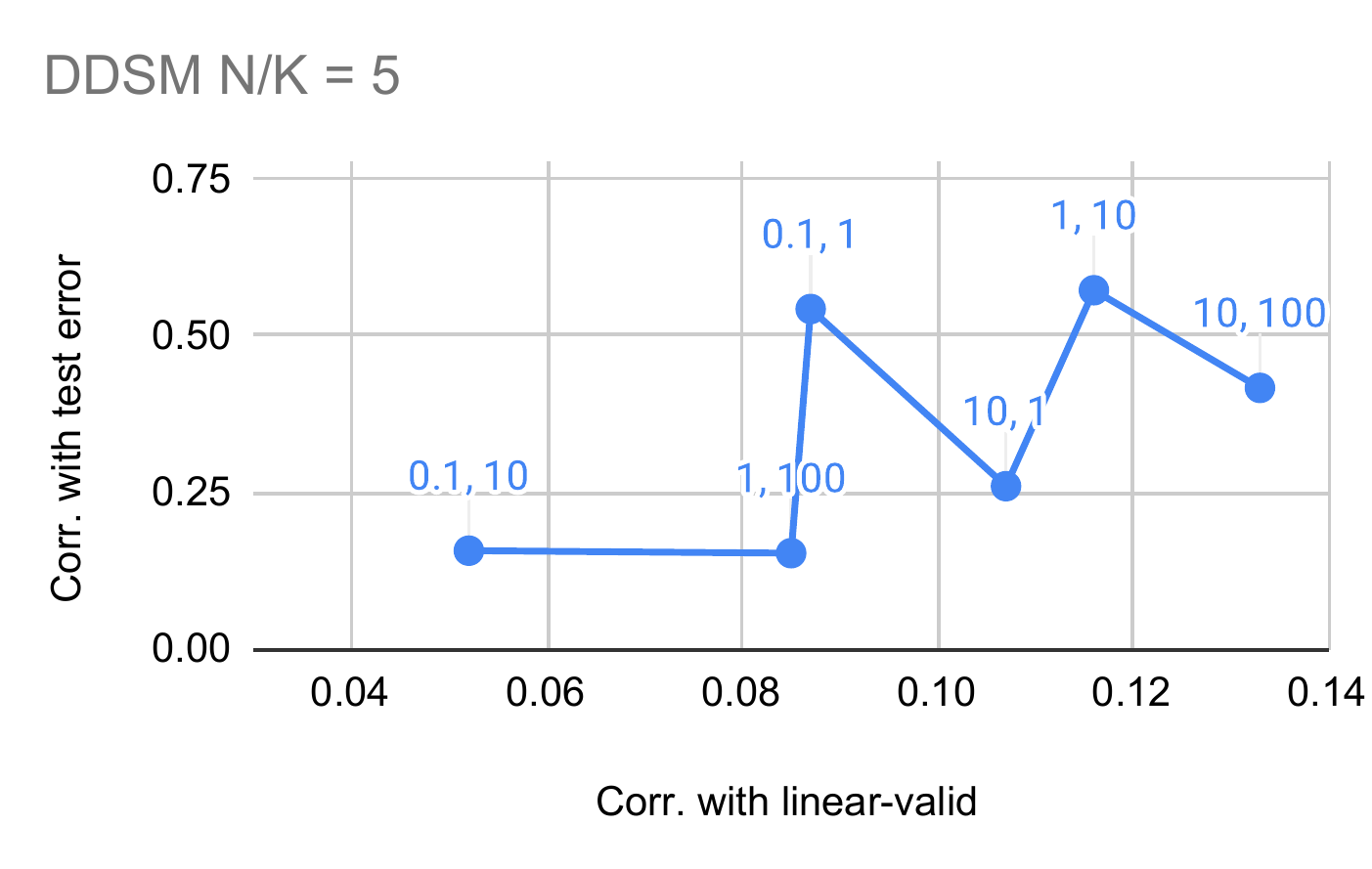}
    \includegraphics[width=0.45\textwidth]{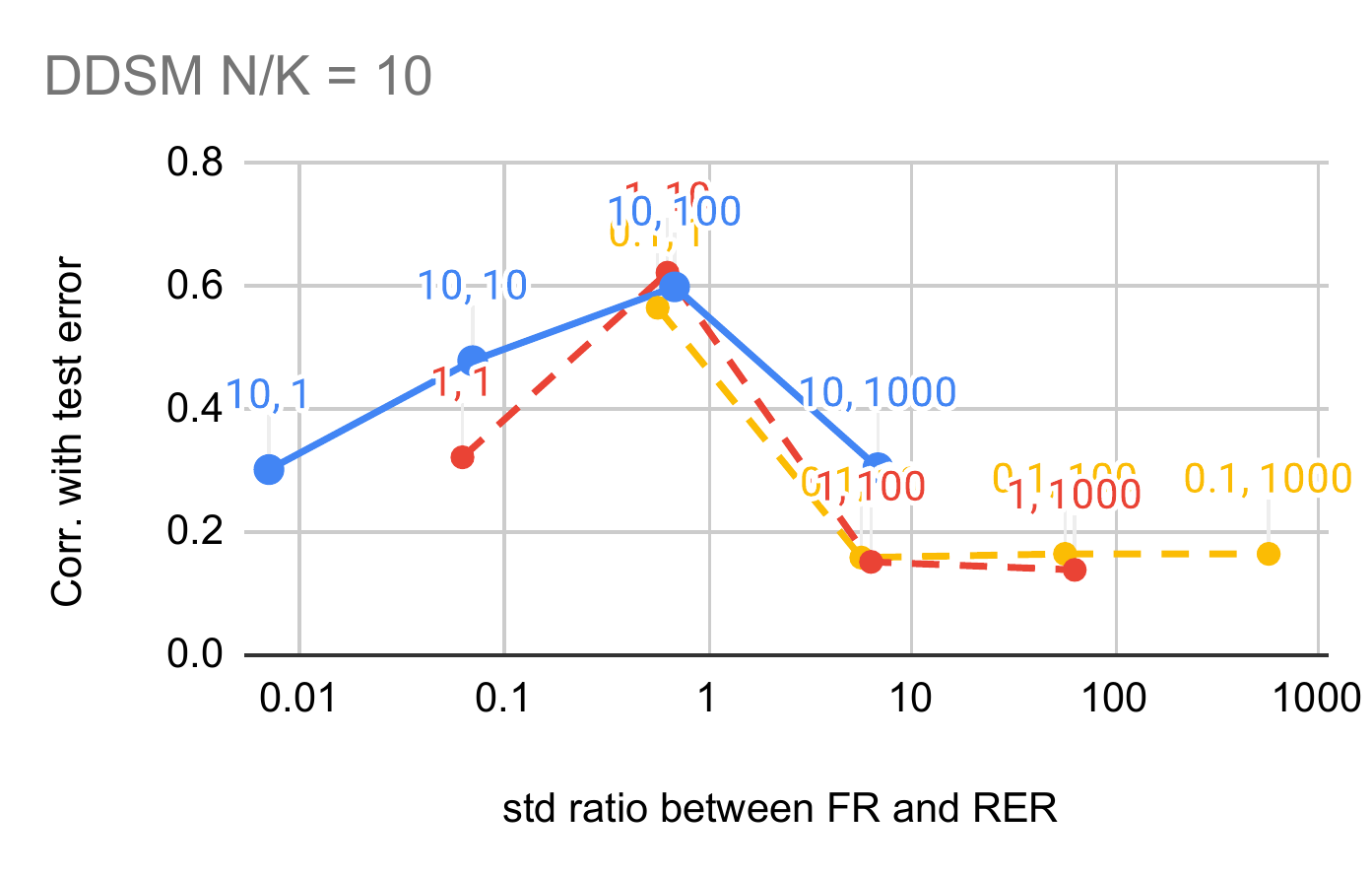}
    \includegraphics[width=0.45\textwidth]{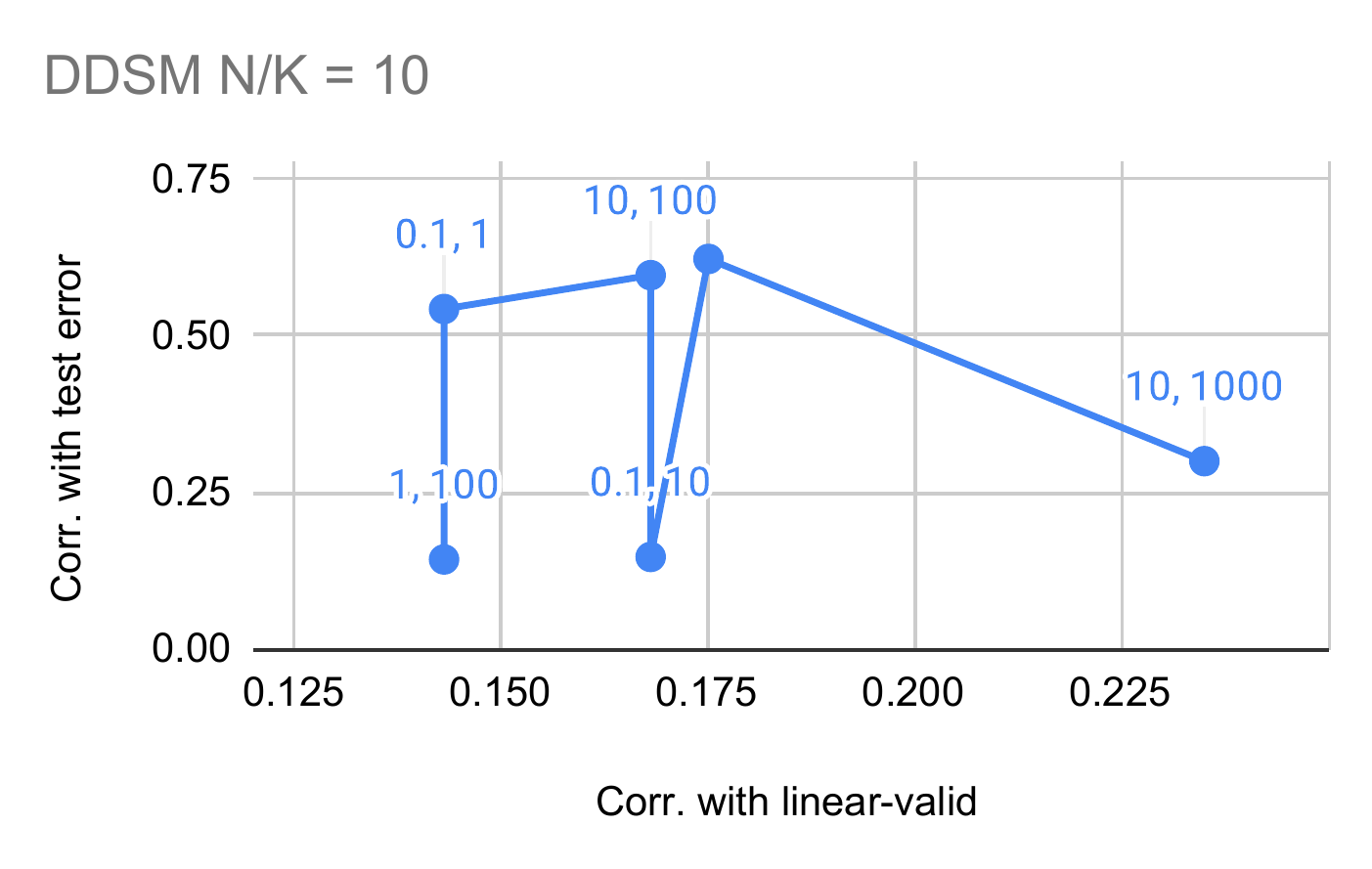}
    \caption{PACTran-Gaussian hyperparameter studies on DDSM. Hyperparameters are labeled as $(a, b)$. High $y$-value indicates a good correlation with the downstream test error.}
    \label{fig:hp_ddsm}
\end{figure}

In Fig.\ref{fig:hp_caltech101}-\ref{fig:hp_ddsm}, we plotted the performance of different hyperparameters, labeled as $(a, b)$, on the 9 VTAB tasks of the NeuCraB experiments. Each figure is composed of two columns and three rows (corresponding to $N/K \in \{2, 5, 10\}$). 
The left column plotted the ratio between the robust standard deviation of the FR and RER term ($x$-axis) vs. the Kendall-Tau between PT-Gauss and the downstream test error ($y$-axis). The right column plotted the Kendall-Tau between PT-Gauss and LINEAR-VALID ($x$-axis) vs. the Kendall-Tau between PT-Gauss and the downstream test error ($y$-axis). 

From the left columns, we find that the ratios between the standard deviation of the FR term and the RER term are very indicative of the performances of the PT-Gauss metric. Intuitively, too low of a ratio ($\le 0.1$) reduces PT-Gauss to the LINEAR metric, while too high of a ratio ($\ge 10$) completely ignores the RER term. Both scenarios are clearly not optimal according to the results, and the optimal ratio is consistently around 1.0 which achieves a balance between FR and RER. For better visualization, we group the hyperparameters pairs by the $a$ values using different colors (Yellow: $a=0.1$, Red: $a=1$, Blue: $a=10$).

From the right columns, we find that the Kendall-Tau between PT-Gauss and LINEAR-VALID is in general well correlated with the Kendall-Tau between PT-Gauss and the test error. This justifies our choice of using LINEAR-VALID as a validation method for choosing hyperparameters. The few exceptions (e.g. SmallNorb) are mostly caused by the poor performance of LINEAR-VALID on that dataset.

\subsection{Results on Each Individual Dataset}
\label{sec:vtab-complete}
In Table \ref{tab:vtab-complete2}, \ref{tab:vtab-complete5}, \ref{tab:vtab-complete10}, we report the complete results on each individual VTAB task.

\begin{table}[!ht]
\centering
  \begin{tabular}{c | c | c | c }
100+ classes & Caltech101 & Oxford-flowers & Sun397 \\\hline
LEEP & 0.253 $\pm$ 0.023 & 0.140 $\pm$ 0.016 & 0.213 $\pm$ 0.010  \\
$\Ncal$-LEEP & 0.747 $\pm$ 0.024 & 0.663 $\pm$ 0.026 & 0.760 $\pm$ 0.019\\
H-score & 0.327 $\pm$ 0.056 & 0.443 $\pm$ 0.035 & 0.470 $\pm$ 0.041\\
LogME & 0.350 $\pm$ 0.000 &0.293 $\pm$ 0.008& 0.280 $\pm$ 0.003\\
LINEAR &0.253 $\pm$ 0.010 & 0.203 $\pm$ 0.006 & 0.237 $\pm$ 0.006\\
LINEAR-VALID &0.778 $\pm$ 0.034 & 0.726 $\pm$ 0.025 & 0.746 $\pm$ 0.038\\ \hline
$\Ncal$-PT-Dir &0.787 $\pm$ 0.030 & 0.713 $\pm$ 0.018 & 0.780 $\pm$ 0.029\\
$\Ncal$-PT-Gam &0.790 $\pm$ 0.032 & 0.713 $\pm$ 0.013 & 0.780 $\pm$ 0.024\\
PT-Gauss$_{grid}$ &0.860 $\pm$ 0.014 & 0.913 $\pm$ 0.015 & 0.830 $\pm$ 0.010\\
PT-Gauss$_{fix}$ &0.800 $\pm$ 0.011 & 0.750 $\pm$ 0.020 & 0.760 $\pm$ 0.011\\\hline\hline
10-99 classes & Cifar-10 & Oxford-IIIT Pet & SmallNorb \\\hline
LEEP & -0.040 $\pm$ 0.035 &0.206 $\pm$ 0.008 & -0.150 $\pm$ 0.062 \\
$\Ncal$-LEEP & 0.419 $\pm$ 0.062 & 0.678 $\pm$ 0.017 & 0.107 $\pm$ 0.027\\
H-score & 0.005 $\pm$ 0.033 & 0.072 $\pm$ 0.057 & 0.242 $\pm$ 0.019\\
LogME & 0.153 $\pm$ 0.003 & 0.206 $\pm$ 0.006 & -0.157 $\pm$ 0.006\\
LINEAR & 0.160 $\pm$ 0.025 & 0.203 $\pm$ 0.003 & -0.147 $\pm$ 0.015\\
LINEAR-VALID & 0.311 $\pm$ 0.079 & 0.672 $\pm$ 0.027 & -0.055 $\pm$ 0.070\\ \hline
$\Ncal$-PT-Dir & 0.413 $\pm$ 0.101 & 0.678 $\pm$ 0.033 & -0.110 $\pm$ 0.032\\
$\Ncal$-PT-Gam & 0.420 $\pm$ 0.105& 0.678 $\pm$ 0.032 & -0.100 $\pm$ 0.040\\
PT-Gauss$_{grid}$ & 0.770 $\pm$ 0.025&0.775 $\pm$ 0.012 & 0.447 $\pm$ 0.037\\
PT-Gauss$_{fix}$ & 0.770 $\pm$ 0.030 & 0.832 $\pm$ 0.012 & 0.447 $\pm$ 0.037\\\hline\hline
2-9 classes & Patch-Camelyon & DMLAB & CBIS-DDSM \\\hline
LEEP & -0.024 $\pm$ 0.030 & -0.003 $\pm$ 0.037 & 0.150 $\pm$ 0.022 \\
$\Ncal$-LEEP &0.162 $\pm$ 0.039 &0.069 $\pm$ 0.088&-0.003 $\pm$ 0.085\\
H-score & 0.393 $\pm$ 0.056 & 0.260 $\pm$ 0.071&-0.097 $\pm$ 0.056\\
LogME & -0.123 $\pm$ 0.013 & 0.073 $\pm$ 0.006&0.263 $\pm$ 0.006\\
LINEAR & -0.043 $\pm$ 0.025 & 0.097 $\pm$ 0.018&0.287 $\pm$ 0.025\\
LINEAR-VALID & 0.294 $\pm$ 0.065 & 0.017 $\pm$ 0.097 & -0.123 $\pm$ 0.070\\ \hline
$\Ncal$-PT-Dir & 0.164 $\pm$ 0.054 & 0.027 $\pm$ 0.053&0.107 $\pm$ 0.062\\
$\Ncal$-PT-Gam & 0.177 $\pm$ 0.050 & 0.027 $\pm$ 0.052&0.120 $\pm$ 0.070\\
PT-Gauss$_{grid}$ & 0.543 $\pm$ 0.035 & 0.437 $\pm$ 0.037&0.383 $\pm$ 0.070\\
PT-Gauss$_{fix}$ & 0.543 $\pm$ 0.044 & 0.600 $\pm$ 0.032&0.383 $\pm$ 0.070\\\hline
  \end{tabular}
  \caption{Kendall-Tau correlations on each of the VTAB tasks when $N/K=2$.}
  \label{tab:vtab-complete2}
 \end{table}
 
\begin{table}[!ht]
\centering
  \begin{tabular}{c | c | c | c }
100+ classes & Caltech101 & Oxford-flowers & Sun397 \\\hline
LEEP & 0.270 $\pm$ 0.008 & 0.163 $\pm$ 0.007 & 0.237 $\pm$ 0.010 \\
$\Ncal$-LEEP & 0.803 $\pm$ 0.012 & 0.743 $\pm$ 0.011 & 0.840 $\pm$ 0.015\\
H-score & 0.503 $\pm$ 0.038&0.393 $\pm$ 0.056 & 0.340 $\pm$ 0.053\\
LogME & 0.450 $\pm$ 0.000 & 0.393 $\pm$ 0.004&0.420 $\pm$ 0.003\\
LINEAR & 0.277 $\pm$ 0.008 & 0.220 $\pm$ 0.006 & 0.263 $\pm$ 0.011\\
LINEAR-VALID & 0.804  $\pm$ 0.021 & 0.801 $\pm$ 0.022 & 0.816 $\pm$ 0.009\\ \hline
$\Ncal$-PT-Dir & 0.837 $\pm$ 0.018&0.793 $\pm$ 0.012 & 0.847 $\pm$ 0.017\\
$\Ncal$-PT-Gam & 0.840 $\pm$ 0.012 & 0.793 $\pm$ 0.009 & 0.843 $\pm$ 0.017\\
PT-Gauss$_{grid}$ & 0.823 $\pm$ 0.008 &0.800 $\pm$ 0.006 & 0.757 $\pm$ 0.004\\
PT-Gauss$_{fix}$ & 0.823 $\pm$ 0.008 & 0.877 $\pm$ 0.006 & 0.797 $\pm$ 0.006\\\hline\hline
10-99 classes & Cifar-10 & Oxford-IIIT Pet & SmallNorb \\\hline
LEEP & 0.073 $\pm$ 0.025 & 0.239 $\pm$ 0.009 & -0.067 $\pm$ 0.029 \\
$\Ncal$-LEEP &0.693 $\pm$ 0.039 & 0.815 $\pm$ 0.017 & 0.100 $\pm$ 0.058\\
H-score & 0.000 $\pm$ 0.065&0.239 $\pm$ 0.027 & 0.184 $\pm$ 0.050\\
LogME & 0.130 $\pm$ 0.003 & 0.296 $\pm$ 0.007 & -0.147 $\pm$ 0.006\\
LINEAR & 0.183 $\pm$ 0.007 & 0.219 $\pm$ 0.006 & -0.150 $\pm$ 0.007\\
LINEAR-VALID & 0.630 $\pm$ 0.033 &0.735 $\pm$ 0.033 & -0.133 $\pm$ 0.044\\ \hline
$\Ncal$-PT-Dir & 0.683 $\pm$ 0.024 & 0.751 $\pm$ 0.041 & -0.060 $\pm$ 0.059\\
$\Ncal$-PT-Gam & 0.683 $\pm$ 0.021 & 0.755 $\pm$ 0.036 & -0.053 $\pm$ 0.054\\
PT-Gauss$_{grid}$ & 0.820 $\pm$ 0.013 & 0.748 $\pm$ 0.006 & 0.580 $\pm$ 0.007\\
PT-Gauss$_{fix}$ & 0.820 $\pm$ 0.009 & 0.808 $\pm$ 0.006&0.397 $\pm$0.017 \\\hline\hline
2-9 classes & Patch-Camelyon & DMLAB & CBIS-DDSM \\\hline
LEEP & -0.090 $\pm$ 0.030 & 0.023 $\pm$ 0.037 & 0.137 $\pm$ 0.022 \\
$\Ncal$-LEEP &0.162 $\pm$ 0.039&-0.007 $\pm$ 0.100 & 0.133 $\pm$ 0.047\\
H-score & 0.393 $\pm$ 0.056&0.032 $\pm$ 0.102 & -0.072 $\pm$ 0.062\\
LogME & -0.123 $\pm$ 0.013 &0.070 $\pm$ 0.003&0.277 $\pm$ 0.007\\
LINEAR & -0.043 $\pm$ 0.025&0.110 $\pm$ 0.006&0.300 $\pm$ 0.010\\
LINEAR-VALID & 0.294 $\pm$ 0.065 & -0.109 $\pm$ 0.118 & -0.054 $\pm$ 0.043\\ \hline
$\Ncal$-PT-Dir & 0.164 $\pm$ 0.054&0.143 $\pm$ 0.094&0.113 $\pm$ 0.071\\
$\Ncal$-PT-Gam & 0.177 $\pm$ 0.050&0.157 $\pm$ 0.097&0.120 $\pm$ 0.068\\
PT-Gauss$_{grid}$ & 0.543 $\pm$ 0.035&0.277 $\pm$ 0.027&0.417 $\pm$ 0.060\\
PT-Gauss$_{fix}$ & 0.543 $\pm$ 0.044&0.577 $\pm$ 0.022&0.417 $\pm$ 0.060\\\hline
  \end{tabular}
  \caption{Kendall-Tau correlations on each of the VTAB tasks when $N/K=5$.}
  \label{tab:vtab-complete5}
 \end{table}  

\begin{table}[!ht]
\centering
  \begin{tabular}{c | c | c | c }
100+ classes & Caltech101 & Oxford-flowers & Sun397 \\\hline
LEEP & 0.320 $\pm$ 0.009 & 0.217 $\pm$ 0.016 & 0.290 $\pm$ 0.014 \\
$\Ncal$-LEEP &0.823 $\pm$ 0.010 & 0.793 $\pm$ 0.008 & 0.850 $\pm$ 0.011\\
H-score &0.497 $\pm$ 0.040 & 0.417 $\pm$ 0.042 & 0.470 $\pm$ 0.043\\
LogME &0.513 $\pm$ 0.003 &0.467 $\pm$ 0.000 & 0.483 $\pm$ 0.000\\
LINEAR &0.327 $\pm$ 0.007 & 0.277 $\pm$ 0.004 & 0.370 $\pm$ 0.006\\
LINEAR-VALID &0.821 $\pm$ 0.015 & 0.827 $\pm$ 0.010 & 0.857 $\pm$ 0.014\\ \hline
$\Ncal$-PT-Dir &0.827 $\pm$ 0.010&0.810 $\pm$ 0.006 & 0.880 $\pm$ 0.011\\
$\Ncal$-PT-Gam &0.827 $\pm$ 0.010 & 0.810 $\pm$ 0.006 & 0.880 $\pm$ 0.011\\
PT-Gauss$_{grid}$ &0.813 $\pm$ 0.009 & 0.767 $\pm$ 0.004& 0.727 $\pm$ 0.008\\
PT-Gauss$_{fix}$ &0.787 $\pm$ 0.009 & 0.820 $\pm$ 0.009&0.727 $\pm$ 0.006\\\hline\hline
10-99 classes & Cifar-10 & Oxford-IIIT Pet & SmallNorb \\\hline
LEEP & 0.113 $\pm$ 0.022 &0.226 $\pm$ 0.013 & -0.103 $\pm$ 0.026 \\
$\Ncal$-LEEP &0.730 $\pm$ 0.012&0.822 $\pm$ 0.023&0.007 $\pm$ 0.049\\
H-score & 0.234 $\pm$ 0.041 &0.296 $\pm$ 0.035&0.425 $\pm$ 0.060\\
LogME & 0.133 $\pm$ 0.000 &0.413 $\pm$ 0.004 & -0.133 $\pm$ 0.000\\
LINEAR & 0.180 $\pm$ 0.015&0.256 $\pm$ 0.006&-0.170 $\pm$ 0.009\\
LINEAR-VALID & 0.704 $\pm$ 0.035&0.811 $\pm$ 0.019 & -0.070 $\pm$ 0.041\\ \hline
$\Ncal$-PT-Dir & 0.690 $\pm$ 0.022&0.832 $\pm$ 0.013&-0.183 $\pm$ 0.051\\
$\Ncal$-PT-Gam & 0.700 $\pm$ 0.018&0.835 $\pm$ 0.010&-0.180 $\pm$ 0.052\\
PT-Gauss$_{grid}$ & 0.700 $\pm$ 0.017&0.778 $\pm$ 0.011&0.557 $\pm$ 0.008\\
PT-Gauss$_{fix}$ & 0.757 $\pm$ 0.014&0.808 $\pm$ 0.006 & 0.263 $\pm$ 0.007\\\hline\hline
2-9 classes & Patch-Camelyon & DMLAB & CBIS-DDSM \\\hline
LEEP & -0.090 $\pm$ 0.030 & 0.090 $\pm$ 0.037 & 0.147 $\pm$ 0.022 \\
$\Ncal$-LEEP &0.162 $\pm$ 0.039 &0.223 $\pm$ 0.027&0.060 $\pm$ 0.150\\
H-score & 0.393 $\pm$ 0.056&-0.057 $\pm$ 0.056&0.138 $\pm$ 0.042\\
LogME & -0.123 $\pm$ 0.013 &0.070 $\pm$ 0.003&0.273 $\pm$ 0.006\\
LINEAR & -0.043 $\pm$ 0.025 & 0.080 $\pm$ 0.006&0.290 $\pm$ 0.014\\
LINEAR-VALID & 0.294 $\pm$ 0.065&0.150 $\pm$ 0.105&-0.075 $\pm$ 0.126\\ \hline
$\Ncal$-PT-Dir & 0.164 $\pm$ 0.054&0.220 $\pm$ 0.050&0.017 $\pm$ 0.124\\
$\Ncal$-PT-Gam & 0.177 $\pm$ 0.050&0.217 $\pm$ 0.051&0.027 $\pm$ 0.126\\
PT-Gauss$_{grid}$ & 0.543 $\pm$ 0.035&0.443 $\pm$ 0.036&0.300 $\pm$ 0.040\\
PT-Gauss$_{fix}$ & 0.543 $\pm$ 0.044&0.463 $\pm$ 0.021&0.597 $\pm$ 0.033\\\hline
  \end{tabular}
  \caption{Kendall-Tau correlations on each of the VTAB tasks when $N/K=10$.}
  \label{tab:vtab-complete10}
 \end{table}

\subsection{Results on Checkpoints with the Same Feature Dimension}
\label{sec:vtab-same-dim}
The sixteen model checkpoints in the Neural Checkpoint Ranking Benchmark (NeuCRaB) have different penultimate feature dimensions. In particular, ten of them have 2048 dimensions: Jigsaw, Relative Patch Location, Exemplar, Rotation, Sup-Rotation, Semi-Rotation-10\%, Semi-Exemplar-10\%, Sup-Exemplar-100\%, Sup-100\%, and Feature Vector; two of them have 1536 dimensions: Cond-BigGAN and Uncond-BigGAN; and four of them have 128 dimensions: WAE-UKL, WAE-GAN, WAE-MMD, VAE.

In the 9 VTAB tasks that have been considered, the dimensionalities of the penultimate features appear to be positively correlated to the performance of the checkpoints (where checkpoints with higher feature dimensions tend to achieve higher testing accuracies). In fact, if the feature dimension is directly used as the metric (in which we need to add a small random perturbation to break the ties for those checkpoints with the same feature dimensions), the averaged Kendall-Tau correlation on the 9 tasks appears to be 0.481, which is higher than most of the other baselines. 

In order to eliminate the effect caused by the differences of the penultimate feature dimensions, we compare all metrics on a subset that contains the ten checkpoints with the same feature dimensions 2048. The rest of the experiment settings are the same as before. The results of their averaged performance on the 9 VTAB tasks are reported in Table \ref{tab:vtab-same-dim}. We can see that PT-Gauss still achieves the highest correlations compared to any other metrics on those 10 checkpoints with the same feature dimensions. 
 \begin{table}[!ht]
\centering
  \begin{tabular}{c | c | c | c }
 & K=2 & K=5 & K=10  \\\hline
LEEP & 0.293 & 0.301 & 0.311\\
$\Ncal$-LEEP & 0.275 &0.435 &0.456  \\
Hscore & -0.003 &-0.096 &-0.059 \\
LogME & 0.335 & 0.360 & 0.385 \\
LINEAR & 0.323 & 0.346 & 0.359 \\
LINEAR-VALID & 0.301 & 0.389  & 0.413 \\ \hline
$\Ncal$-PT-Dir &0.351 &0.432 &0.475  \\
$\Ncal$-PT-Gam &0.349 &0.429 &0.472 \\
PT-Gauss$_{grid}$ & 0.414 & 0.511 & 0.495 \\
PT-Gauss$_{fix}$ & {\bf 0.433} & {\bf 0.521} & {\bf 0.527} \\ \hline
  \end{tabular}
  \caption{Averaged Kendall-Tau correlations over the 9 VTAB tasks on the 10 checkpoints with the same feature dimensions 2048.}
  \label{tab:vtab-same-dim}
 \end{table}

\subsection{Kendall-Tau Rank Correlation}
\begin{figure}[t]
  \centering  \includegraphics[width=0.3\linewidth]{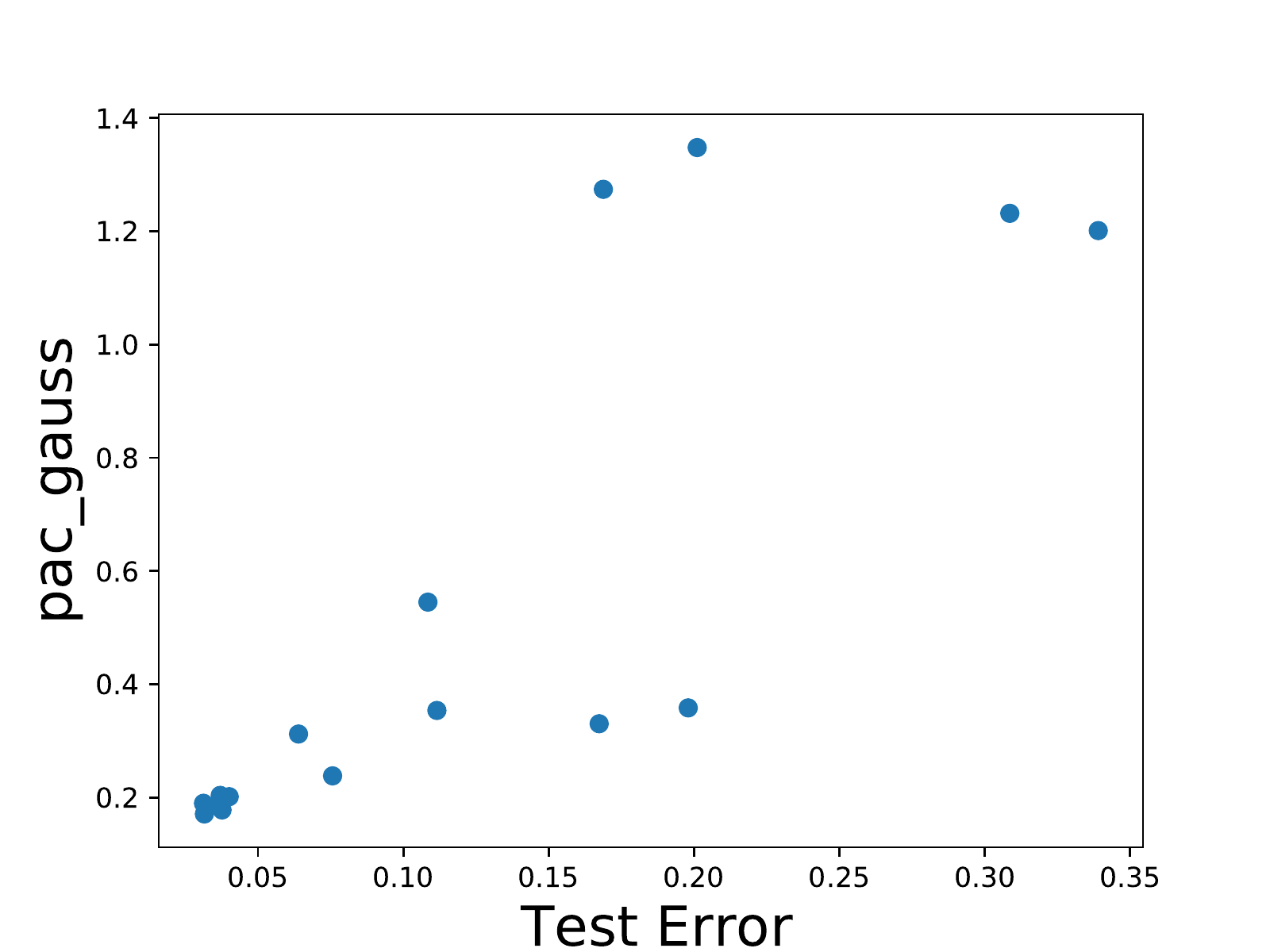}
  \includegraphics[width=0.3\linewidth]{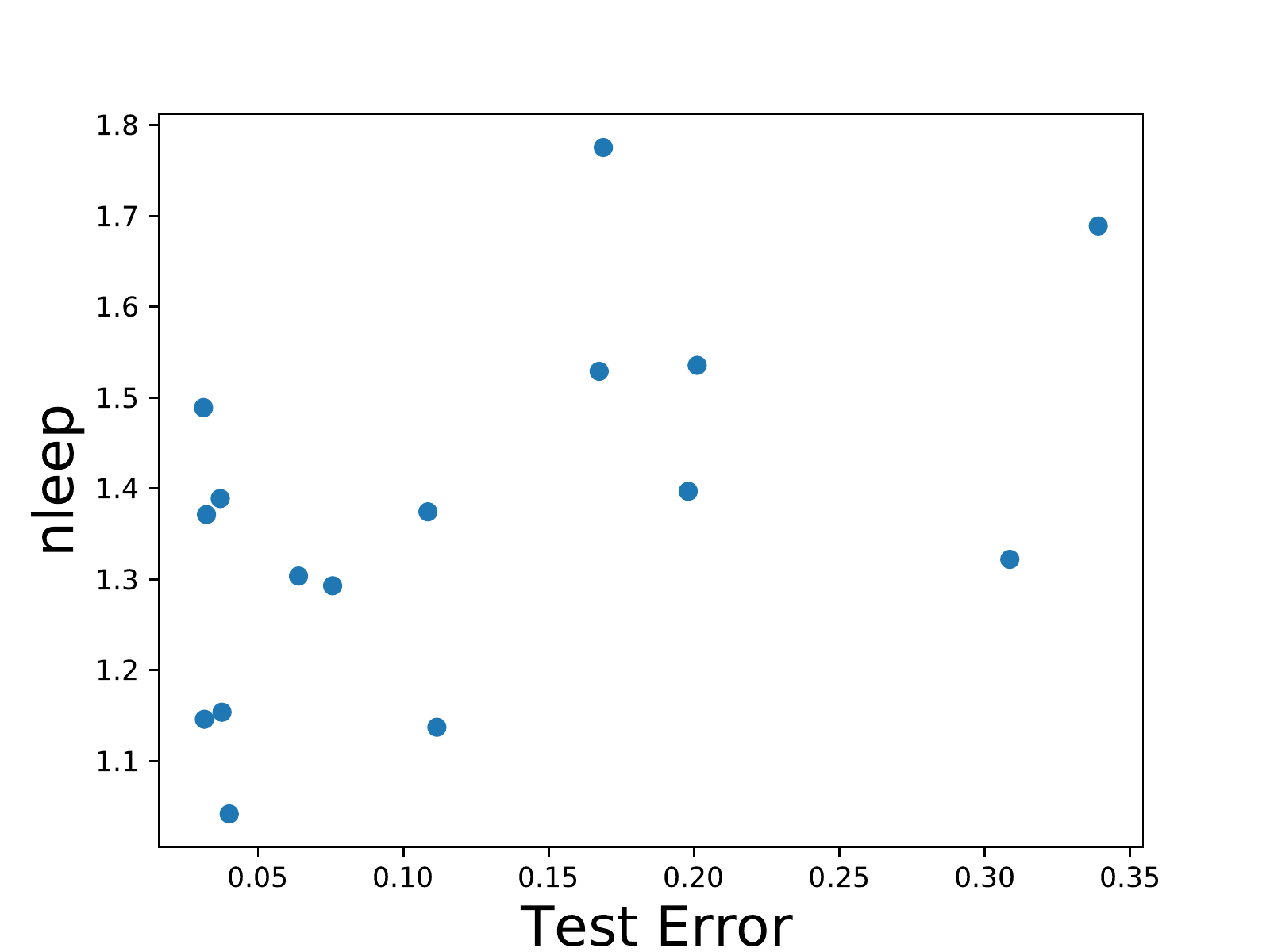}
  \includegraphics[width=0.3\linewidth]{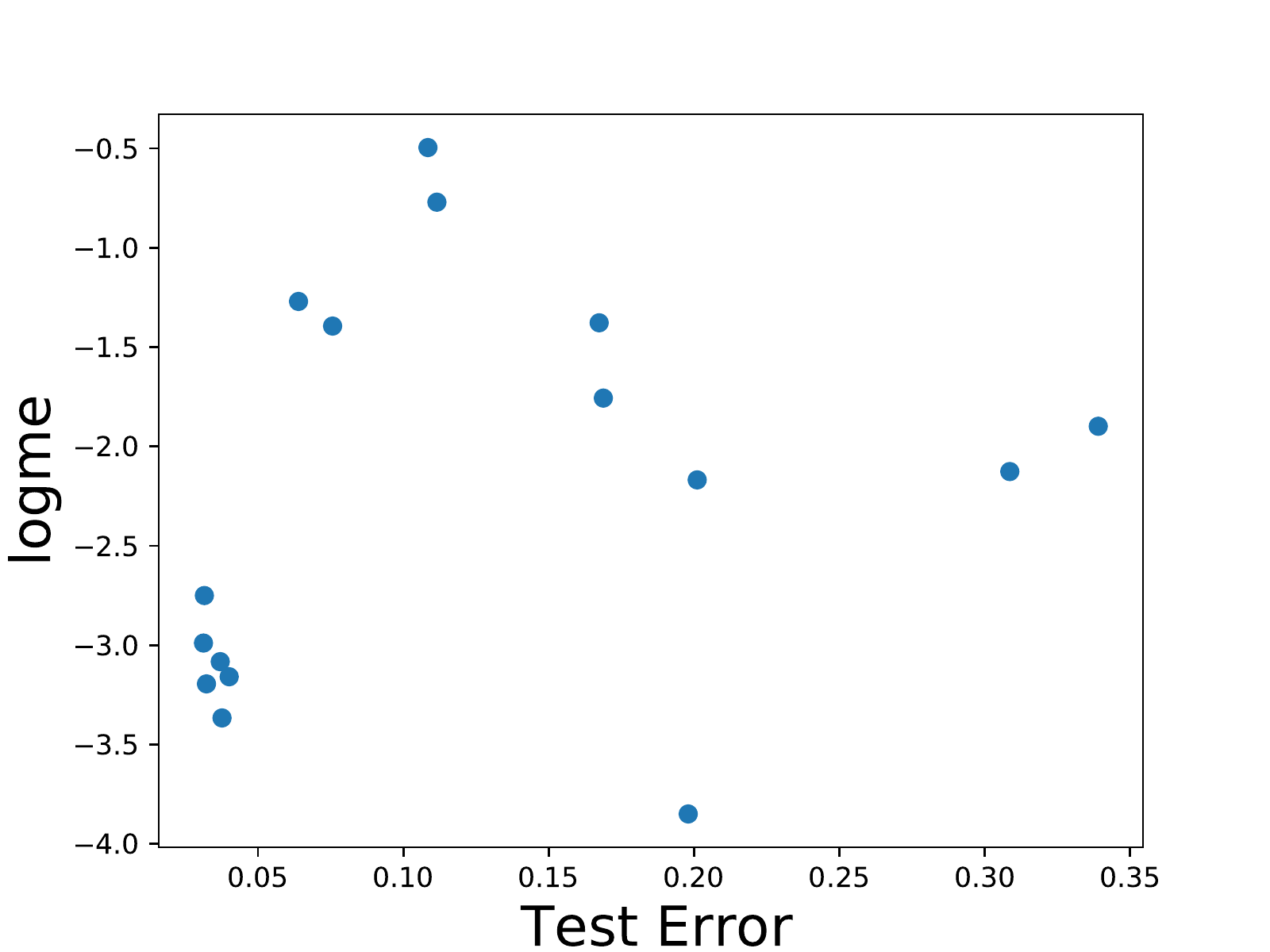}
   \caption{Test error vs PT-Gauss ($\tau$=0.77) , N-LEEP ($\tau$=0.42), LogME ($\tau$=0.15) of 16 checkpoints on the Cifar-10 task.}
   \label{fig:correlation}
\end{figure}

We use the Kendall-Tau rank correlation coefficient to correlate between the transferability metric scores and the testing error of the finetuned checkpoints. 
Kendall-Tau correlation is a classic metric for estimating the correlation between rankings and has been used in previous similar works such as \cite{li2021ranking,you2021logme,48754}. In particular, among $C$ checkpoints $\thetab_i$ with test error $e(\thetab_i)$ and a metric $m(\thetab_i)$:
\begin{align*} 
    \tau = \frac{1}{C(C-1)} \sum_{i \neq j} \sign(m(\thetab_i) - m(\thetab_j))\sign(e(\thetab_i) - e(\thetab_j)).
\end{align*}
More broadly, \cite{li2021ranking} explored various ranking measure including Top-k Recall/ accuracy, Pearson and Kendall-Tau. They showed that Kendall-Tau is highly correlated with the other metrics, and is a more stable indicator than other metrics such as the top-k accuracy.
In Fig.\ref{fig:correlation}, we provide a visual illustration comparing PT-Gauss to two other methods. We see that a high $\tau$ value is a strong indicator for better metric-error correlation as well as picking the best checkpoints with lowest test error (PT-Gauss, left). 
% Furthermore, comparing the checkpoints with the lowest metric (y-value) of each of the three figures, PT-Gauss clearly picked the best checkpoint with lowest test error (x-value).

\section{Additional Details of the VQA Experiments}
\subsection{VQA Architecture}
\label{sec:vqa_arch}
\begin{figure}
\resizebox{\linewidth}{!}{%
\includegraphics{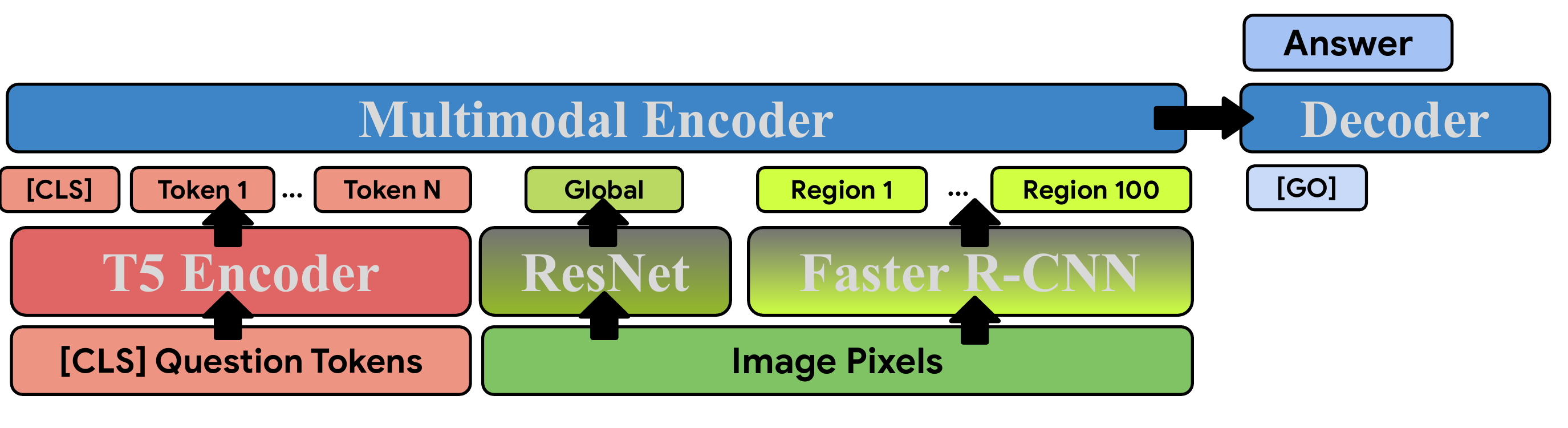}}
\caption{\small \textbf{VQA architecture} used in our experiments. The text encoder is initialized from a T5-base checkpoint, while the multimodal encoder is initialized from scratch. The parameters of ResNet152 and Faster R-CNN are frozen during VQA training.}
\label{fig:vqa_model}
\vspace{-15pt}
\end{figure}

We applied the state-of-art VQA model architecture, which fuses the image and question representations in a multimodal Transformer model~\cite{transformers}. See Fig.\ref{fig:vqa_model} as an illustration. On the image side, we take a global image feature from ResNet152~\cite{resnet} pretrained on ImageNet~\cite{imagenet} plus 100 region-of-interest image features from Faster R-CNN~\cite{frcnn} pretrained on Visual Genome~\cite{krishnavisualgenome}. The parameters of both ResNet152 and Faster R-CNN are frozen during training. On the question side, we use the text-encoder of a pretrained T5-base checkpoint~\cite{t5}. 
Finally, the decoder takes the [GO] symbol as the input and applies cross-attention to the outputs from the multimodal encoder, and outputs the answer. 

\subsection{VQA Datasets Descriptions}
\label{sec:vqa-pretrain-datasets}
\begin{enumerate}
    \item VQA v2.0: Visual Question Answering (VQA) v2.0~\cite{vqa2} is designed for answering open-ended questions about images. These questions require an understanding of vision, language and commonsense knowledge to answer.
    It is the second version of the VQA dataset~\cite{vqa1}. It has 265,016 images from COCO and abstract scenes and at least 3 questions (5.4 questions on average) per image.
    \item V7W: Visual7W~\cite{visual7w} is a large-scale visual question answering dataset, with object-level groundings and multimodal answers. Each question starts with one of the seven "W"s: what, where, when, who, why, how and which. It is collected from 47,300 COCO images and it has 327,929 QA pairs, together with 1,311,756 human-generated multiple-choices and 561,459 object groundings from 36,579 categories.
    \item GQA: The GQA dataset~\cite{gqa} centers around real-world reasoning, scene understanding and compositional question answering. It consists of 113K images and 22M questions of assorted types and varying compositionality degrees, measuring performance on an array of reasoning skills such as object and attribute recognition, transitive relation tracking, spatial reasoning, logical inference and comparisons.
    \item CNETVQA: the CNETVQA dataset was created based on the ConceptNet~\cite{speer2017conceptnet}. In particular, a T5 model was first finetuned for question generation with VQA v2.0 dataset, where each training example included a pair of phrases as the input: a randomly selected entity from the question and the answer, as well as the original question as the target. Once the T5-based question generation model was trained, it was applied on the edges of the ConceptNet which created about 300k question-answer pairs. Next, each of the QA pairs was matched by the top five images from the Google Image Search. After filtering out some too large or too small images, the final CNETVQA dataset contains about 1M total examples.
    \item TP-COLOR-COCO, TP-COLOR-CC3M, TP-COLOR-CC12M: Given image captions, we created three template-based VQA datasets consisting of visual questions about colors following a similar approach to color question generation proposed in COCOQA \cite{ren2015exploring}. More specifically, we first detected color mentions in the captions using a list of simple Wikipedia colors. We then used SpaCy dependency parsing instead of Stanford constituency parsing to extract the noun or the noun phrase associated with each color mention, as well as to group multiple colors for the same noun together. Finally, we filled in the templates (only singular variants shown): What color is the/this/that [object], What is the color of the/this/that [object], Is the/this/that [object] [color], Is the/this/that [object] [wrong color].
    We explored three sources of image captions: COCO-Captions~\cite{cococap}, CC3M~\cite{cc3m}, and CC12M~\cite{cc12m}. The number of question-answer pairs for TP-COLOR-COCO, TP-COLOR-CC3M, TP-COLOR-CC12M are 2.1M, 7,1M, and 38.9M, respectively.
    \item VQ2A-COCO, VQ2A-CC3M: We also took another approach to create VQA datasets from image captions, following \cite{vq2a}. These are the datasets generated by selecting various types of answer candidates from captions and then using a T5 XXL model trained for question generation and answering to generate questions and perform filtering. This results in 3.5M and 13.3M question-answer pairs for VQ2A-COCO and VQ2A-CC3M, respectively.
\end{enumerate}

\subsection{Finetuning}
\label{sec:finetune-vqa}
Similar to the NeuCRaB experiments, the finetunings on OKVQA was also done in two ways for each pretrained checkpoint: a full-model finetuning of the checkpoints following the learning schedule at pretraining time for another 100,000 iterations; and 5 top-layer-only finetunings using an L-BFGS solver with weight decay $\frac{1}{B} \cdot \cbr{0.01, 0.1, 1., 10., 100.}$, where $B$ is the size of the training set. 
The lowest test error of the above finetunings was used as the testing error of the checkpoint on the downstream task.

\subsection{Computation Platform}
The checkpoint pretraining, the pretrained feature extractions as well as the pretrained model finetunings were all done on the Google Cloud V2 4x4 TPUs. 
All the transferability metrics were computed on the Google Cloud Intel Skylake CPU (2GHz per core) with 1 core and 5GB RAM per run.

\subsection{Hyperparameters of PACTran-Gaussian}
\label{sec:vqa-hparams-pac-gauss}
\begin{figure}[!ht]
    \centering
    \includegraphics[width=0.45\textwidth]{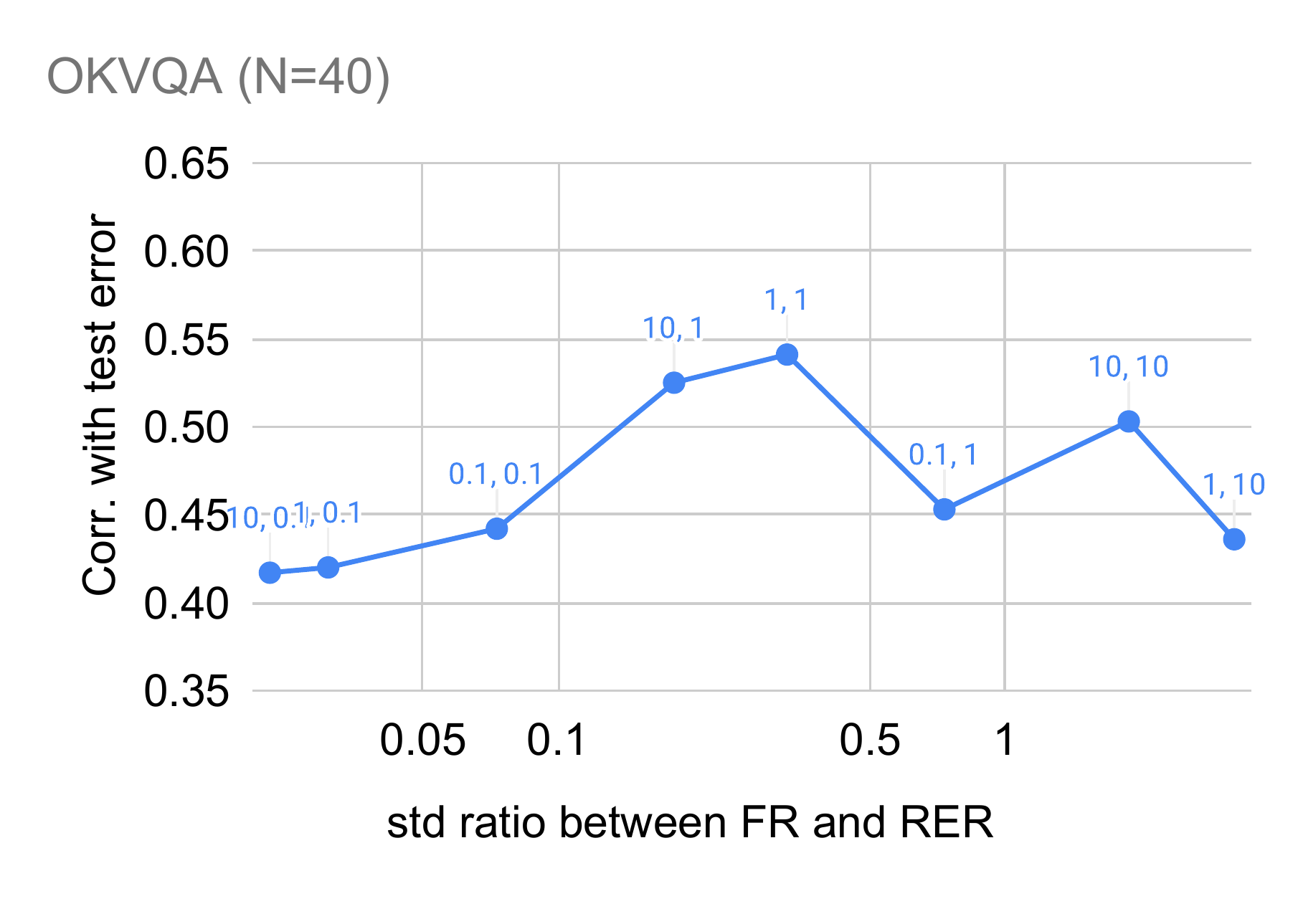}
    \includegraphics[width=0.45\textwidth]{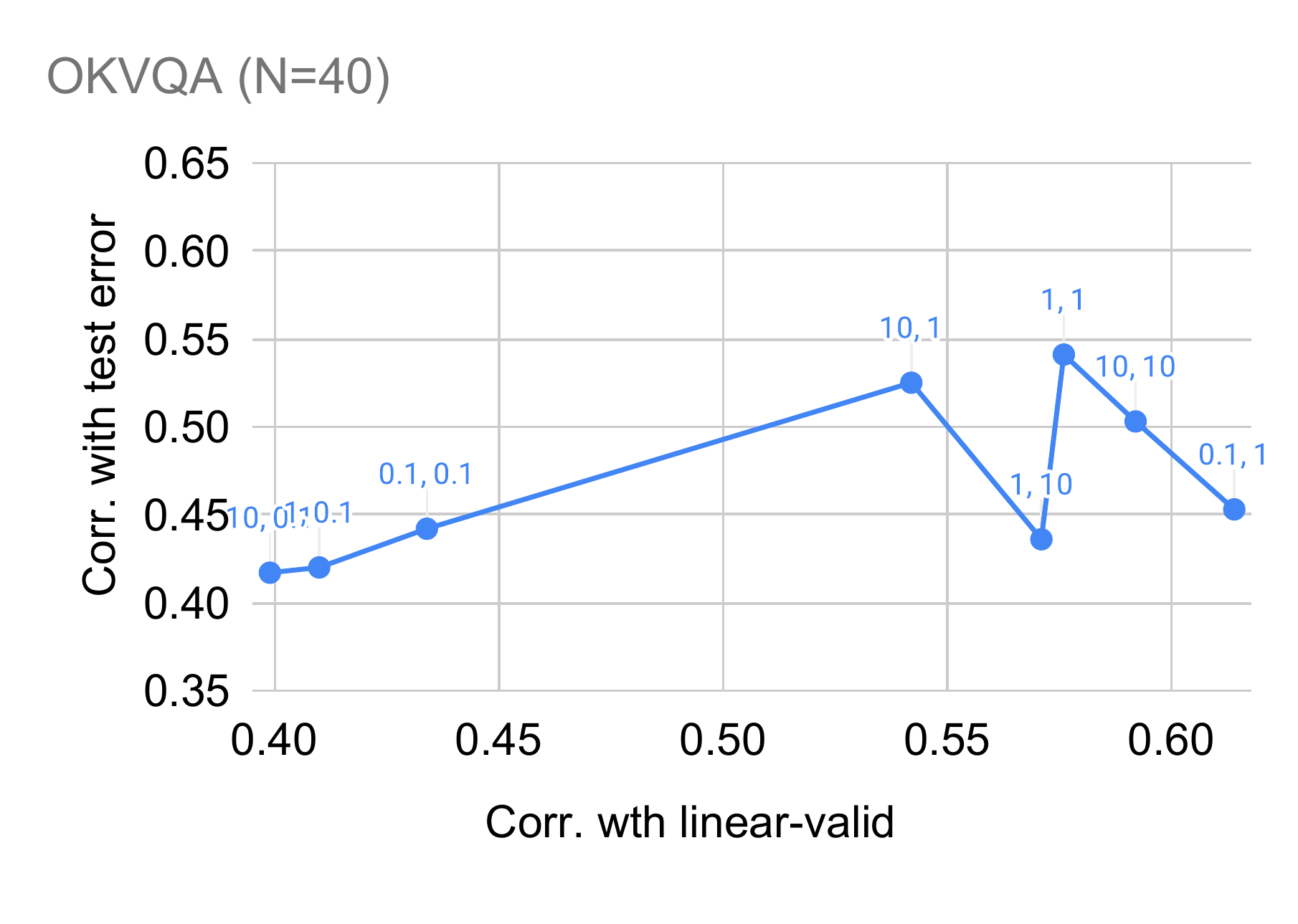}
    \includegraphics[width=0.45\textwidth]{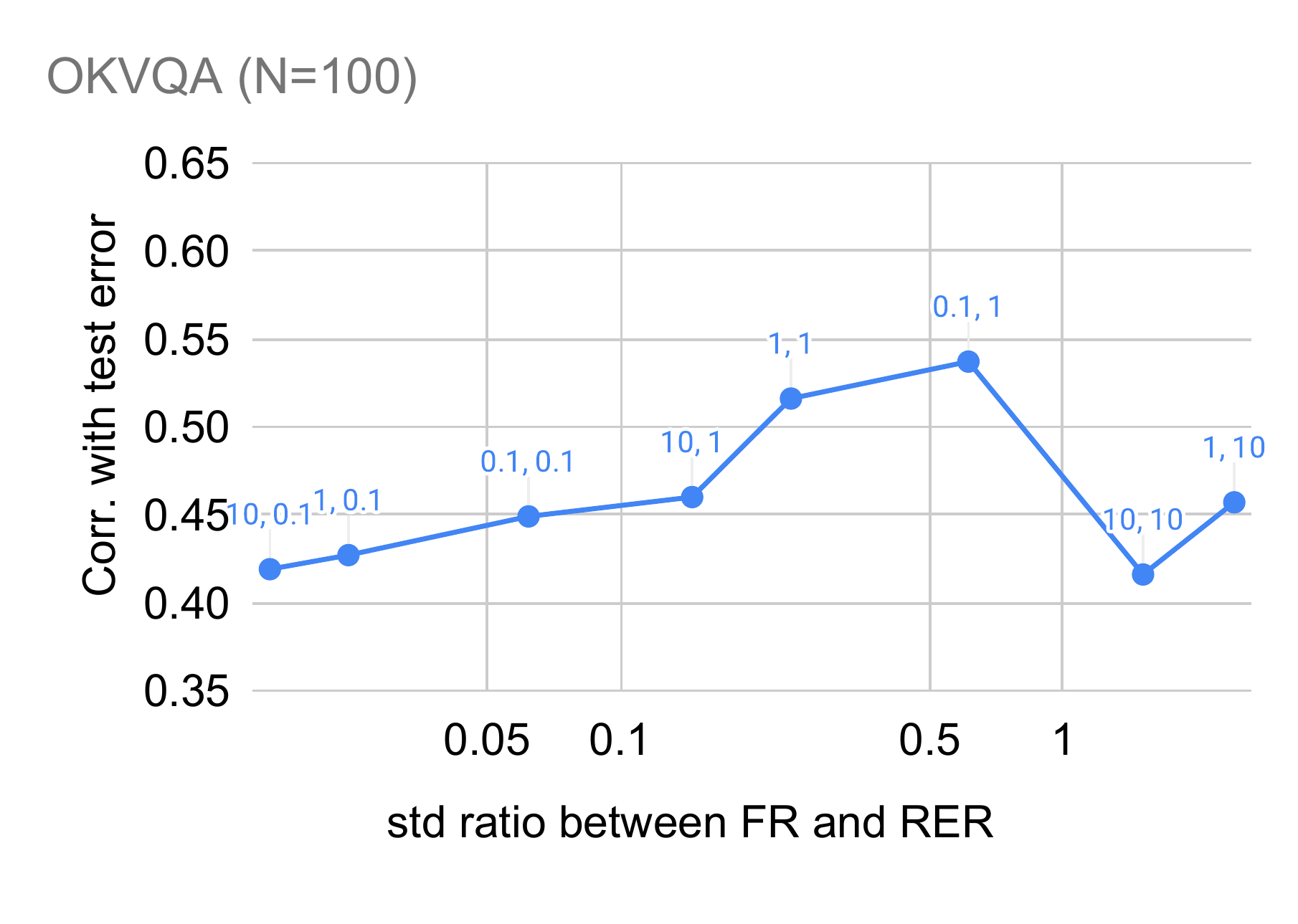}
    \includegraphics[width=0.45\textwidth]{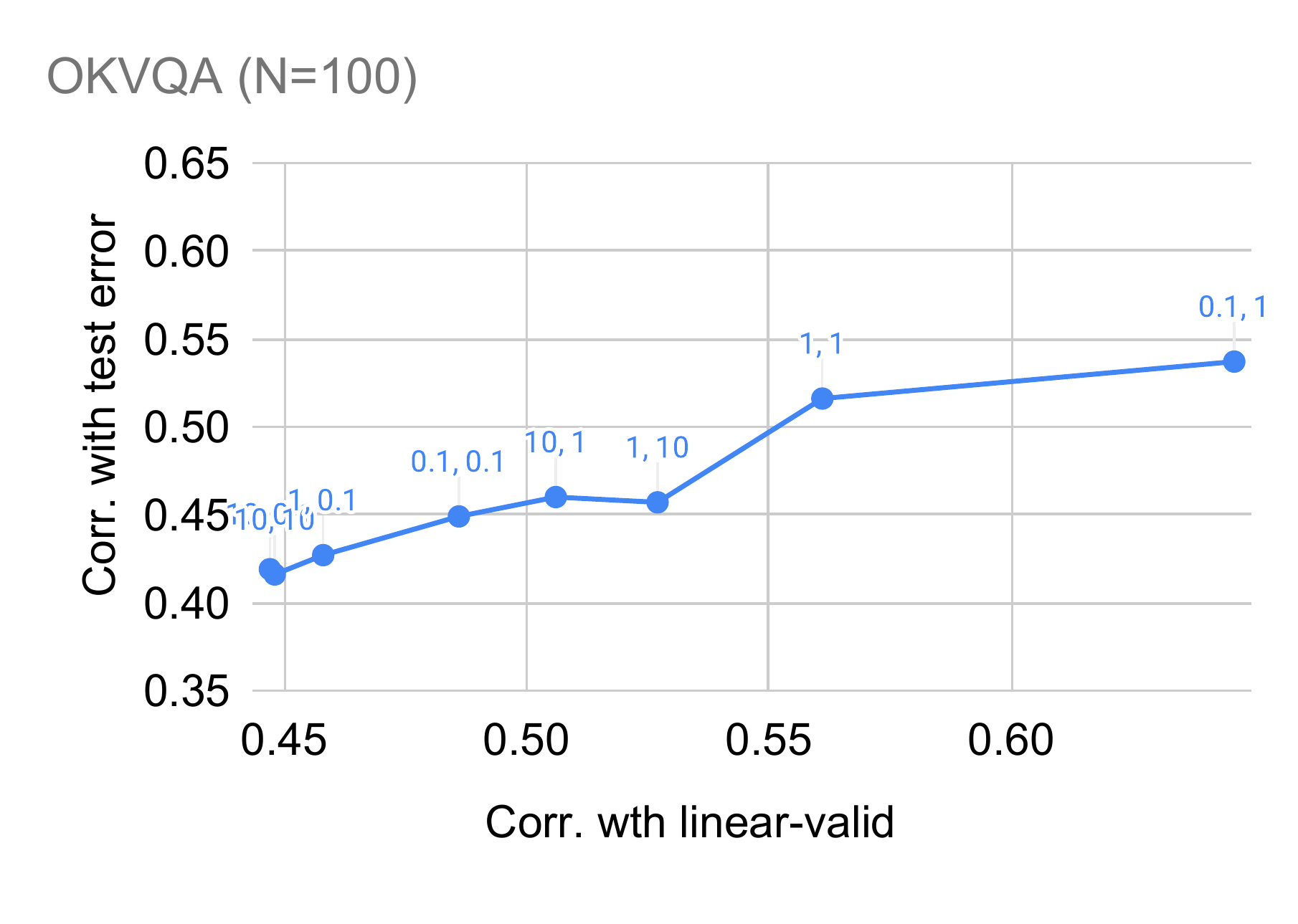}
    \includegraphics[width=0.45\textwidth]{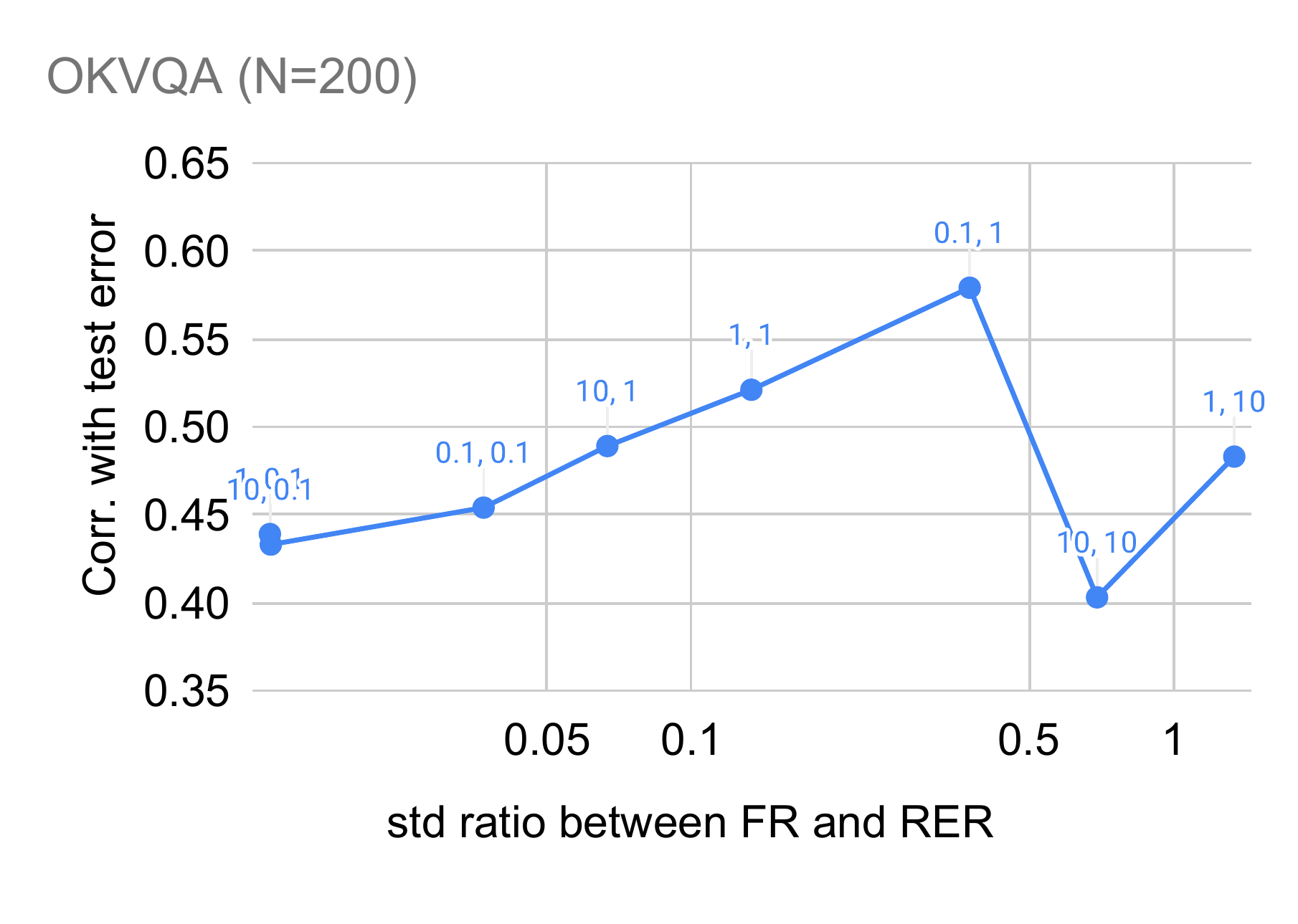}
    \includegraphics[width=0.45\textwidth]{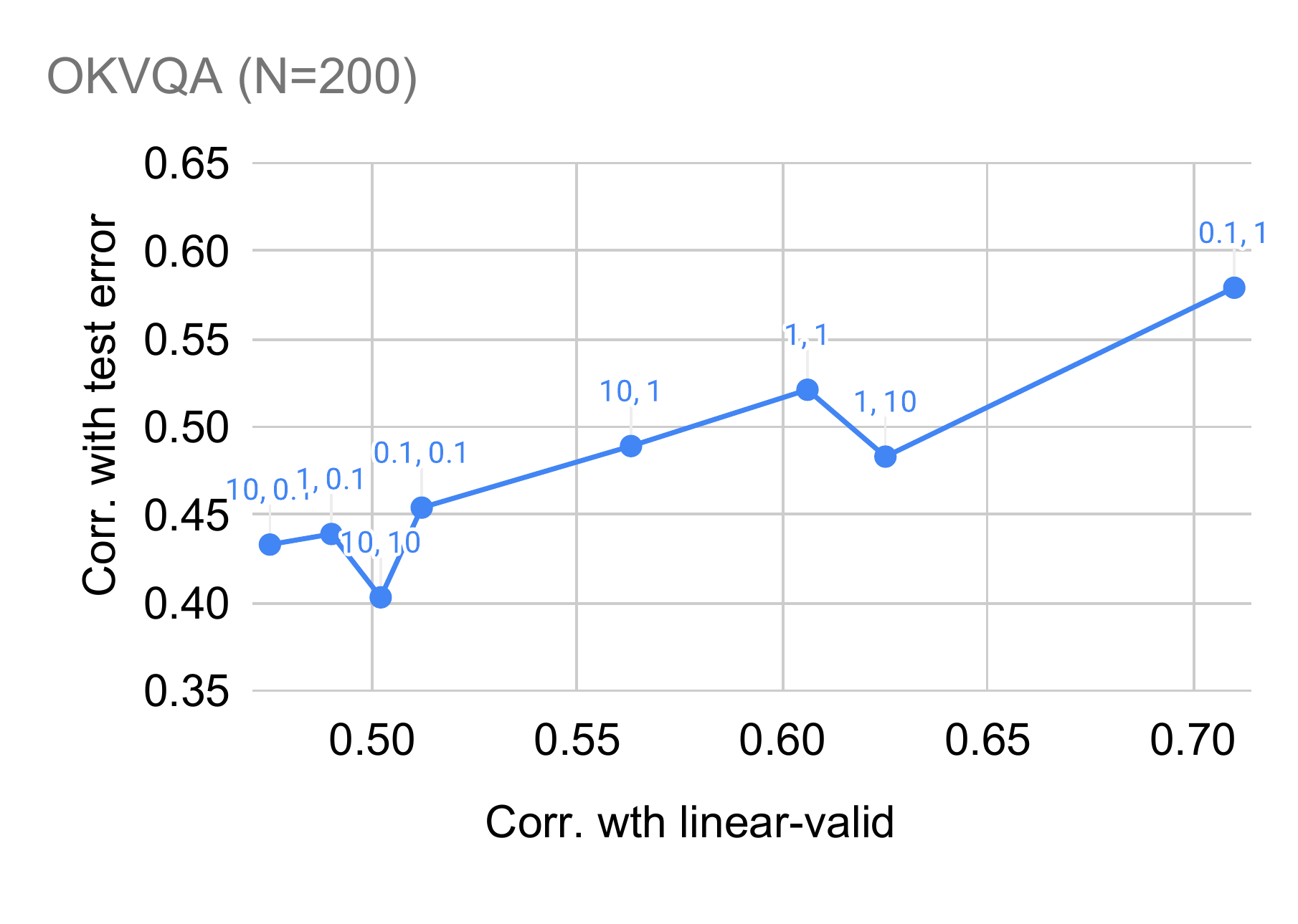}
    \caption{PACTran-Gaussian hyperparameter studies on OKVQA. Hyperparameters are labeled as $(a, b)$. High $y$-value indicates a good correlation with the downstream test error.}
    \label{fig:hp_okvqa}
\end{figure}

In Fig.\ref{fig:hp_okvqa}, we did an analysis of the different hyperparameters on the OKVQA experiment. We see a similar pattern to the previous ones in \ref{sec:hparams-pac-gauss}, except the optimal std ratio appears to be slightly lower ($\simeq 0.5$).

\subsection{GFLOPS in the OKVQA Experiment}
\label{sec:vqa-gflops}
In Table~\ref{tab:vqa-gflops} we report the GFLOPS of running the metrics as well as the GFLOPS of the feature extraction stage from the pretrained checkpoints in the OKVQA experiment. As we can see, the bottleneck is also the penultimate feature extraction, which is about 3-4 orders of magnitude slower than running the metrics themselves.
 \begin{table}[!ht]
\centering
\vspace{-0.1in}
  \begin{tabular}{c | c | c | c }
GFLOPS & $N=40$ & $N=100$ & $N=200$  \\\hline
LEEP & 7.44E-3 & 1.85E-2 & 3.69E-2\\
$\Ncal$-LEEP & 1.85E-2 &1.58E-1 &3.69E-1  \\
Hscore & 6.82E0 &6.86E0 &6.92E0 \\
LogME & 6.83E0 & 6.87E0 & 6.93E0 \\
LINEAR & 8.54E-1 & 2.10E0 & 4.19E0 \\
LINEAR-VALID & 2.85E-1 & 7.03E-1  & 1.40E0 \\ \hline
PT-Dir &7.58E-2 &1.86E-1 &3.71E-1  \\
PT-Gam &7.44E-2 &1.85E-1 &3.70E-1 \\
$\Ncal$-PT-Dir &1.86E-2 &1.58E-1 &7.93E-1  \\
$\Ncal$-PT-Gam &1.86E-2 &1.58E-1 &7.93E-1  \\
PT-Gauss$_{grid}$ & 8.58E-1 & 2.11E0 & 4.21E0 \\
%PT-Gauss$_{fix}$ & 1.91E-1 & 4.70E-1 & 9.36E-1 \\ 
\hline
Penultimate Feature (6, 3) &8.89E2 & 2.22E3 & 1.22E3 \\
Penultimate Feature (9, 5) &1.05E3 & 2.63E3 & 3.05E3 \\
Penultimate Feature (12, 7) &1.16E3 & 5.27E3 & 6.09E3 \\
  \end{tabular}
  \caption{GFLOPS of running each metrics and the penultimate-layer feature extractions on OKVQA. }
  \label{tab:vqa-gflops}
\vspace{-0.2in}
 \end{table}

\end{document}